\documentclass[article,onefignum,onetabnum]{siamart190516}

\usepackage{amssymb}
\usepackage{bm,bbm}
\usepackage{lipsum}
\usepackage{amsfonts}
\usepackage{graphicx}
\usepackage{cite}
\usepackage{epstopdf}
\usepackage{algorithmic}
\usepackage{etoolbox}
\usepackage{stackrel}
\usepackage{multirow}
\usepackage{enumerate}
\usepackage{overpic}
\usepackage[caption=false]{subfig}
\usepackage[normalem]{ulem}

\usepackage{xr-hyper}
\makeatletter
\newcommand*{\addFileDependency}[1]{
  \typeout{(#1)}
  \@addtofilelist{#1}
  \IfFileExists{#1}{}{\typeout{No file #1.}}
}
\makeatother

\usepackage{breqn}
\ifpdf
  \DeclareGraphicsExtensions{.eps,.pdf,.png,.jpg}
\else
  \DeclareGraphicsExtensions{.eps}
\fi

\renewenvironment{equation*}{\[}{\]\ignorespacesafterend}

\newcommand{\mcl}{\mathcal}
\newcommand{\mbf}{\mathbf}
\newcommand{\mbb}{\mathbb}

\newcommand{\eps}{\epsilon}

\newcommand{\Langle}{\left\langle}
  \newcommand{\Rangle}{\right\rangle}

\newcommand{\iidsim}{\stackrel{iid}{\sim}}
\newcommand{\dequal}{\stackrel{d}{=}}

\newcommand{\by}{\mbf y}

\newcommand{\bu}{\mbf u}

\newcommand{\bv}{\mbf v}

\newcommand{\be}{\mbf e}

\newcommand{\bx}{\mbf x}
\newcommand{\dd}{\mathrm{d}}
\newcommand{\R}{\mathbb{R}}

\newcommand{\PP}{\mathbb{P}}
\newcommand{\pom}{{\mbf u}^{\ast}}
\newcommand{\poU}{U^\ast}
\newcommand{\poC}{C^\ast}
\newcommand{\bchi}{\bar{\bm{\chi}}}

\newcommand{\bphi}{\bm{\phi}}

\newcommand{\Fnorm}[1]{\left\| #1 \right\|_F}
 
\newcommand{\tG}{\widetilde{G}}
\newcommand{\tW}{\widetilde{W}}
\newcommand{\tZ}{\widetilde{Z}}
\newcommand{\tN}{\widetilde{N}}
\newcommand{\tL}{\widetilde{L}}
\newcommand{\tD}{\widetilde{D}}
\newcommand{\tk}{\mathfrak{k}}

\newcommand{\biasc}{|\poC_\epsilon B\bu^\dagger_m/\gamma^2 - \bu^\dagger_m |}

\DeclareMathOperator*{\argmax}{arg\,max}

\crefname{hypothesis}{Hypothesis}{Hypotheses}
\newsiamthm{assumption}{Assumption}
\newsiamthm{thm}{Theorem}

\newsiamthm{claim}{Claim}
\newsiamthm{informaldefinition}{Informal Definition}
\newsiamremark{conjecture}{Conjecture}
\newsiamremark{remark}{Remark}
\newsiamremark{example}{Example}
\newsiamremark{hypothesis}{Hypothesis}
\newsiamremark{problem}{Problem}
\crefname{hypothesis}{Hypothesis}{Hypotheses}
\AtEndEnvironment{problem}{\null\hfill$\Diamond$}
\AtEndEnvironment{example}{\null\hfill$\Diamond$}
\AtEndEnvironment{remark}{\null\hfill$\Diamond$}

\headers{Posterior Consistency of SSR}{A. L. Bertozzi, B. Hosseini, H. Li, K. Miller, and A. M. Stuart}

\title{Posterior Consistency of Semi-Supervised Regression on Graphs
\thanks{Submitted to the editors DATE.
\funding{\color{black}This work is supported by NSF grant DMS 1818977, AFOSR grant FA9550-17-1-0185, NSERC PDF fellowship, a Caltech Von K{\'a}rm{\'a}n instructorship, DOD NDSEG Fellowship, and DARPA grant FA8750-18-2-0066.}}}

\author{
	Andrea L. Bertozzi\footnotemark[3]\and
	Bamdad  Hosseini\thanks{Computing and Mathematical Sciences, Caltech, Pasadena, CA
		(\email{bamdadh@caltech.edu}, \email{astuart@caltech.edu}).}
	\and
	Hao Li\thanks{Department of Mathematics, University of California, Los Angeles, Los Angeles, CA
		(\email{bertozzi@math.ucla.edu}, \email{lihao0809@math.ucla.edu}, \email{millerk22@math.ucla.edu}).}
	\and
	Kevin Miller\footnotemark[3]
	\and
	Andrew M. Stuart\footnotemark[2]
}

\usepackage{amsopn}

\ifpdf
\hypersetup{
  pdftitle={Bayesian Posterior Consistency},
  pdfauthor={Hao Li, Kevin Miller}
}
\fi

\usepackage[colorinlistoftodos,prependcaption,textsize=tiny]{todonotes}

\usepackage{comment}

\begin{document}
\maketitle
\begin{abstract}
Graph-based semi-supervised regression (SSR) involves
estimating the value of a function on a weighted graph
from its values (labels) on a small subset of the vertices; it can be formulated as a Bayesian inverse problem.
This paper is concerned with the consistency of SSR 
in the context of classification, in the setting where the labels 
have small noise and the underlying graph weighting is consistent
with well-clustered vertices.  
We present a Bayesian formulation of SSR in which the weighted graph 
defines a Gaussian prior, using a graph Laplacian, and the labeled 
data defines a likelihood. We analyze the rate of contraction of
the posterior measure around the ground truth in terms of parameters
that quantify the small label error and inherent clustering
in the graph. We obtain bounds on the rates of contraction 
and illustrate their sharpness through numerical experiments.
The analysis also gives insight into the choice of hyperparameters
that enter the definition of the prior.
\end{abstract}



\section{Introduction} \label{sec:intro}
Semi-supervised learning (SSL) is the problem of labeling all points 
within a dataset (the {\it unlabeled data}) by combining knowledge of a 
subset of noisy observed labels (the {\it labeled data}); this is done
by exploiting correlations and geometric information present in the dataset combined with label information.
We study this problem in the framework
of Bayesian inverse problems (BIPs), building on a widely adopted 
semi-supervised regression (SSR) approach to SSL 
developed in the machine-learning community.
In this context, the Bayesian formulation has a novel 
structure in which the unlabeled data defines the prior distribution 
and the labeled data defines the likelihood. The goal of this article 
is to study posterior consistency; that is, the contraction of the 
resulting Bayesian posterior distribution onto the ground-truth solution in 
certain parametric limits related to parameters underlying our model.
We adopt ideas from spectral clustering in unsupervised learning to construct and
analyze the prior arising from a similarity graph constructed 
from the unlabeled data. This prior information interacts 
with the labeled data via the likelihood.
An interesting feature of the Bayesian SSR posterior consistency analysis 
is the fact that the unlabelled data is used to construct the prior while the labelled data enters the likelihood;
this is in contrast to standard 
formulations of posterior consistency in BIPs where data is only used to define the likelihood while 
the prior is fixed, independently of data.
In the setting we consider, and when the prior information  and the likelihood complement
each other, then a form of Bayesian posterior consistency can be established
and the posterior measure on the predicted labels contracts 
around the ground truth. 
Furthermore our analysis elucidates how 
hyperparameter choices in the prior, quantitative measures of clustering
in the dataset and the noise in labels combine to affect the 
contraction rates of the posterior. In the following three subsections, 
we review relevant literature, formulate the problem mathematically 
and describe our contributions.

\subsection{Relevant Literature}\label{sec:relevant-literature}%
Many approaches to SSL and SSR have been developed in the literature
and a detailed discussion of all of them is outside the scope of this article.
We refer the reader to the  review articles  \cite{zhu2005semi}
and \cite{kostopoulos2018semi} for, respectively, the state-of-the-art in 2005
and a more recent appraisal of the field.

The consistency of
supervised learning and regression is well-developed; see \cite{tewari2007consistency}
for a literature review, as well as the preceding work
in \cite{steinwart2001influence,steinwart2005consistency,wu2006analysis}
which establish the problem in the framework of Vapnik \cite{vapnik1998statistical}.
All of this work on
supervised classification focuses
on the large data/large number of features setting,
and often considers only linearly separable unlabeled data.
Therefore, these previous works do not leverage the power of graph-based techniques to extract
geometric information in large unlabeled datasets, a primary feature of the
SSR problems studied in this work.

Graph-based techniques are widely used in
unsupervised learning
\cite{belkin2002laplacian,von2007tutorial}, a subject
that has seen significant analysis in relation to consistency. The papers
\cite{spielmat1996spectral, spielman2007spectral} perform a careful analysis
of the spectral gaps of graph Laplacians resulting from clustered data,
studying recursive methods for multi-class clustering. The paper
\cite{ng2002spectral} introduced an approach for the analysis of
multi-class unsupervised learning based on perturbations of a perfectly clustered case.
The paper \cite{von2008consistency} introduced the idea of studying the
consistency of spectral clustering in the limit of large independent and identically distributed (i.i.d.)~datasets
in which the graph Laplacians converge to a limiting integral
operator. The articles \cite{trillos2018error, trillos2016variational} took 
this idea further by proving the convergence of graph Laplacian operators to local differential
operators by controlling the local connectivity of the graph as a function
of the number of vertices. 

In this paper, our focus is on transductive SSL \cite{kostopoulos2018semi}
in the framework of
the influential papers \cite{zhu2003semi,zhu2003combining} where the
categorical labels $\{1, \dots, M\}$ are
embedded in $\R^M$ and the SSR approach to SSL is adopted.
Bertozzi and Flenner \cite{{bertozzi2016diffuse}} introduced an interesting relaxation of
this assumption, by means of a Ginzburg-Landau penalty term which favors real-values
close to $\pm 1$ but does not enforce the categorical values $\pm 1$ exactly. In contrast to these relaxations,
the probit approach to classification, described in the classic text on Gaussian process
regression \cite{rasmussen2006gaussian} and analyzed in \cite{HHRS19} in the context
of SSL, works directly with the
categorical labels and does not rely on the embedding step.

The idea of regularization by graph Laplacians for SSL was developed in different
contexts such as manifold regularization
\cite{belkin2006manifold}, Tikhonov regularization \cite{belkin2004regularization}
and local learning regularization \cite{wu2007transductive} as well as 
more recent articles focusing on large data settings \cite{vanZanten18, vanZanten2019}.
However, while graph regularization methods are widely applied in practice
the rigorous analysis of their properties,
and in particular asymptotic consistency and posterior contraction rates, are
not well-developed within the context of SSL and SSR. 
Indeed, to the best of our knowledge the Bayesian consistency of SSR
has not been analyzed. Studying
SSL/SSR in a Bayesian setting introduces new challenges that require careful consideration about assumptions regarding graph structure and statistical properties of the resulting model \cite{kirichenko2017}. We build on the spectral analysis of the
graph Laplacian introduced in \cite{ng2002spectral} to study 
unsupervised learning, and refined in \cite{HHRS19} 
to study the consistency of optimization-based
approaches to binary and one-hot SSL. 

The preceding discussion shows connections between SSL/SSR and the classical
theory of inverse problems, and in particular the use of graph-based
analogs of Tikhonov regularization \cite{engl}. The use of data to construct regularizers
is emerging as an important area in the classical approach to inverse problems
\cite{arridge2019solving,soh2019learning}. Here we work in the context of Bayesian
inverse problems \cite{dashti2013bayesian,kaipio2006statistical}.
The subject of Bayesian posterior consistency is aimed at
reconciling the large data limits of frequentist and Bayesian approaches
to statistical inference problems. Early influential works in this field concentrated on
negative results concerning the Bayesian nonparametric setting
where the prior and likelihood were inconsistent \cite{diaconisf}. Subsequent
work in this area concentrated on
positive results, demonstrating that minimax rates of convergence
can be obtained within the Bayesian setting \cite{ghosal,bvm2} by studying
posterior measure concentration through Bernstein--Von Mises-type 
theorems \cite{bvm,bvm2} provided that priors are constructed
carefully. The celebrated paper \cite{brownlow} demonstrates
how large data  and small noise limits are intimately related,
and this link underpins subsequent studies of inverse problems from the
perspective of Bayesian posterior consistency. This line of work was initiated
in the paper \cite{vdw} where the small noise limit of linear inverse
problems was studied. A number of papers in this area followed
\cite{agapiou,ray} and it is currently an active research area, particularly in
relation to nonlinear inverse problems \cite{nickl2016infdimstat}.

In some problems, optimization approaches rather than fully Bayesian
approaches are adopted, and the study of consistency for inverse
problems in this setting is overviewed in \cite{engl}. Linking this to
maximum a posteriori (MAP) estimators for inverse problems was
a subject developed in \cite{dashti} and the study of consistency for
MAP estimators in semi-supervised learning, and in particular use of the
probit likelihood model, is undertaken in \cite{HHRS19}. 
The aforementioned papers \cite{zhu2003semi,zhu2003combining} correspond
precisely to application of MAP estimation to the SSR we study in this paper.
Since the MAP estimator of a Gaussian posterior distribution is equal to its mean,
and since our results directly imply consistency of the posterior mean, this paper
may be viewed as providing theoretical justification for the algorithm proposed in
\cite{zhu2003semi,zhu2003combining}.

\subsection{Problem Setup} \label{sec:prob-setup}
Consider a set of vertices $Z = \{ 1, \cdots, N \}$
and an associated
set of {\it feature vectors}
$X = \{ \bx_1, \bx_2, \cdots, \bx_N\}$. Each feature
vector $\bx_j$ is assumed to be a point in $\mbb R^d$.
$X$ may thus be viewed as a function $X: Z \mapsto \mbb R^d$ or
as a matrix in $\mbb R^{d \times N}$ with columns given by $\bx_j$. We refer to $X$ as the {\it unlabeled data}.
Throughout
this article we assume that every element of
$Z$ belongs to one of $M$ classes and
employ the one-hot encoding to represent the label of each point.
More precisely,
we assume there exists
a function $l: Z \mapsto \{\be_1, \cdots, \be_M\}$ where the $\be_j \in \mbb R^M$
are the standard coordinate vectors. A point $j \in Z$ then belongs to class $m$
if $l(j) = \be_m$.

Now let $Z' \subseteq Z$ be a subset of $J\le N$ vertices and define a
function $Y: Z' \mapsto \R^M$, noting that this may also be
viewed as a matrix $Y \in \mbb R^{M \times  J}$. The columns of $Y$
are denoted by $\{\by'_1, \cdots, \by'_J\}$ and comprise a 
collection of {\it noisy observed  labels} on $Z'$;
in practice, we use $\by'_j \in \{\be_1, \cdots, \be_M\}$, the one-hot vectors,
or small noisy perturbations of this setting.
We refer to $Y$ as the {\it labeled data}.
Underlying this paper is the assumption that the labeled data 
is determined by a generative model of the form
\begin{equation}
  \label{label-model}
  Y = U^\dagger H^T + \gamma \eta.
\end{equation}
Here  $U^\dagger \in \mbb R^{M \times N}$ is the {\it ground-truth latent variable} that
gives the true labels of all of the vertices in $Z$,
$H \in \R^{J \times N}$ is the submatrix consisting of the $Z'$ rows of the identity matrix $I_N \in \R^{N \times N}$ and 
$\eta \in \R^{M \times J}$ is a matrix
with independent standard Gaussian entries, i.e.,
$ \eta_{mj} \iidsim \mcl N(0,1)$.
The parameter $\gamma >0$ is the standard deviation of the observation noise.  
It is instructive to think of the columns of $U^\dagger$ as being chosen
from $\{\be_1, \cdots, \be_M\}$, although generalizations of this setting
are possible. 

The model \eqref{label-model} casts the SSL problem of inferring the
true labels on $Z$ as the SSR problem of
finding $U^\dagger$, adopting the terminology of\cite{kostopoulos2018semi}:
our modeling assumption makes the observations $Y$ real-valued,  
rather than categorical as in classification, and therefore is considered 
a regression problem.
The SSR problem is ill-posed, requiring the learning of $NM$ parameters
from $JM$ noisy data points, since we typically have far fewer
labels than the total number of unlabeled data points, i.e. $J \ll N$.
The labeled data may be viewed as providing prior information that renders this ill-posed problem tractable. 
To this end, we formulate SSR 
in the framework of Bayesian linear inverse problems \cite[Ch.~8]{calvetti2007introduction}.

The main goal of this article is to analyze the consistency of the Bayesian SSR problem by identifying the conditions
under which the posterior measure $\mu^Y$ (defined in \eqref{bayes-rule} below) contracts around the ground-truth matrix $U^\dagger$ in \eqref{label-model}.
Formally, we define the following functional 
as a measure of  posterior contraction
\begin{equation}
  \mcl I :=
  \mathbb{E}_{Y|U^\dagger}\mathbb{E}_{\mu^Y} \Fnorm{U - U^\dagger}^2,
  \label{eq:contraction}
\end{equation}
where the inner expectation is with respect to the posterior measure
$\mu^Y$ on $U$ while the outer expectation is with respect 
to the law of
$Y | U^\dagger$ following \eqref{label-model}; $\Fnorm{\cdot}$
denotes the Frobenius norm. In other words, for a given ground truth $U^\dagger$, $\mcl I$ measures the mean-squared error of the posterior measure $\mu^Y$ on $U$, averaged over the possible observations $Y$ arising from $U^\dagger$ according to the generative 
model given by equation \eqref{label-model}. With this notation,
our aim is to solve the following problem:

\begin{problem}[Posterior consistency of Bayesian SSR]\label{prob-post-consistency-bayesian-SSR}
  Under what conditions on the graph $G$, the labeled set $Z'$, 
the ground truth $ U^\dagger$ and other hyperparameters entering the definition of the
prior
can we ensure that $\mcl I \downarrow 0$ as the noise-level $\gamma$ in the
unlabeled data, and
some measure $\epsilon$ of closeness to perfect clustering in the labeled data, tend to zero.
\end{problem}

Indeed we will find explicit  bounds on $\mcl I$ 
which give consistency in the limit $(\epsilon,\gamma) \to 0$
and reveal the role of model parameters in the
form of the contraction rate. Our bounds are applicable 
for small values of $\gamma, \eps$ (the explicit condition under which the bounds hold
will be presented) and not just in the asymptotic regimes where $ ( \eps, \gamma) \to 0$. 

\subsection{Main Contributions} \label{sec:informal-results}
We study posterior contraction, as measured by the quantity 
$\mcl I$. In the theory we develop, the quantity 
of labeled data and unlabeled data will be fixed, a 
practically useful setting in which to study algorithms based
around SSR. The prior that we use is a discrete analog of the
Mat{\'e}rn prior with graph Laplacian used in place of the
continuum Laplacian in the differential operator formulation
popularized in \cite{lindgren2011explicit}. 
We show in Section~\ref{sec:bayes-form-ssl} that the resulting posterior measure $\mu^Y(\dd U)$ takes the form
\[
    \mu^Y(\dd U)
 \propto \exp\left(- \frac{1}{2\gamma^2} \Fnorm{U H^T - Y}^2
  - \frac{1}{2} \left\langle U^T, C_{\tau}^{-1}U^T\right\rangle_F
  \right) \dd U,
\]
where $\| \cdot \|_F$ and $\langle \cdot, \cdot, \rangle_F$ denote the Frobenius norm and  inner product while 
$C_\tau = \tau^{2\alpha}(L + \tau^{2\alpha}\mathrm{I}_N)^{-\alpha}$ is the prior covariance matrix with $L$ denoting a 
graph Laplacian matrix and parameters $\alpha, \tau^2 >0$. We interpret $\tau$ as 
an inverse length-scale in the space of the feature vectors, and $\alpha$ as a 
regularity parameter for the prior controlling the number of relevant eigenvectors of $L$ which are active in the
Bayesian SSR approach. The parameter $\gamma$ is the noise standard deviation
in \eqref{label-model}. We also introduce a  parameter $\epsilon$ which characterizes the geometry of the underlying graph.  
This parameter is formally defined  through the notion of a weakly connected 
graph as introduced in \cite{ng2002spectral} and used in \cite{HHRS19}:
 
\begin{informaldefinition}[Weakly connected graph]\label{weakly-connected-graph-informal}
Let $0 < \eps \ll 1$, then a graph $G = \{ Z, W\}$ is weakly connected with $K$ clusters if it consists of pathwise connected components $\tG_k= \{ \tZ_k, \tW_k\} $ for $k=1, \dots, K$ so that the edge weights between elements in different $\tG_k$ are $\mcl O(\epsilon)$.  In other words, up to a reordering of $Z$, the matrix $W$ is an $\mcl O(\eps)$ perturbation of a block diagonal weight matrix, and the graph Laplacian associated with each block has a one-dimensional null-space.  
\end{informaldefinition}

This informal definition will be made precise later on in Subection~\ref{sec:weakly-conn-graph}. We now present an informal version of our 
main result whose precise statement is given as 
 as Corollary~\ref{cor:main-a} to 
Theorem~\ref{thm:main}, both stated and proved 
in Section \ref{sec:main-contribution}.

\vspace{0.1in}
{\sc Main Theorem.}
 \emph{Let $G = \{ Z, W\}$ be weakly connected with $K$ components $\tG_k$ and  perturbation
  parameter $0 < \eps \ll 1$ as in
Definition~\ref{weakly-connected-graph-informal}.
  Suppose that the rows of the
   ground-truth matrix $U^\dagger \in \mbb R^{M \times N}$  belong to the
  span of the indicator functions of the $\tG_k$,
fix $\alpha>0$ and choose $\tau$ so that
$$ \eps = \eps_0\tau^{\max\{2, 2\alpha\}}.$$
Then, for appropriately chosen $\eps_0$,
there exists $\Xi>0$, independent of $\eps$ and $\gamma$, so that
   \begin{equation*}
     \mcl I \le \Xi \max \left\{ \gamma^2, \eps^{\min\{1, \alpha\}} \right\}.
   \end{equation*}
}

\begin{figure}
\centering
\includegraphics[width=1\textwidth]{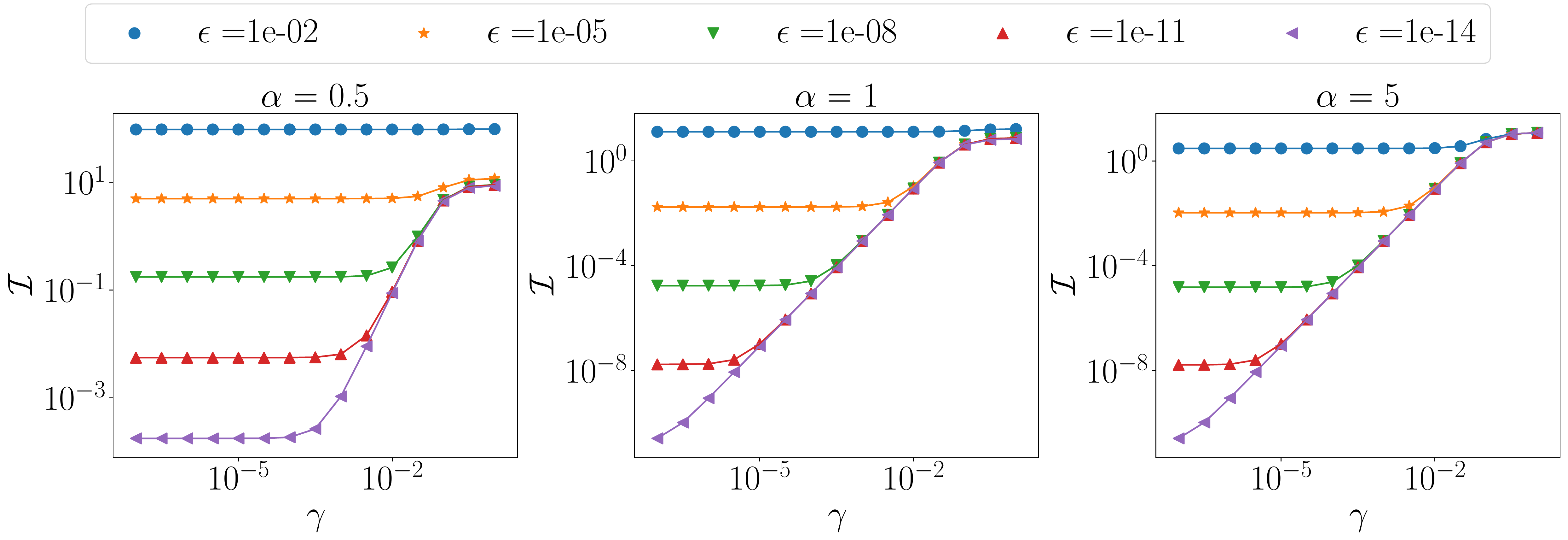}
\vspace{-0.3in}
\caption{A numerical demonstration of the Main Theorem on 
a synthetic dataset (detailed in Subsection \ref{sec:synthetic-data}). Details of this experiment are described in Section~\ref{sec:numerical-experiments}.
 The value of $\mcl I$ reduces with $\gamma$ up to the point where $\gamma^2 \approx \eps^{\min\{ 1, \alpha\}}$ where the errors saturate as predicted by the upper bound in
  the Main Theorem.
    Smaller values of $\eps$ result in smaller values of $\mcl I$
    that indicates higher concentration of posterior probability mass
    around the ground truth $U^\dagger$. }
  \label{fig:I vs gamma}
\end{figure}

Let us give insight into this theorem. The parameters $\eps$ and $\gamma$ are
inherent to the specific SSR problem and the dataset at hand. Broadly speaking $\eps$ is a geometric property of
the point cloud $X$ of unlabeled data, quantifying how clustered it is, and $\gamma$ is the noise standard deviation of 
the labels, quantifying how accurate the labels are. Hence these parameters
are fixed, although they are generally unknown. Then the Main Theorem
implies the following regarding the error in Bayesian SSR: 

\begin{itemize}
\item If $\eps^{\min\{1, \alpha\}} \le \gamma^2$, then
 the label measurement noise ($\gamma$) dominates over the measure of closeness to perfect
clustering ($\eps$) and so posterior contraction is controlled by
 the $\gamma$ parameter.
\item If $\gamma^2 < \eps^{\min\{1, \alpha\}}$,  then the measure
of closeness to perfect
clustering is dominant in comparison to the
 label measurement noise, and posterior contraction is controlled 
by the $\eps$ parameter.
\end{itemize}
The Main Theorem also has the following implications regarding choice of
parameters $\alpha,\tau$ entering the prior:
\begin{itemize}
\item{The length-scale $\tau$ needs to be tuned depending on  $\epsilon$ the measure of how clustered the
data is}.
\item In the case  $\gamma^2 < \eps^{\min\{1, \alpha\}}$ we observe that choosing
$\alpha <1$  gives a sublinear 
contraction rate in $\eps$ while a linear rate is achieved if $\alpha \ge 1$. Thus it is preferable to tune  $\alpha$ so that
  $\alpha \ge 1$.
\footnote{For reasons related to the large data limit $N \to \infty$, it is natural when $N \gg 1$
to choose $\alpha>\frac{d}{2}$ and since $d$ is typically larger than $2$,
this enforces $\alpha>1$; see \cite{HHOS19}.} 
\end{itemize}
Since $\epsilon$ is not known this suggests the importance of estimating it from data, and more
generally of using hierarchical methods to determine $\alpha,\tau.$

These insights are also supported by our numerical experiments in 
Section~\ref{sec:numerical-experiments}; furthermore these
experiments also verify the sharpness of the upper bound 
in the Main Theorem. As a prelude to these detailed experiments,
Figure~\ref{fig:I vs gamma} contains the results of a computational 
example which illustrates our main theorem on a synthetic dataset. We 
postpone details of this experimental set-up to 
Section~\ref{sec:numerical-experiments}, but studying the figure
at this point already gives useful insight:
for fixed values of $\eps$ the  value of $\mcl I$ goes to zero at a rate
proportional to $\gamma^2$ until an inflection point,
around $\gamma^2 \approx \eps^{\min\{1,\alpha\}}$,
after which the error saturates; the saturation levels themselves 
go to zero like $\eps^{\min\{1, \alpha\}}$. These facts are
exactly as predicted by our theory. 

The rest of this article is structured as follows. 
We outline the details of the Bayesian SSR problem in Section~\ref{sec:bayes-form-ssl}, introducing the
likelihood and the prior in Subsections~\ref{sec:likelihood} and \ref{sec:prior-constr-using}
followed by an analytic expression for the posterior measure 
in Subsection~\ref{sec:posterior}.
Section~\ref{sec:main-contribution} is dedicated to our consistency analysis and presents detailed
versions of our primary results that are summarized in the Main
Theorem.
We first analyze the disconnected graph case in Subsection~\ref{sec:disconnected-graph}
to gain some insight into the behavior of the posterior. We then study the weakly connected graph
setting in Subsection~\ref{sec:weakly-conn-graph}.
We present the proofs of these results, relying on lemmata that are stated in Section~\ref{sec:main-contribution}, but deferring their proof to Appendix~\ref{sec:proof-main-results}.
We collect numerical experiments in Section~\ref{sec:numerical-experiments} that demonstrate the
sharpness of the contraction rates and bounds obtained in Section~\ref{sec:main-contribution}. 
We present experiments which illustrate situations in 
which the label noise dominates the
closeness to clustering, and vice versa.
We conclude the article in Section~\ref{sec:conclusions} 
with further discussion, including potential new lines of
research stemming from  our results. 
Appendix~\ref{sec:proof-main-results} contains the detailed proofs of the 
lemmata that support the main theoretical results developed in Section~\ref{sec:main-contribution}; 
these are also illustrated by numerical results presented in Subsection~\ref{sec:numer-supp-lemm-1} in the supplemental material. 
We include a summary of spectral analysis results from~\cite{HHRS19} that are used routinely
throughout the  proofs in Appendix~\ref{app:results-HHRS}.

\section{Bayesian Formulation Of SSR}\label{sec:bayes-form-ssl}

In this section we outline the Bayesian formulation of the SSR problem in detail. We derive the likelihood potential
$\Phi$ in Subsection~\ref{sec:likelihood} and construct the prior
measure in Subsection~\ref{sec:prior-constr-using}.
An analytic expression for the posterior measure is given in Subsection~\ref{sec:posterior}.

Throughout the following we let $\langle\cdot, \cdot\rangle$ denote the
Euclidean inner product and $|\cdot|$ the Euclidean norm;
we use $\| \cdot \|_2$ to denote the induced operator Euclidean norm
on matrices. Recall that $\|  \cdot \|_F$ denotes the Frobenius norm 
on matrices and define $\langle A,B\rangle_F := \mathrm{Tr}\left(A^TB\right)$,
the inner-product which induces this norm. We  
 use $\otimes$ to denote the Kronecker product between matrices.
Occasionally we use $|S|$ to denote the  cardinality of a set $S$;
confusion with the Euclidean distance should not arise as we
will clarify the notation based on the context.

\subsection{The Likelihood}\label{sec:likelihood}
Based on the generative model \eqref{label-model} for the labeled data 
$Y \in \R^{M\times J}$, we define 
the likelihood distribution $\PP(Y|U)$ with density proportional to
\begin{align}
\exp\left(-\frac{1}{2\gamma^2}\left\|UH^T-Y\right\|_F^2\right),
\end{align}
recalling that $\eta = \frac{1}{\gamma}(U H^T - Y) \in \mbb R^{M \times J}$ has independent standard Gaussian entries $\eta_{mj} \sim \mcl N(0,1)$.
It is therefore convenient to define the likelihood potential
\begin{equation}\label{SSR-likelihood}
  \Phi: \R^{M \times N} \times \R^{M \times J} \mapsto \R^+, \qquad
  \Phi(U; Y) := \frac{1}{2\gamma^2} \| U H^T - Y \|_F^2.
\end{equation}

\begin{remark}
  We note that if the entries of the noise $\eta$ are not independent but rather correlated, then the expression
  \eqref{SSR-likelihood} needs to be modified by weighting
the $\Fnorm{\cdot}$ norm 
  by the inverse square root of the covariance operator of $\eta$.
This will make no significant difference to what follows and we work
with i.i.d.~noise only to simplify the exposition.
\end{remark}

\subsection{The Prior}\label{sec:prior-constr-using}
We now detail the construction of the Gaussian prior measure for $U \in \mbb R^{M \times N}$, whose $M$ rows reflect the labeling of the $N$ data points in $X$ into $M$ corresponding classes. We also demonstrate how this prior expresses the geometric information in the unlabeled data $X$. We construct a weighted graph $G = \{Z, W\}$ with vertices $Z$ 
and self-adjoint weighted adjacency matrix $W = (w_{ij})$. The weights
$w_{ij} \ge 0$ reflect the affinity of data pairs  $(x_i, x_j) \in X \times X$,
the edge set of the graph.
For example,  we may  construct $W$
using a kernel $\kappa: \R_+ \rightarrow \R_+$
by setting
  \begin{equation}
    \label{w-ij-definition}
    w_{ij} = \kappa( |\bx_i - \bx_j|).
  \end{equation}
The kernel $\kappa$ is assumed to be positive, non-increasing, and
with bounded variance; a natural example is the
Gaussian kernel   $\kappa(t) = \exp\left( - |t|^2/r^2 \right)$,
or 
the 
indicator function of the interval $[0,r]$, both 
  with bandwidth $r \in \mathbb{R}^+$. \footnote{We note that data-driven choice of kernel $\kappa$ is an active area of research
  in supervised learning \cite{jacot2018neural,owhadi2019kernel,owhadi2020ideas}, and is also potentially relevant here, but is outside the scope of our present discussion.}
Note that \eqref{w-ij-definition} implies that $W$ is symmetric
and the suggested weight constructions lead to $w_{ij}$ which 
encode the pairwise similarities between the  points in $X$.

Given a weight matrix $W$ with the properties illustrated by this
explicit construction, we introduce a \emph{graph Laplacian} operator 
on $G$ of the form
\begin{equation}
  L = D^{-p}(D - W)D^{-p},\label{graph-laplacian}
\end{equation}
where  $D = \mathrm{diag}\{d_{i}\}$ with entries $d_{i}:=
  \sum_{j\in Z}w_{ij}$ is the diagonal degree matrix and  $p \in \R$ is a user-defined parameter.
Taking $p = 0$ gives the
  {\it unnormalized}  Laplacian while  $p = 1/2$ gives
the {\it normalized}  Laplacian. Other normalizations of $L$ are also
possible and can result in non-symmetric operators; see \cite[Sec.~5.1]{HHOS19} for a detailed discussion.

With the graph Laplacian matrix identified we finally
define the prior covariance matrix $C_\tau \in \R^{N \times N}$
with hyperparameters $\tau^2, \alpha > 0$ to be
\begin{equation}\label{C-tau-def}
C_\tau := \tau^{2\alpha}
( L + \tau^2 I_N)^{-\alpha}.
\end{equation}
Graph Laplacian operators are positive semi-definite (see \cite[Prop.~1]{von2007tutorial});
the matrix $C_\tau$ is therefore strictly positive definite thanks to the shift by $\tau^2 I_N$.
The normalization by $\tau^{2\alpha}$ ensures that the largest eigenvalue of $C_{\tau}$ is one, while $\alpha >0$ controls the rate of decay  of the rest of the eigenvalues of $C_\tau$; when the graph Laplacian is constructed from
nearly clustered data, $C_\tau$ will exhibit a spectral gap and the eigenvectors
associated with eigenvalues near one will contain geometric information about the
clusters; we refer to this phenomenon as the \emph{smoothing effect} of $C_\tau.$ In other words, the eigenvectors of $C_\tau$ that correspond to the eigenvalues near $1$ represent {\it smooth functions} with respect to the topology of the similarity graph $G$; therefore, our prior built around $C_\tau$ will favor functions (i.e. rows of $U$) that give similar outputs to datapoints $\bx_i, \bx_j$ that are ``closer'' in the similarity graph. We refer the reader to \cite{belkin2004regularization}
for further discussion about the smoothing assumptions inherent in using graph Laplacian matrices for SSL.

With $C_\tau$ at hand, we conclude our definition of the prior on the unknown $U$, 
the Gaussian measure $\mu_0(\dd U) = \mcl N(0, I_M \otimes C_\tau)$
with Lebesgue density 
\begin{equation}
  \label{mu-0-generic-form}
  \mu_0(\dd U) := \frac{1}{\big[ (2 \pi)^N {\rm det}(C_\tau)  \big]^{\frac{M}{2}}}
  \exp \left( - \frac{1}{2} \langle U^T, C_\tau^{-1} U^T \rangle_F \right)
    \dd U.
  \end{equation}
If we introduce
the rows $\{\bu_1, \cdots, \bu_M\}$ of $U$, then we note the
prior can be written as 
\begin{equation*}
  \begin{aligned}
    \mu_0(\dd U)  = \frac{1}{\big[ (2 \pi)^N {\rm det}(C_\tau)  \big]^{\frac{M}{2}}}
    \prod_{\ell=1}^M \exp\left(-\frac{1}{2}\left\langle \bu_\ell,
    C_{\tau}^{-1}\bu_\ell\right\rangle \right) \dd \bu_\ell.
  \end{aligned}
\end{equation*}
The above expression reveals that, a priori, each row of $U$ has
the same distribution, and is independent of the others,
and that this distribution on rows favours 
structure across $Z$ which reflects the eigenvectors of the
largest eigenvalues of $C_\tau.$ The matrix $C_\tau$ is chosen
so that this eigenstructure reflects clustering present in the
unlabeled data, for appropriately chosen $\tau$, determined
through the analysis in this paper.

\begin{remark}
The prior covariance $C_\tau$ defined in \eqref{C-tau-def}
depends on the unlabeled data $X$ through the matrix $L$ and the weight 
matrix $W$.  This perspective differs significantly
from standard BIPs, where the data only appears in the likelihood 
and the prior is constructed independent of the data
(other than, perhaps, a noise-dependent scaling)\cite{vdw2}. In our formulation of SSR, the labeled
  data appear in the likelihood potential $\Phi$ while the unlabeled data are used to
  construct the prior measure $\mu_0$.
\end{remark}

\subsection{The Posterior}\label{sec:posterior}
 Using Bayes' rule, 
 we can determine 
the posterior $\mu^Y$ from the likelihood $\PP(Y|U)$ 
and prior $\mu_0$ defined through the Radon-Nikodym derivative
\begin{equation}
  \label{bayes-rule}
  \frac{\dd \mu^Y}{\dd \mu_0}(U) = \frac{1}{\vartheta(Y)} \exp \Big( - \Phi( U; Y) \Big).
\end{equation}
The posterior measure $\mu^Y$ is the Gaussian defined by
\begin{align}\label{eq:posterior}
  \mu^Y(\dd U)
  = \frac{1}{ \vartheta(Y)} \exp\left(- \frac{1}{2\gamma^2} \Fnorm{U H^T - Y}^2
  - \frac{1}{2} \left\langle U^T, C_{\tau}^{-1}U^T\right\rangle_F
  \right) \dd U.
\end{align}

It is well-known that linear inverse problems with additive Gaussian noise and
a Gaussian prior result in Gaussian posteriors; this is due to the conjugacy of the prior and
the likelihood\cite{rasmussen2006gaussian}. In this case, 
we have the additional property that the independence
of the rows $\bu_\ell$ of $U$ under the prior $\mu_0$ is preserved under the
posterior $\mu^Y$. To see this, we introduce
the rows $\{\by_1, \cdots, \by_M\}$ of $Y$ and note that we may write 
\begin{dmath*}
  \mu^Y(\mathrm{d}U) \propto\exp \left[-\frac{1}{2}\sum_{m=1}^M
\frac{1}{\gamma^2}\left|H\bu_m-\by_m\right|^2 +   \left\langle \bu_m, C_{\tau}^{-1}\bu_m\right\rangle \right]. \\
\end{dmath*}
Using this structure as the product of i.i.d.~Gaussians in each
of the $M$ rows of $U$, Proposition~\ref{prop:posterior} shows that
 $\mu^Y = \mcl N( U^\ast, I \otimes C^\ast)$, where $U^\ast \in \mbb R^{M \times N}$ is
the matrix with rows
\begin{equation*}
  \pom_m = \frac{1}{\gamma^2}\poC H^T\by_m, \qquad m=1,\dots, M,
\end{equation*}
and  $C^\ast$ is the covariance matrix
\begin{equation*}
   \poC = \left(C_{\tau}^{-1} + \frac{1}{\gamma^2}H^T H \right)^{-1}.
\end{equation*}

\section{Consistency Of Bayesian SSR}\label{sec:main-contribution}

In this section, we prove consistency of the posterior
$\mu^Y$. We study consistency with respect to two small
parameters: $\gamma$, which measures noise in the the labeled data
$Y$, and $\epsilon$ which measures the closeness to perfectly clustered
unlabeled data $X$.  Recall from the Main Theorem
that our goal is to show that the measure of contraction $\mcl I$ (defined 
in \eqref{eq:contraction}) is controlled with the noise standard 
deviation $\gamma$ or the geometric perturbation parameter $\eps$,
whenever the prior hyperparameters  $\tau, \alpha$ are chosen appropriately.
We will show that letting $\gamma \to 0$
results in posterior contraction, until a floor is reached that is determined
by $\epsilon.$ Furthermore the analysis will reveal guidance about
the choice of the hyperparameters $\tau$ and $\alpha$ in the
prior. In Section~\ref{sec:disconnected-graph} we consider the
case of a disconnected graph with $\eps=0$ and obtain
contraction rates with respect to $\gamma.$ 
In Section~\ref{sec:weakly-conn-graph} we build on the  disconnected
case  to obtain our desired results 
for weakly connected graphs with $\epsilon$ small.

\subsection{Disconnected Graph}
\label{sec:disconnected-graph}

Consider a weighted graph $G_0 = \{ Z, W_0\}$ consisting of $K < N $ components (subgraphs) $\tG_k$. 
Without loss of generality, we assume that 
the vertices in $Z$ are ordered so that $Z = \{\tZ_1, \tZ_2, \cdots, \tZ_K\}$ with the 
$\tZ_k$ denoting the index set of vertices in subgraph $\tG_k$. We refer to $\tZ_k$ as
the clusters and let $\tN_k = |\tZ_k|$ denote the number of vertices in the $k$-th 
cluster. We make the following assumptions on the graph $G_0$.
\begin{assumption}
  \label{assumption:G0}
  The graph $G_0=\{Z,W_0\}$ satisfies the following conditions:
  \begin{itemize}
    \item[(a)] The weighted adjacency matrix $W_0 \in \mbb R^{N \times N}$ is block diagonal
          \[
            W_0 = \mathrm{diag}(\tW_1,\tW_2,\cdots, \tW_K),
          \]
          with $\tW_k \in \mbb R^{\tN_k \times \tN_k}$
          denoting the weight adjacency matrices of the subgraphs $\tG_k$.
    \item[(b)]  Let $\tL_k$ be the
          graph Laplacian matrices of the subgraphs $\tG_k$, i.e.,
          \begin{equation*}
            \tL_k:= \tD_k^{-p}(\tD_k - 
            \tW_k) \tD_k^{-p}
          \end{equation*}
          with $\tD_k$ denoting the degree matrix of $\tW_k$.
          There exists a uniform constant $\theta >0$ so that for $k=1,\cdots, K$ the
          submatrices $\tL_k$ satisfy
          \begin{equation}\label{uniform-spectral-gap-sub-matrices}
            \langle \bx, \tL_k \bv \rangle \ge \theta \langle \bv, \bx \rangle,
          \end{equation}
          for all vectors $\bv \in \mbb R^{\tN_k}$ and $\bv \bot \tD_k^{p} \mbf 1$
          with $\mbf 1\in \mbb R^{\tN_k}$ denoting the  vector of ones. In other words,
          the $\tL_k$ 
          have a uniform spectral gap.
  \end{itemize}
\end{assumption}
\begin{remark}
  The existence of such $\theta$ as in 
\eqref{uniform-spectral-gap-sub-matrices} is equivalent to assuming the subgraphs $\tG_k$ are pathwise connected; i.e. any two vertices in $\tG_k$ can be joined by a path within $\tG_k$. This is a direct consequence of 
\cite[Props.~2 and 4]{von2007tutorial} stating that the graph $\tG_k$ is connected  if and only if 0 is an 
  eigenvalue of $\tL_k$ with multiplicity 1 and that
the corresponding eigenvector is $\tD_k^p\mbf 1$.
\end{remark}

With a disconnected graph $G_0$ as above, we proceed as 
in Section~\ref{sec:prior-constr-using} and define
graph Laplacian and covariance matrices of the form
\begin{equation}\label{L-0-and-C-tau-0-def}
  L_0 := D_0^{-p} (D_0 - W_0) D_0^{-p} \quad \text{and}
  \quad C_{\tau, 0} := \tau^{2\alpha}( L_0  + \tau^2 I_N)^{- \alpha},
\end{equation}
with $D_0$ denoting the diagonal degree matrix of $W_0$ and parameters $\tau, \alpha >0$. We recall that because $L_0$ is symmetric and positive semidefinite, then the matrix power $(L_0 + \tau^2 I_N)^{-\alpha}$ for non-integer $\alpha > 0$ in \eqref{L-0-and-C-tau-0-def} is well-defined in terms of the eigendecomposition of $L_0$.
Note that
\[
            L_0 = \mathrm{diag}(\tL_1,\tL_2,\cdots, \tL_K),
          \]
and that $C_{\tau,0}$ inherits a similar block-diagonal structure.
We use the covariance matrix $C_{\tau, 0}$ to define prior measures $\mu_0$ of
the form \eqref{mu-0-generic-form}. In order to show posterior contraction with
such a prior, we also need to make some assumptions on
the index set of labeled data $Z'$ and the ground-truth matrix $U^\dagger$;
these encode the idea that the labels are coherent with the geometric
structure implied by the perfect clustering of the data.

\begin{assumption}\label{assumption:Zprime}
  At least one label is observed in each cluster $\tZ_k$; that is,
          \begin{equation*}
            |Z' \cap \tZ_k| \ > 0 \qquad \forall k =1, \dots, K.
          \end{equation*}
\end{assumption}

\begin{assumption}
  \label{assumption:U_dagger}
   Let $(\bu_m^\dagger)^T$
          for $m = 1, \dots, M$ denote the rows of $U^\dagger$.
          Then $\bu_m^\dagger \in {\rm span} \{ \bchi_1, \dots, \bchi_K\},$
          where the weighted set functions are defined by
          \begin{equation}
            \label{bchi-k-definition}
            \bchi_k :=  \frac{D_0^p \mbf 1_k}{\left|D_0^p \mbf 1_k \right|},
          \end{equation}
          with $\mbf 1_k \in \R^{N}$ denoting indicator of the cluster $\tZ_k$.
        \end{assumption}
        \begin{remark}
         Assumption~\ref{assumption:U_dagger} strongly enforces the notion that the 
          desired ground-truth functions $U^\dagger$ are consistent with the underlying perfect clustering structure of the $\{\tilde{Z}_k\}_{k=1}^K$. We do not expect a posterior consistency result without an assumption
          of this type and 
           note here that our current exposition does not address  posterior contraction   when Assumption~\ref{assumption:U_dagger} is violated.
            While this is an interesting and practically pertinent question, we delay it for future study. We conjecture that as long as the ground-truth variable $U^\dagger$ is consistent
            with the observed labeling and  the true underlying clustering structure of the unlabeled data $X$, then posterior contraction
            will occur around the projection of $U^\dagger$ onto ${\rm span} \{ \bchi_1, \dots, \bchi_K\}$.
     \end{remark}

With the above assumptions in hand we are ready to present our first
posterior contraction result in the case of disconnected graphs.
Recall $\mcl I(\gamma, \alpha, \tau)$ as defined in \eqref{eq:contraction}.

\begin{theorem}\label{thm:post-contraction-disc-graph}
Suppose that Assumptions~\ref{assumption:G0}, \ref{assumption:Zprime}
and \ref{assumption:U_dagger} are satisfied in turn by
the disconnected graph $G_0$, the labeled  set $Z'$ and
the ground-truth matrix $U^\dagger$. 
    Consider the label model \eqref{label-model}, the
    prior measure $\mu_0(\dd U) = \mcl N(0, C_{\tau,0})$ as
    in \eqref{mu-0-generic-form}, and  the resulting posterior measure $\mu^Y(\dd U)$ as
    in \eqref{eq:posterior}. Then there is a constant $\Xi > 0$ which is independent of hyperparameters $\gamma, \tau, $ and $\alpha$, so that
for every fixed $\gamma, \tau, \alpha> 0$, we have
    \begin{equation*}
      \mcl I( \gamma, \alpha, \tau) \le \Xi \max\left\{ \gamma^2,
      \tau^{2\alpha}\right\} \left(1 + \max\left\{ \gamma^2,
      \tau^{2\alpha}\right\}\|U^\dagger\|_F^2 \right).
      \end{equation*}
    \end{theorem}

We prove this theorem in Section~\ref{sec:proof-thm-disconnected}; 
here we discuss the intuition behind it.
If  $U \sim \mu_0$  as above then $\bu_m \iidsim \mcl N(0, C_{\tau,0})$
where we recall $(\bu_m)^T$ are the rows of $U$. Thus by
the Karhunen-Lo{\'e}ve (KL) theorem~\cite{loeve_1978},
\begin{equation*}
  \bu_m  \dequal  \sum_{j=1}^N \frac{1}{\sqrt{\lambda_{j,0}}} \xi_{mj} \bphi_{j,0},
\end{equation*}
with $\{ (\lambda_{j,0}, \bphi_{j,0})\}_{j=1}^N$ denoting the eigenpairs of $C_{\tau, 0}$ and
$\xi_{mj} \iidsim \mcl N(0,1)$.
The matrix $L_0$ has a $K$ dimensional null-space
spanned by the $\bchi_k$ and this null-space is associated to the
eigenvalue $1$ for $C_{\tau,0}.$ Furthermore, when $\tau^2$ is small
the remaining eigenvalues of $C_{\tau,0}$ are also small. These
ideas are made rigorous in \cite[Lemma~35 and Prop.~36]{HHRS19}.
From those results it follows that
\begin{equation}\label{KL-expansion-disconnected-graph}
  \bu_m \dequal  \sum_{j=1}^K  \xi_{mj} \bchi_j + \mcl O(\tau^{2\alpha}),
\end{equation}
meaning that the prior is concentrated on $\text{span} \{\bchi_1, \dots, \bchi_K\}$. On the other hand
the posterior $\mu^Y$ also decouples along the rows $\bu_m$ following Proposition~\ref{prop:posterior}
and so the SSR problem can be viewed as $M$ separate BIPs for each row
of $\bu_m$, all with the same structure.
As $\tau \to 0$ the prior mass concentrates on the $K$ dimensional subspace spanned
by the set-functions $\bchi_k$. Since the posterior is absolutely continuous with respect
to the prior, the posterior mass will also concentrate on the same subspace. 
The assumptions on the ground truth $U^\dagger$ ensure that the data
is consistent with the rows $\bu_m$ lying in this subspace and give information
on assignation of labels, corresponding to weights on the $\bchi_m.$ Hence, 
letting $\gamma \to 0$ yields concentration of the posterior
around the ground-truth matrix $U^\dagger$ under
Assumptions~\ref{assumption:Zprime} and \ref{assumption:U_dagger}.

\begin{remark}
\label{r:withoute}
Theorem~\ref{thm:post-contraction-disc-graph} suggests that, in this perfectly clustered setting, choosing
$\tau$ to achieve $\tau^{2\alpha} = \gamma^2$ is optimal, since
it balances the two sources of error in the contraction rate.
However, in the next subsection, we study the case that the unlabeled data is not perfectly clustered, where we measure the
proximity of it to being perfectly clustered with the parameter $\epsilon$.
We state our theorems in a setting in which $\tau$ scales as
a power of $\epsilon$, rather than $\gamma$. We make this choice because
$\tau$ and $\epsilon$ are linked intrinsically
through the unsupervised learning task encapsulated in the
prior measure, based on the unlabeled data,
whilst $\gamma$ enters separately through the likelihood, which
captures the labeled data.
In a broader picture, these considerations about the choice of $\tau$
suggest the importance of choosing 
this hyperparameter in a data-adaptive fashion and the importance 
of using hierarchical Bayesian methods to learn such parameters. 
\end{remark}

\subsubsection{Proof of Theorem~\ref{thm:post-contraction-disc-graph}}\label{sec:proof-thm-disconnected}

We first bound the inner expectation in \eqref{eq:contraction}, which is the mean square error of the estimator $U|Y$. We define the matrix $\poC_0$  to be the the posterior covariance obtained by substituting  the
prior covariance $C_{\tau,0}$ from \eqref{L-0-and-C-tau-0-def} into (\ref{eq:mean and cov}), i.e., 
\begin{equation}\label{eq:C-0-ast-definition}
\poC_0 := \left( C^{-1}_{\tau,0} + \frac{1}{\gamma^2} B \right)^{-1}.
\end{equation}
For brevity we suppress the dependence of $\poC_0$ on $\tau, \alpha,$ 
and $\gamma$. We then have
\[
 \mbb{E}_{U|Y} \| U - U^\dagger \|_F^2 
   = 	\sum_{m=1}^M \mathbb{E}_{\bu_m | \by_m}
  \left|\bu_m-\bu^\dagger_m \right|^2         
   =	M \text{Tr}(\poC_0) + \sum_{m=1}^M
  \left|
  \frac{1}{\gamma^2}\poC_0 H^T\by_m - \bu^\dagger_m
  \right|^2.
\]
The first identity relies on the independence of the rows $\bu_m^T$ of
$U$ under the posterior distribution, as established in 
Proposition~\ref{prop:posterior}. 
The second identity comes from the fact that the mean square error is the sum of the variance and squared bias of the estimator of each row.

We may now apply the outer expectation in definition of $\mcl I$ with respect to the
data $Y|U^\dagger$, and since $\mathrm{Tr}(\poC_0)$ does not depend on $Y$, we may pull it out of the outer expectation and write
\begin{equation}\label{eq:I-bias-variance-decomp-1}
  \mcl I(\gamma, \alpha, \tau)  =
  M \text{Tr}(\poC_0) + \mathbb{E}_{Y|U^\dagger} \left( \sum_{m=1}^M 
  \left|\frac{1}{\gamma^2}\poC_0 H^T\by_m
  - \bu^\dagger_m \right|^2 \right).
\end{equation}
Since we assumed
\begin{equation}\label{eq:noise-model-repeat}
\by_m | \bu^{\dagger}_m \sim \mathcal{N}(H\bu^\dagger_m, \gamma^2I_{J})
\end{equation}
and  the rows $\left\{\by_m^T \right\}_{m=1}^M$ are independent conditional on $U^\dagger$,
we can write
\[
  \mathbb{E}_{Y|U^\dagger}
  \left|\frac{1}{\gamma^2}\poC_0 H^T\by_m
  - \bu^\dagger_m \right|^2
  =  \mathbb{E}_{\by_m | \bu^\dagger_m}
  \left|\frac{1}{\gamma^2}\poC_0 H^T\by_m
  - \bu^\dagger_m\right|^2.
\]
This expectation is the mean square error of the posterior mean estimator of $\bu^\dagger_m$, which  can
be decomposed into a 
 variance and a squared
bias term:
\begin{align*}
  \mathbb{E}_{\by_m|\bu^\dagger_m} \left|\frac{1}{\gamma^2}\poC_0 H^T\by_m
  -\bu^\dagger_m\right|^2
  = \text{Tr}&\left(
  \text{Cov}\left(\frac{1}{\gamma^2}\poC_0 H^T\by_m\right)\right)
  + \\
  &\left| \mbb{E}_{\by_m|\bu^\dagger_m}
  \left(\frac{1}{\gamma^2}\poC_0 H^T\by_m\right)
  - \bu^\dagger_m \right|^2,
\end{align*}
where $\mathrm{Cov}(\cdot)$ denotes the covariance
matrix of a random vector. We compute the variance term
using \eqref{eq:noise-model-repeat}:
\[
  \text{Cov}\left(\frac{1}{\gamma^2}\poC_0 H^T\by_m\right)
  = \frac{1}{\gamma^2}\poC_0 H^T \text{Cov}\left(\by_m\right)
  \frac{1}{\gamma^2}H(\poC_0)^T
  = \frac{1}{\gamma^2} \poC_0 B\poC_0,
\]
where we used the fact that $\mathrm{Cov}(\by_m) = \gamma^2I_{J}$ and
$B = H^TH \in\mathbb{R}^{N\times N}$. For the bias term, we can write
\[
  \mathbb{E}_{\by_m|\bu^\dagger_m}
  \left( \frac{1}{\gamma^2}\poC_0 H^T\by_m\right)
  = \frac{1}{\gamma^2} \poC_0 H^TH\bu^\dagger_m
  = \frac{1}{\gamma^2} \poC_0 B\bu^\dagger_m.
\]
Putting these terms  together yields
\[
  \mathbb{E}_{\by_m|\bu^\dagger_m}
  \left|\frac{1}{\gamma^2}\poC_0 H^T\by_m - \bu^\dagger_m\right|^2
  = \frac{1}{\gamma^2}\text{Tr}\left(\poC_0 B\poC_0\right)
  + \left|\frac{1}{\gamma^2}\poC_0 B\bu^\dagger_m-\bu^\dagger_m \right|^2.
\]
Substituting this identity back into \eqref{eq:I-bias-variance-decomp-1} yields
\begin{equation}
  \mcl I(\gamma, \alpha, \tau)
  = M \text{Tr}(\poC_0) + \frac{M}{\gamma^2}\text{Tr}(\poC_0 B \poC_0)
  + \sum_{m=1}^M
  \left|\frac{1}{\gamma^2}\poC_0 B\bu^\dagger_m - \bu^\dagger_m \right|^2.
  \label{eq:three terms_1}
\end{equation}
The desired bound now follows from Lemmata \ref{lemma:1-disc},
\ref{lemma:2-disc}, and \ref{lemma:3-disc} below that in turn bound the first, second, and
third term in
the right hand side of  \eqref{eq:three terms_1}. These Lemmata are proved in Appendix~\ref{ssec:A2}. We note that the constant $\Xi$ is common to each of the results and that the results indeed hold for any $\gamma, \tau, \alpha > 0$.

\begin{lemma}\label{lemma:1-disc}
Suppose Assumptions~\ref{assumption:G0} and \ref{assumption:Zprime} are
 satisfied by the disconnected graph $G_0$ and the labeled set $Z'$, respectively. 
Then  there exists a constant $\Xi>0$,
such that for any $\gamma, \tau, \alpha > 0$, we have
  %
  %
\begin{equation}\label{eq:trace-bound-C-0-ast}
     {\rm Tr}(\poC_0) \leq \Xi \max\{ \gamma^2, \tau^{2\alpha}\},
   \end{equation}
  where $\poC_0$  is
  the posterior covariance matrix in \eqref{eq:C-0-ast-definition}.
  %
  %
\end{lemma}

\begin{lemma}\label{lemma:2-disc}
  Suppose Lemma~\ref{lemma:1-disc} is satisfied. Then for  any $\gamma, \tau, \alpha > 0$, we have
  \[
    \frac{1}{\gamma^2} \mathrm{Tr}(\poC_0 B\poC_0) \le \mathrm{Tr}(\poC_0) \leq \Xi \max\left\{ \gamma^2, \tau^{2\alpha} \right\},
  \]
  with the same constant $\Xi> 0$   as in \eqref{eq:trace-bound-C-0-ast}.
\end{lemma}

\begin{lemma}\label{lemma:3-disc}
Suppose Assumptions~\ref{assumption:G0},  \ref{assumption:Zprime}, and \ref{assumption:U_dagger} are
   in turn satisfied by the disconnected graph $G_0$, 
  the labeled set $Z'$, and the ground-truth function $U^\dagger$. Then for any $\gamma, \tau, \alpha >0$ and $m =1, \dots, M$, we have
  \[
    \left| \frac{1}{\gamma^2}\poC_0 B\bu_m^\dagger - \bu_m^\dagger\right|
    \le \mathrm{Tr}(\poC_0) \le  \Xi \max\{\gamma^2, \tau^{2\alpha}\},
  \]
  where $\Xi >0$ is the same constant as in \eqref{eq:trace-bound-C-0-ast}.
\end{lemma}

\subsection{Weakly Connected Graph}\label{sec:weakly-conn-graph}

We now consider a generalization of the setting in the previous
subsection, in which
the disconnected graph $G_0=\{Z, W_0\}$ is perturbed, and the
perturbation results in a  graph $G_\epsilon = \{Z, W_{\epsilon}\}$. 
Similar to \cite{HHRS19} we collect the following set of assumptions 
on this perturbed graph $G_{\epsilon}$.

\begin{assumption}
  \label{assumption:G_k}
  The graph $G_\eps= \{Z, W_\eps\}$ satisfies the following three conditions.
  \begin{enumerate}[(a)]
    \item The weighted adjacency matrix $W_\eps$ can be expanded in the form
          \begin{equation}\label{W-eps-expansion}
            W_{\epsilon} = W_0 + \sum_{h=1}^\infty \epsilon^h W^{(h)},
          \end{equation}
          where $W_0$ is the weighted adjacency matrix of a
          disconnected graph $G_0$.
    \item The  matrices $W^{(h)}$ are self-adjoint and
          $ \{\| W^{(h)} \|_2\}_{h =1}^\infty \in \ell^\infty$.
    \item Let $w^{(0)}_{ij}$ and $w^{(h)}_{ij}$ denote the entries of $W_0$ and
          $W^{(h)}$ respectively. Then, for $h \ge 1$, we assume
          \begin{equation}
            \label{W-k-condition}
            \left\{
            \begin{aligned}
               & w^{(h)}_{ij} \ge 0,\quad \text{if} \quad w_{ij}^{(0)} = 0 \quad
              \text{for} \quad i,j \in Z, i \neq j \\
              & w^{(h)}_{ii} = 0.
            \end{aligned}
            \right.
          \end{equation}
  \end{enumerate}
\end{assumption}

The assumptions (b) and (c) above ensure that  $W_\eps$ is a well-defined adjacency matrix.
Also note that (c) allows for $w^{(h)}_{ij}$, $h \ge 1$, 
to be negative whenever $w^{(0)}_{ij} >0$.
Even in this case, $W_\eps$ will be non-negative only as long as $\eps > 0$ is sufficiently small. With the above assumptions identified we can proceed analogously to Section~\ref{sec:prior-constr-using}
to define Laplacian and covariance matrices
\begin{equation}\label{L-eps-and-C-tau-eps-def}
  L_\eps := D_\eps^{-p} (D_\eps - W_\eps) D_\eps^{-p} \quad \text{and}
  \quad C_{\tau, \eps} := \tau^{2\alpha}( L_\eps  + \tau^2 I_N)^{- \alpha},
\end{equation}
with $D_\eps$ denoting the diagonal degree matrix of $W_\eps$ and parameters $\tau, \alpha >0$.
We then use the covariance matrix $C_{\tau, \eps}$ to define a prior 
measure $\mu_0$ of
the form~\eqref{mu-0-generic-form}  on the weakly connected graph $G_\eps$.
With the assumptions made about the disconnected set-up
in Subsection~\ref{sec:disconnected-graph} and the
above new assumptions  on the weakly connected set-up, we can now present our 
main posterior contraction result, the analogue of 
Theorem~\ref{thm:post-contraction-disc-graph}, for weakly connected 
graphs $G_\eps$.

\begin{theorem}\label{thm:main}
    Suppose Assumptions~\ref{assumption:G0}, \ref{assumption:Zprime}, 
\ref{assumption:U_dagger} and \ref{assumption:G_k} are satisfied in turn 
by the disconnected graph $G_0$, the labeled set $Z'$, the ground-truth 
matrix $U^\dagger$ and the weakly connected graph  $G_\eps$. 
Fix $\alpha>0$. Then there exist constants $(\epsilon_0, \Xi_0, \Xi_1) \in (0,1) \times (0, +\infty)^2$  
such that whenever $\eps < \eps_0$ then
  \begin{equation*}
    \begin{aligned}
      \mcl I(\gamma, \alpha, \tau, \eps) & \le \Xi \max\left\{ \gamma^2,  \left( \frac{\tau^{2}}{1 - \Xi_1 \eps/\tau^2} \right)^\alpha\right\} \\
      & \times \left( 1 + \max\left\{ \gamma^2,  \left( \frac{\tau^{2}}{1 - \Xi_1 \eps/\tau^2} \right)^\alpha\right\} \left[ \eps + \frac{\eps}{\tau^{2\alpha}} + \left( 1 + \frac{\eps}{\tau^2} \right)^\alpha  \right]^2
    \|U^\dagger\|^2  \right).
\end{aligned}
  \end{equation*}
\end{theorem}

The 
intuition behind the proof is that we use the same ideas which underlie
Theorem~\ref{thm:post-contraction-disc-graph}, which
concerns the case $\eps=0$, coupled with new arguments which control 
perturbations to the spectrum of $C_{\tau,\eps}$ with respect to that
of $C_{\tau,0}.$ Specifically $C_{\tau,\eps}$ now has a one-dimensional
null-space associated with the eigenvalue $1$, but has an additional 
$K-1$ eigenvalues of size $1 - \mcl{O}(\eps/\tau^2)$. The remaining eigenvalues
are small, of $\mcl{O}(\tau^{2\alpha})$, if an appropriate relationship 
between $\eps$ and $\tau$ is imposed. The eigenfunctions associated with 
the $K$ eigenvalues at, or near, $1$, nearly span the same space as the N
weighted set-functions $\{\bchi_k\}_{k=1}^K$. Let $(\bu_m)^T$ denote
the rows of $U \sim \mu_0$. Then it
follows from \cite[Prop.~41]{HHRS19} that 
these rows concentrate on the span of the $\bchi_k$
with errors of the form
$\mcl O \left( \eps^2 \tau^{-4} + \tau^{4\alpha} + \eps^2\right)$
 when $\eps = o(\tau^2)$ 
and of the form
$\mcl O \left( \tau^{4\alpha} + \eps^2\right)$
when $\eps = \Theta(\tau^2)$ and $\tau^2$ is small.
These approximation results for the rows $(\bu_m)^T$ under
the prior underlie the proof.  The rest of the argument follows
in the footsteps of Theorem~\ref{thm:post-contraction-disc-graph}. 
First, we decouple the posterior on the rows of $U$ using 
Proposition~\ref{posterior-product-form} to obtain $M$ independent 
BIPs. In each BIP the prior concentration on the span of $\bchi_k$
results in  posterior concentration along the same subspace, at which 
point, the noise standard deviation $\gamma$ in the likelihood 
potential $\Phi$ controls the contraction of the posterior around 
the ground-truth matrix $U^\dagger$ under 
Assumptions~\ref{assumption:Zprime} and \ref{assumption:U_dagger}.

\begin{remark}
    We briefly note that our bounds on the measure of posterior contraction $\mcl I$ are dependent on the number of vertices $N$ in the graph, which for large graphs could be considered worrisome. As $N$ grows, the eigenvalues $L_\eps$ grow as well, and so in practice one could apply  $1/N^s$ scaling for an appropriate power $s>0$ on the graph Laplacian in order to control such growth. 
    Such scalings ensure the convergence of the graph Laplacian to families of integral or differential 
    operators in the $N \rightarrow \infty$ limit; see for example \cite{HHOS19, trillos2016variational, trillos2018error}.
    It is interesting, and relevant, to study Bayesian posterior consistency in this continuum limit as a future direction.
\end{remark}

\subsubsection{Proof of Theorem~\ref{thm:main}}\label{sec:proof-thm-weakly-connected}

Let us define the perturbed posterior covariance matrix
\begin{equation}\label{eq:C-eps-ast-definition}
\poC_\eps := \left( C^{-1}_{\tau,\eps} + \frac{1}{\gamma^2} B \right)^{-1},
\end{equation}
following \eqref{L-eps-and-C-tau-eps-def}
with the prior covariance matrix $C_{\tau, \eps}$.
  Observe that the arguments leading up to the upper bound \eqref{sec:proof-thm-disconnected}
  hold with $\poC_0$ replaced with $\poC_\eps$. Thus we immediately obtain the identity
\begin{equation}
  \mcl I (\gamma, \alpha, \tau, \epsilon)
  = M \text{Tr}(\poC_\eps) + \frac{M}{\gamma^2}\text{Tr}(\poC_\eps B \poC_\eps)
  + \sum_{m=1}^M
  \left|\frac{1}{\gamma^2}\poC_\eps B\bu^\dagger_m - \bu^\dagger_m \right|^2.
  \label{eq:three-terms-eps}
\end{equation}

Similarly to Section~\ref{sec:proof-thm-disconnected} we prove Theorem~\ref{thm:main}
by bounding each term in the right hand side of \eqref{eq:three-terms-eps} in the Lemmata
 \ref{lemma:1},
\ref{lemma:2}, and \ref{lemma:3} below. The proofs are collected in Appendix~\ref{ssec:A3}.

\begin{lemma}\label{lemma:1}
Suppose Assumptions~\ref{assumption:G0}, \ref{assumption:Zprime}, and \ref{assumption:G_k} 
are satisfied in turn by the disconnected graph $G_0$, the labeled set $Z'$, and 
the weakly connected graph $G_\eps$. 
Fix $\alpha>0$. 
Then there exist constants $(\epsilon_0, \Xi_0, \Xi_1)  \in (0,1) \times (0, + \infty)^2$
  such that whenever $\eps < \eps_0$ then 
\[
    {\rm Tr}(\poC_\eps)
    \leq \Xi_0 \max \left\{ \gamma^2, \left( \frac{ \tau^{2}}{ 1- \Xi_1 \eps/ \tau^{2} } \right)^\alpha\right\},
  \]
  with $C^\ast_\eps$ as in \eqref{eq:C-eps-ast-definition}.
\end{lemma}

\begin{lemma}\label{lemma:2}
  Suppose that the conditions of Lemma~\ref{lemma:1} are satisfied and
fix $\alpha>0$. Then there exist constants $(\epsilon_0, \Xi_0, \Xi_1) \in (0,1) \times (0, +\infty)^2$ (the 
same constants as in Lemma~\ref{lemma:1})
such that whenever $\eps < \eps_0$ then
  \begin{equation*}
    \frac{1}{\gamma^2} \mathrm{Tr}(\poC_\eps B \poC_\eps)
\leq \Xi_0 \max \left\{ \gamma^2, \left( \frac{ \tau^{2}}{ 1- \Xi_1 \eps/ \tau^{2} } \right)^\alpha\right\}.
\end{equation*}
\end{lemma}

\begin{lemma}\label{lemma:3}
Suppose Assumptions~\ref{assumption:G0}, \ref{assumption:Zprime}, 
\ref{assumption:U_dagger}, and \ref{assumption:G_k} are satisfied 
by the disconnected graph $G_0$, the labeled set $Z'$, 
the ground-truth matrix $U^\dagger$ and the weakly
    connected graph $G_\eps$ respectively and
fix $\alpha>0$. Then there exist constants $(\epsilon_0, \Xi_0, \Xi_1) \in (0,1) \times (0, +\infty)^2$
such that whenever $\eps < \eps_0$ then 
    \begin{equation*}
    \left|\frac{1}{\gamma^2} \poC_\eps B\bu^\dagger_m - \bu^\dagger_m\right|
    \le \Xi_2 \max\left\{ \gamma^2,  \left( \frac{\tau^{2}}{1 - \Xi_1 \eps/\tau^2} \right)^\alpha\right\}
    \left[ \eps + \frac{\eps}{\tau^{2\alpha}} + \left( 1 + \frac{\eps}{\tau^2} \right)^\alpha  \right]
    |\bu^\dagger_m|.
    \end{equation*}
\end{lemma}


We now present a corollary of Theorem~\ref{thm:main} that is
the precisely stated version of our informal Main Theorem from
Section \ref{sec:intro}.

\begin{corollary}\label{cor:main-a}
  Suppose that the conditions of Theorem~\ref{thm:main} are satisfied and 
that for a fixed $\alpha >0$, the hyperparameters
  $(\eps, \tau)$ are chosen to satisfy
  \begin{equation*}
    2 \Xi_1 \eps =\tau^{\max\{2, 2\alpha\}}.
  \end{equation*}
Then there exists $\Xi_2 >0$ depending on $\alpha$ and the
constants $\Xi, \Xi_1$ from Theorem~\ref{thm:main} but independent of $\eps$ and $\gamma$, so that
   \begin{equation*}
     \mcl I \le \Xi_2 \max \left\{ \gamma^2, \eps^{\min\{1, \alpha\}} \right\}.
   \end{equation*}
\end{corollary}

\begin{remark}
The reader is encouraged to study the discussion following
the informal Main Theorem for an interpretation
of this result in terms of asymptotic consistency. We also note that 
an application of Markov's inequality can immediately extend the bound
in Corollary~\ref{cor:main-a} to a bound on the expected probabilities
of posterior samples being found far from the ground truth $U^\dagger$. 
More precisely, for any $\delta > 0$, we have
  \begin{equation*}
      \mathbb{E}_{Y|U^\dagger}\Bigl\{\mu^Y\left(\Fnorm{U-U^\dagger}> \delta\right)\Bigr\}
    \le \frac{\mcl I}{ \delta^2}.
  \end{equation*}
\end{remark}

\section{Numerical Experiments}\label{sec:numerical-experiments}

In this section, we provide numerical experiments that elucidate 
our main theoretical results and in particular examine the convergence 
rate of the contraction functional $\mcl I$ with respect to both 
the $\epsilon$ and $\gamma$ parameters. We use a
synthetic example in Subsection~\ref{sec:synthetic-data}
as well as the MNIST database of handwritten 
digits \cite{lecun2010mnist} in Subsection~\ref{sec:mnist-data}. 
In both examples, we compute $\mathcal{I}$ via the decomposition given in \eqref{eq:three terms_1},
which provides us with an explicit formula to numerically compute the contraction measure.
We then vary $\epsilon$ and $\gamma$ parameters while choosing $\tau = \epsilon^{1/\max\{2, 2\alpha\}}$.
We numerically differentiate $\log(\mathcal{I})$ with respect to $\log(\epsilon)$ and $\log(\gamma)$ to estimate the rate of convergence with respect to these two parameters.
A surface plot of these derivatives is then presented in  
Figures~\ref{fig:slope} and \ref{fig:MNIST slope}, for the two respective datasets, in which  the color encodes the estimated rate of convergence in terms of the respective variables.
The dark blue colors in these plots indicate a rate of convergence  of $\mcl I$ that is close to zero, meaning that convergence has approximately ceased,
while bright yellow colors indicate larger convergence rates of $\mcl I$.
Further numerical results are presented in Subsection~\ref{sec:numer-supp-lemm-1} in the supplemental material,
taking a closer look at the rates of convergence of different bias and variance terms that contribute to $\mcl I$.

\subsection{Synthetic Data}\label{sec:synthetic-data}

We construct a synthetic weakly connected graph consisting of three clusters of $100$ vertices each, where each cluster represents a different class.
We obtain the weight matrix $W_\epsilon$ following \eqref{W-eps-expansion}; we truncate the expansion at the $\epsilon^3$ level.
Each entry of weight matrices $W_0$ and $W^{(h)},\,h=1,2,3$ are drawn independently from a uniform distribution on $[0,1]$.
The matrices $W_0$ and $W^{(h)},\,h =1,2,3$ are fixed once sampled and are used to construct $W_\epsilon$ for different $\epsilon$ values.
Each $W_\epsilon$ is then symmetrized via the transformation $W_\eps \mapsto (W_\epsilon + W_\epsilon^T)/2$. 
We pick one vertex from each cluster to be labeled and choose ground truth $U^\dagger = [\bchi_1, \bchi_2, \bchi_3]^T$.
We vary $\epsilon$ values from $10^{-1}$ to $10^{-15}$ and $\gamma$ ranging from $10^{-1}$ to $10^{-7.5}$; $\tau$ is taken to be $\epsilon^{1/\max\{2, 2\alpha\}}$.

In Figure~\ref{fig:I vs gamma}, we demonstrate the convergence of $\mathcal{I}$ in the limit of the noise standard deviation
  $\gamma$ going to zero, for different values of $\alpha$ and $\epsilon$. 
We see posterior contraction with respect to $\gamma$
until a floor is reached; this floor depends on $\epsilon$, the degree of
clustering in the data, and is smaller for smaller $\epsilon.$

\begin{figure}
\subfloat[$\alpha = 0.5$]{\begin{overpic}[width=0.32\textwidth]{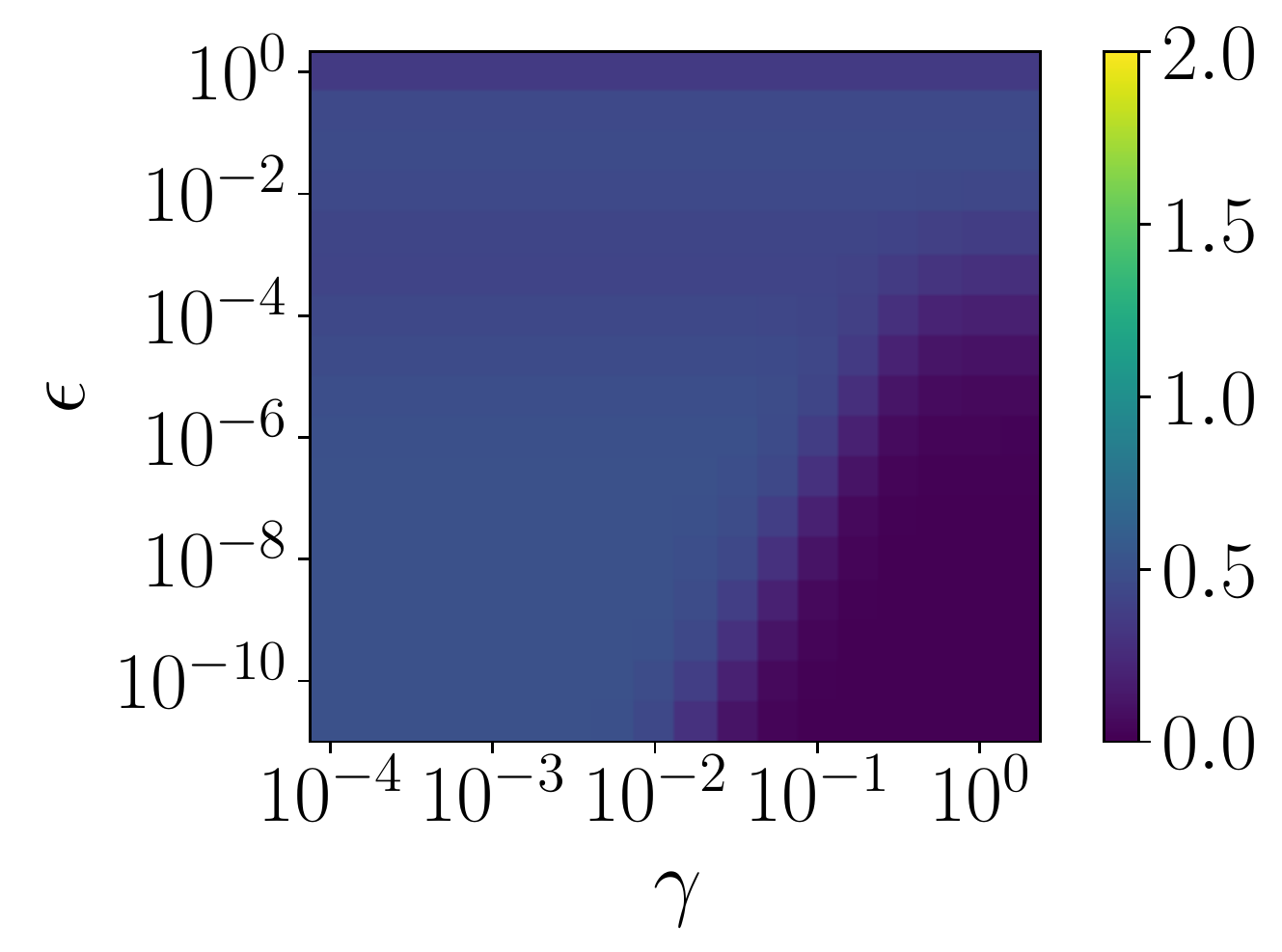}
\put(30,50){\color{white}$c_\epsilon = 0.5$}
\end{overpic}
}
\subfloat[$\alpha = 1$]{\begin{overpic}[width=0.32\textwidth]{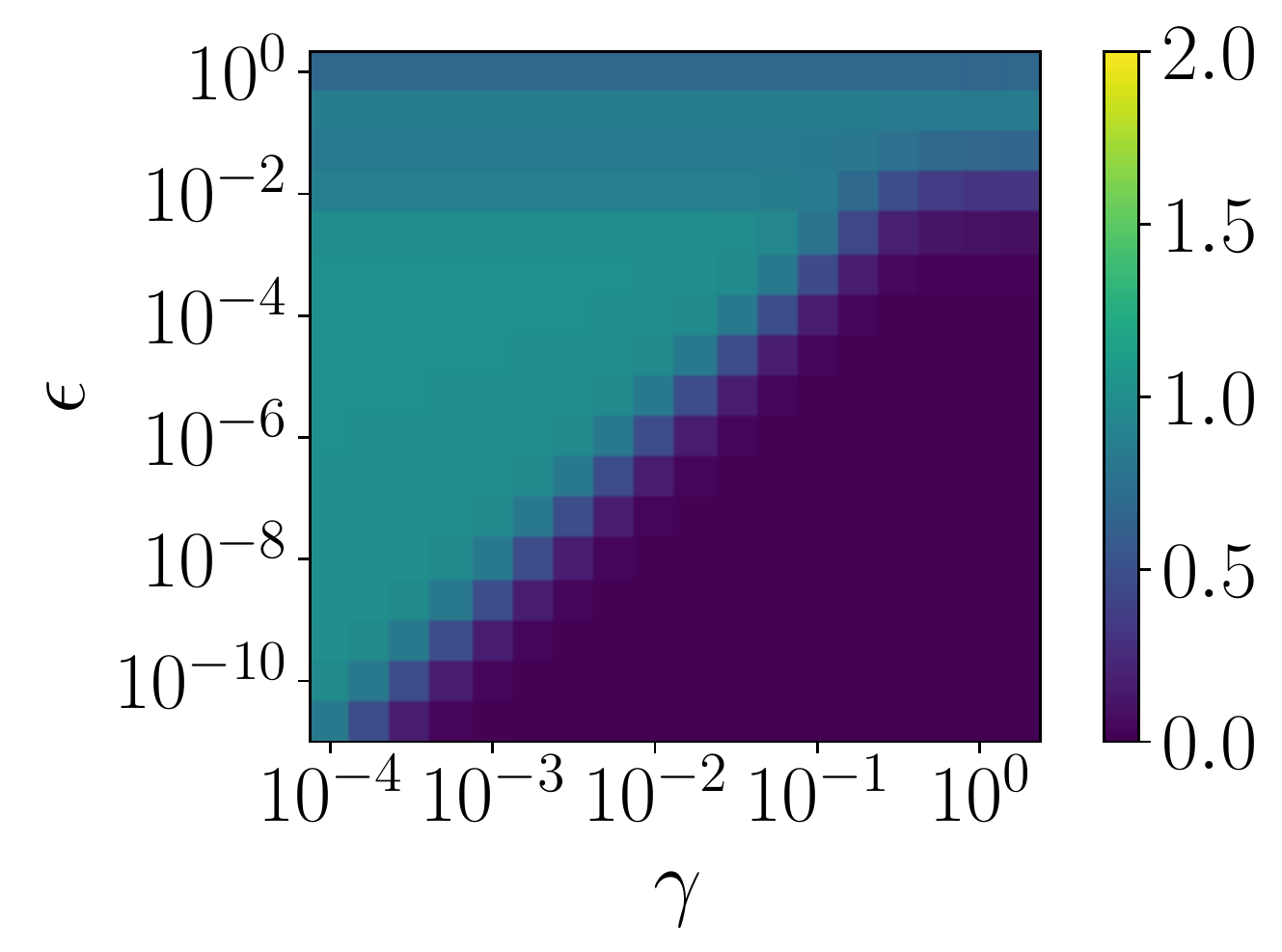}
    \put(30,50){$c_\epsilon = 1$}
\end{overpic}
}
\subfloat[$\alpha = 5$]{\begin{overpic}[width=0.32\textwidth]{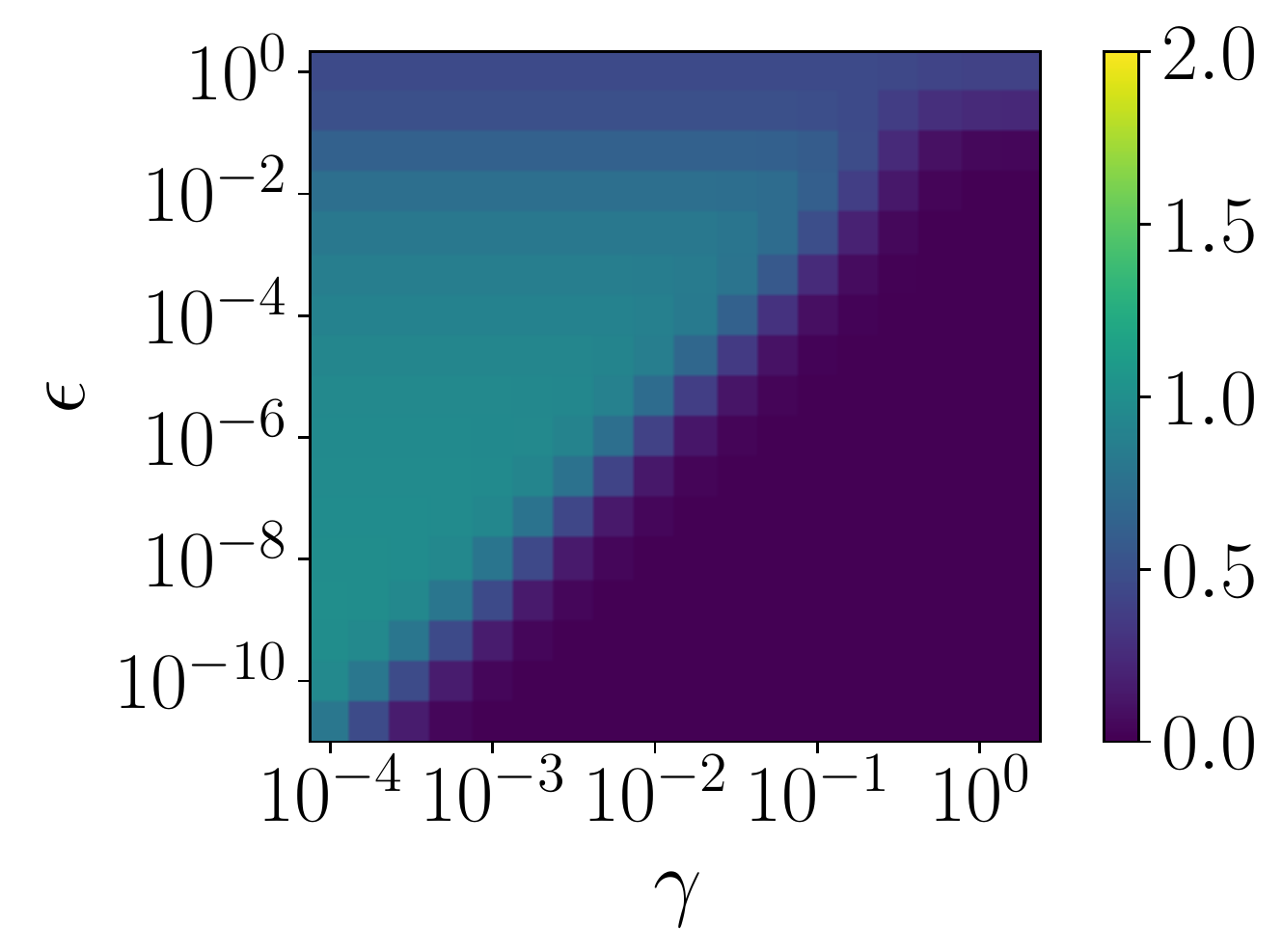}
\put(30,50){$c_\epsilon = 1$}
\end{overpic}}\\
\subfloat[$\alpha = 0.5$]{\begin{overpic}[width=0.32\textwidth]{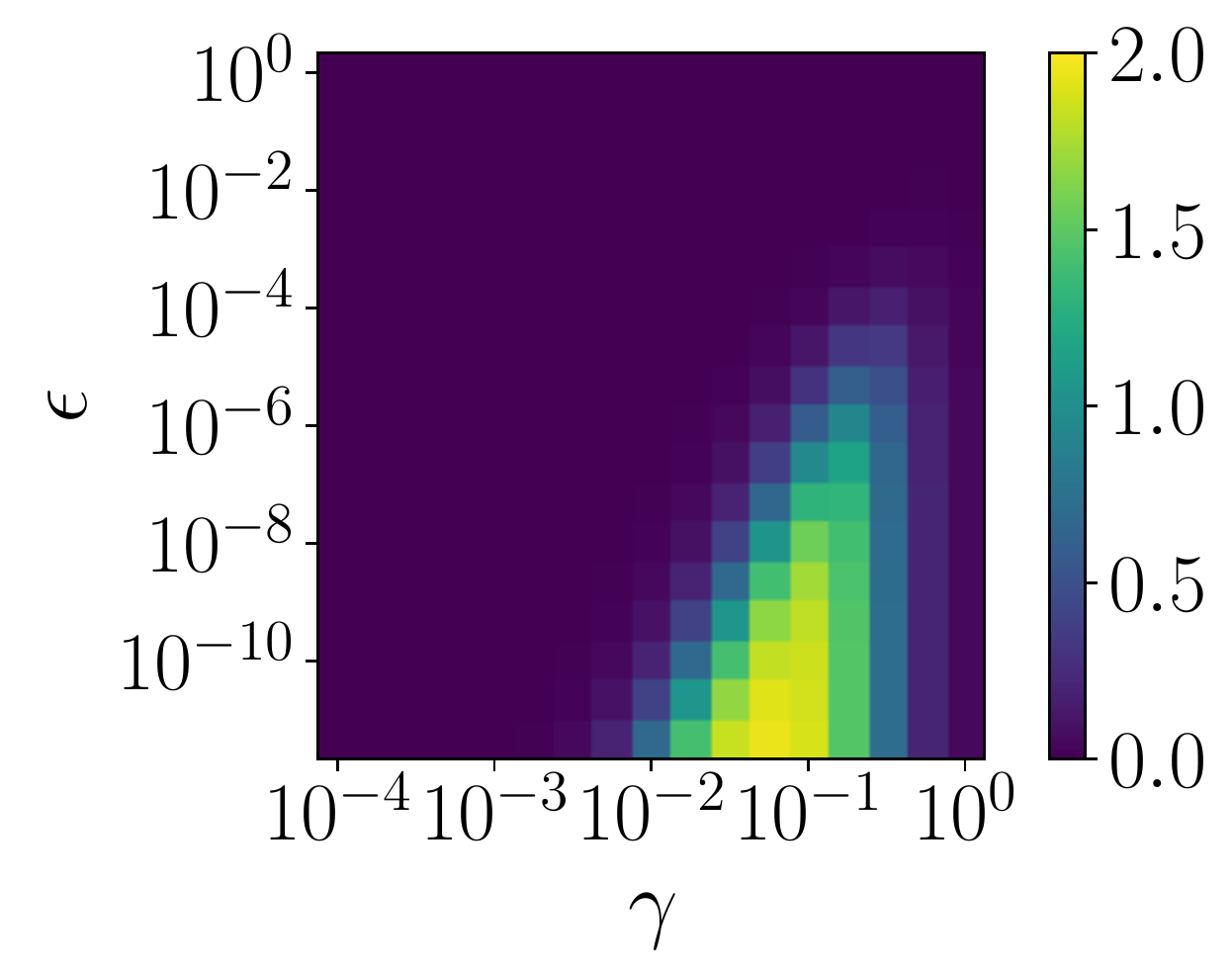}
\put(50,25){\color{white}$c_\gamma = 2$}
\end{overpic}
}
\subfloat[$\alpha = 1$]{\begin{overpic}[width=0.32\textwidth]{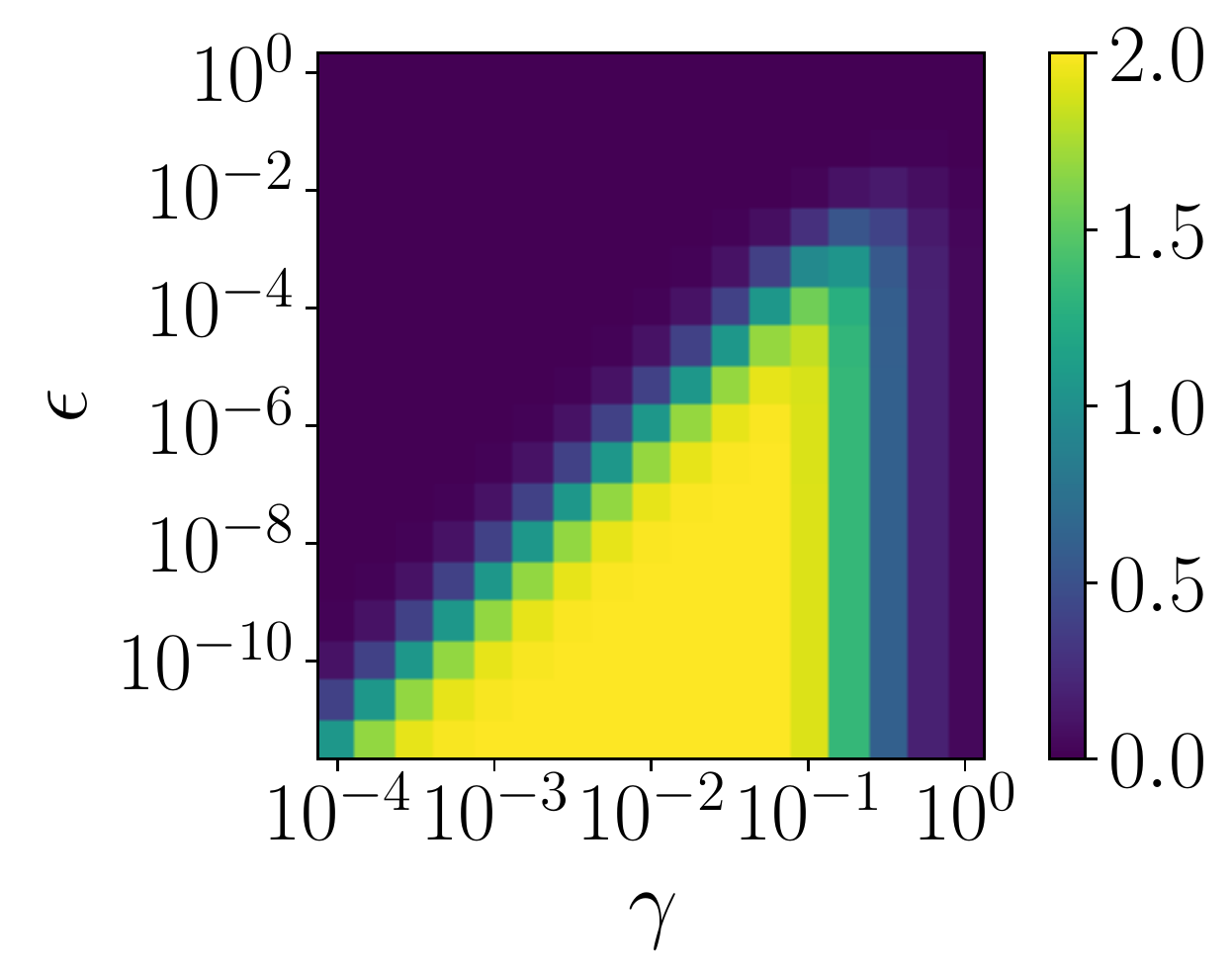}
\put(42,25){$c_\gamma = 2$}
\end{overpic}
}
\subfloat[$\alpha = 5$]{\begin{overpic}[width=0.32\textwidth]{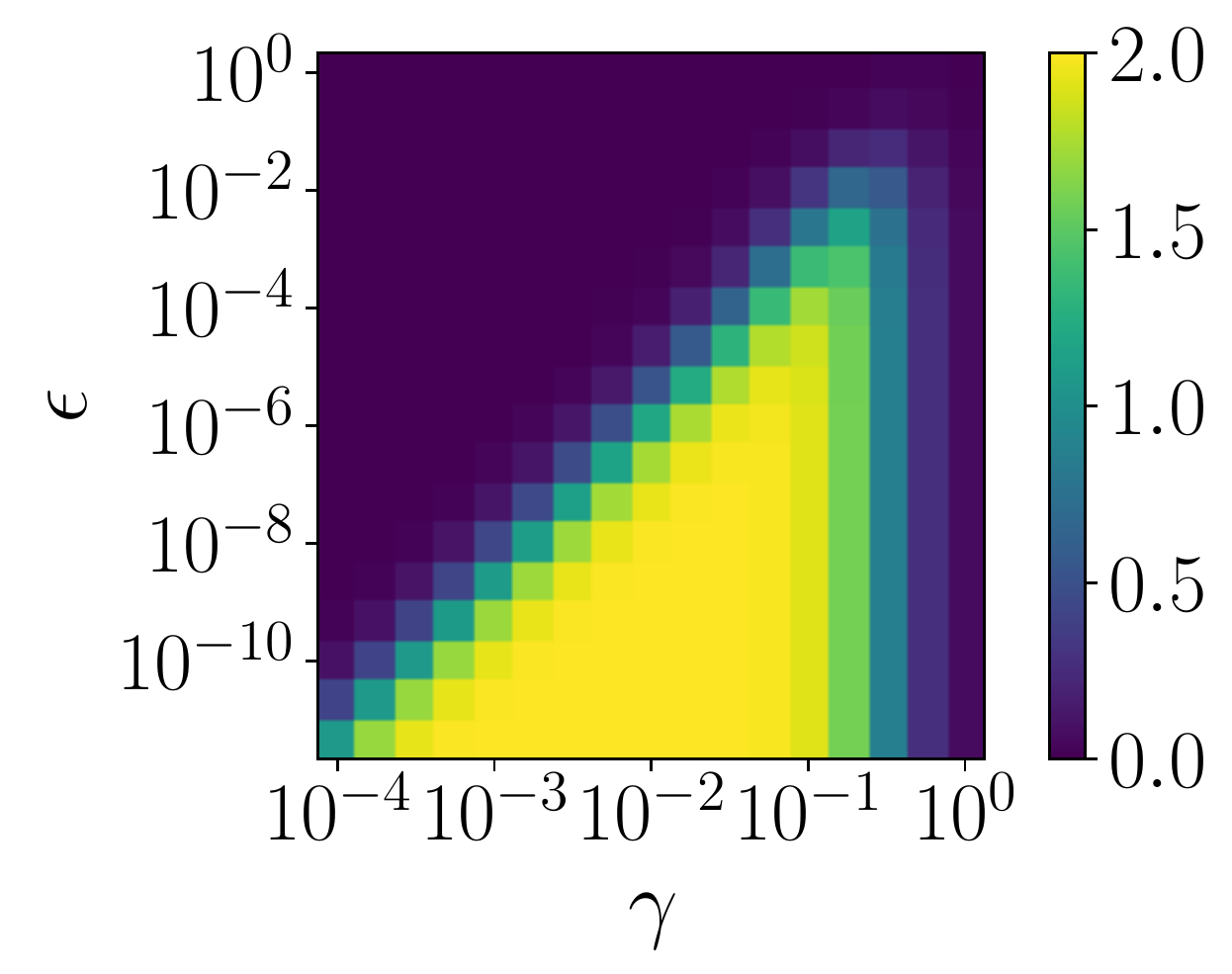}
\put(42,25){$c_\gamma = 2$}
\end{overpic}}
\caption{
    A numerical demonstration of the Main Theorem on a synthetic data set. The top panels showcase numerical estimates of $c_\eps = \frac{\partial \log(\mathcal{I})}{\partial \log(\epsilon)}$
    for different $\alpha$ values and the bottom panels showcase the numerical estimates of $c_\gamma = \frac{\partial \log(\mathcal{I})}{\partial \log(\gamma)}$. In the dark blue regions, $c_\eps, c_\gamma \approx 0$, indicating that $\mathcal{I}$ stays approximately flat with
    respect to the respective variable $\epsilon$ or $\gamma$ and so contraction has approximately ceased; the slope of the brighter regions is annoated in each panel and implies posterior contraction.
    The transition between the dark and bright regions occurs approximately at $\epsilon = \gamma^{2/\min\{1, \alpha\}}$.\label{fig:slope}
}
\end{figure}

In Figure~\ref{fig:slope} we study this phenomenon in more
detail. Let us define
$$
c_\epsilon := \partial \log(\mcl I)/\partial \log (\epsilon)  \ge 0 \text{ and } c_\gamma := \partial \log(\mcl I)/\partial \log (\gamma)  \ge 0,
$$
which correspond to contraction rates of $\mcl O(\eps^{c_\eps})$ and  $\mcl O(\gamma^{c_\gamma})$ respectively.
We present surface plots in Figure~\ref{fig:slope} of $c_\epsilon$ (top row) and $c_\gamma$ (bottom row) as functions of $\epsilon, \gamma$ for various values of $\alpha$.
Darker (lighter) regions correspond to smaller (larger) values of the logarithmic slopes $c_\eps, c_\gamma$. In regions with lighter values (i.e. $c_\eps, c_\gamma > 0$), we observe posterior contraction because the logarithmic slopes are nonzero.
The darker regions correspond to instances where the contraction has approximately ceased as indicated by the logarithmic slopes being zero. This is the phenomenon that is displayed in
Figure~\ref{fig:I vs gamma}, where the value of $\mcl I$ reduces with respect 
to $\gamma$ up to the point where the errors saturate  at an 
$\epsilon$-dependent value as predicted by the bounds in Theorem~\ref{thm:main}.

In the bottom row of Figure~\ref{fig:slope}, horizontal 
``slices'' of the plot correspond to a fixed value of $\epsilon$ which is how Figure~\ref{fig:I vs gamma} can
be obtained.
Going from right to left, we observe that the contraction rate is on the order of $\gamma^2$, until the point that $\gamma^2 \approx \eps^{\min\{1, \alpha\}}$
when our theory predicts that the $\mcl I$ will saturate and contraction has stopped, i.e., $c=0$. These plots illustrate the sharpness of our theoretical bounds
of Theorem~\ref{thm:main} for the posterior contraction measure $\mcl I$.
Similar results, with the roles of $\epsilon$ and $\gamma$ swapped, are
seen in the top row of Figure~\ref{fig:slope}.

\subsubsection{Numerics In Support Of Lemmata~\ref{lemma:1} to \ref{lemma:3}}\label{sec:numer-supp-lemm-1}

\begin{figure}
\subfloat[$\alpha = 0.5$]{\begin{overpic}[width=0.32\textwidth]{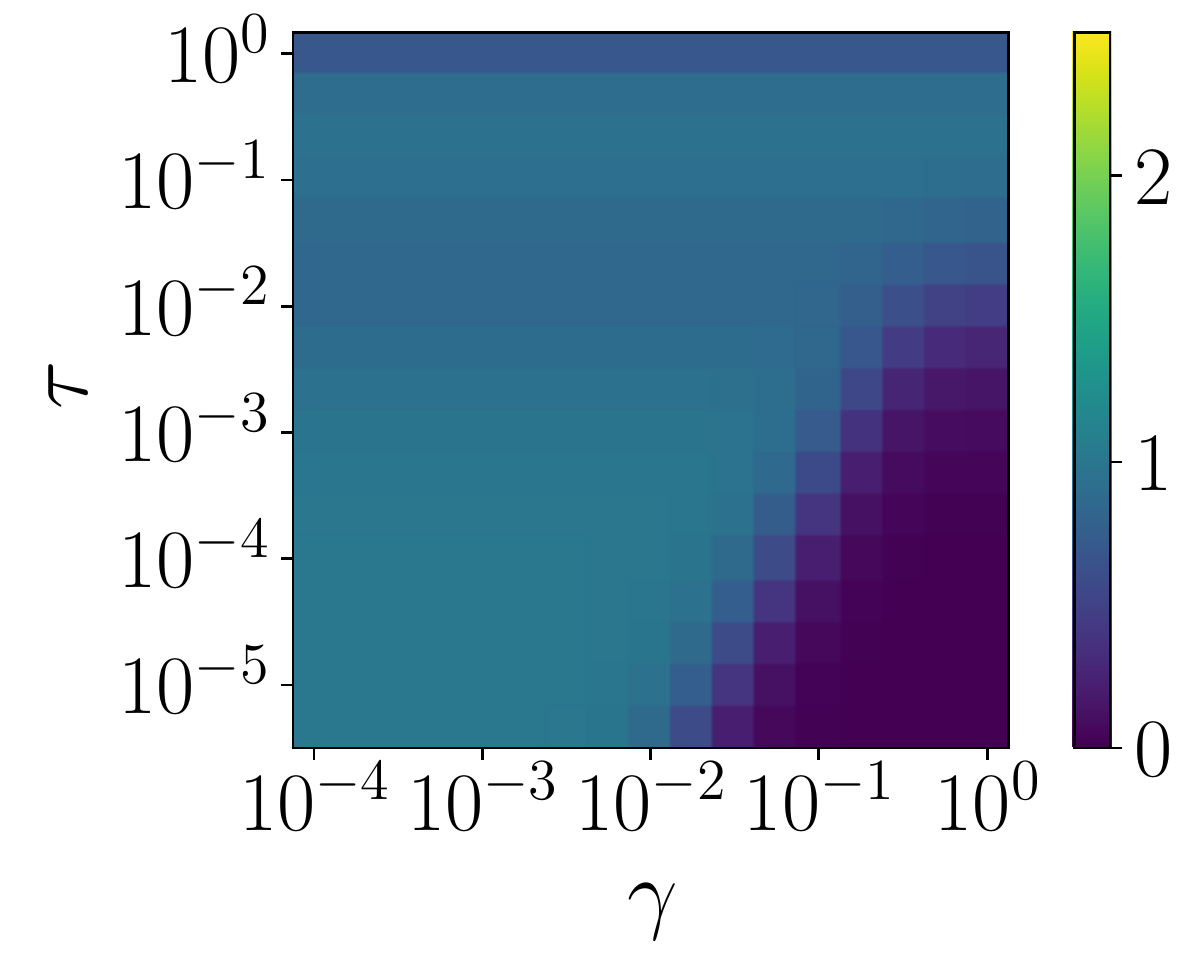}
\put(30,60){\color{white}$c_\tau = 1$}
\end{overpic}
}
\subfloat[$\alpha = 1$]{\begin{overpic}[width=0.32\textwidth]{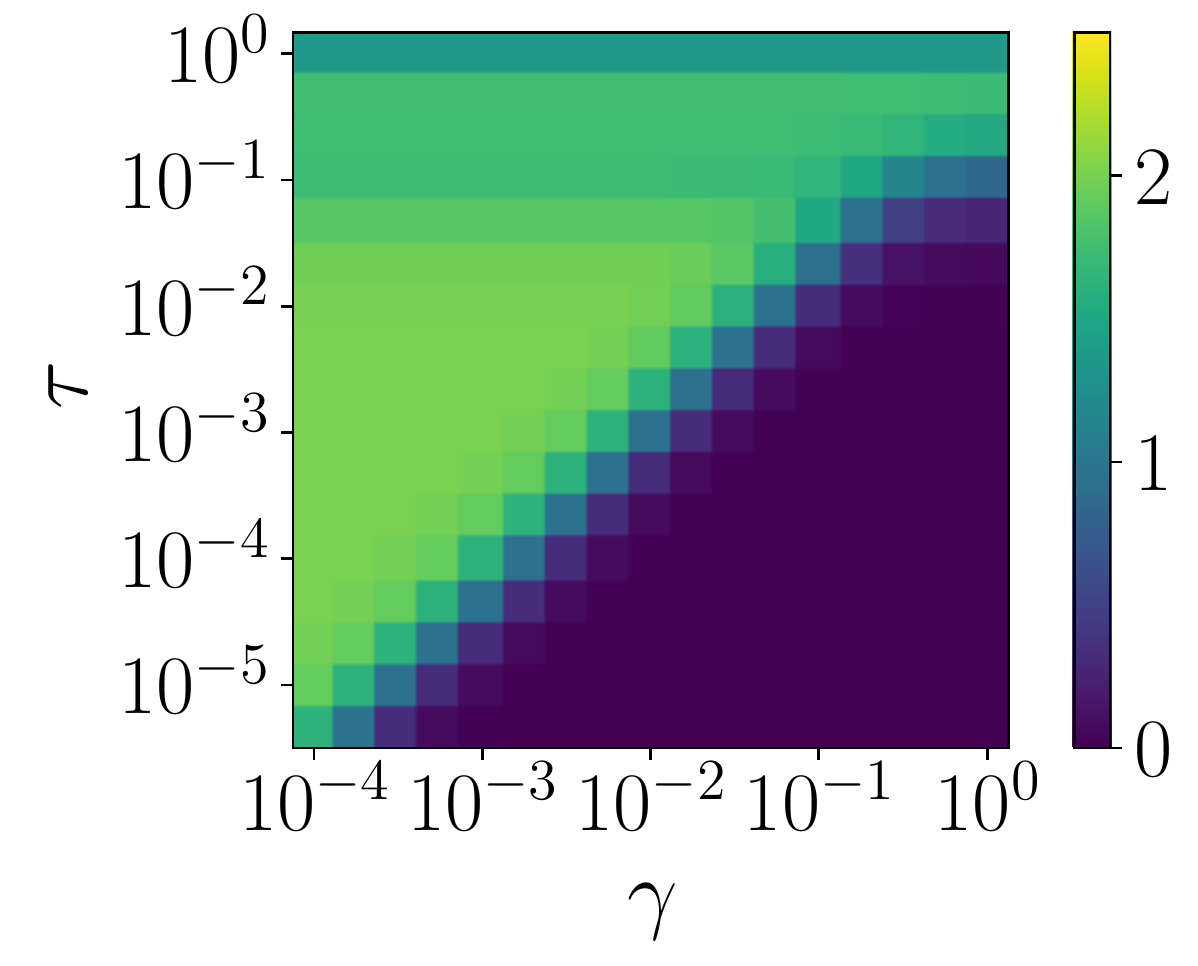}
\put(30,60){$c_\tau = 2$}
\end{overpic}
}
\subfloat[$\alpha = 1.25$]{\begin{overpic}[width=0.32\textwidth]{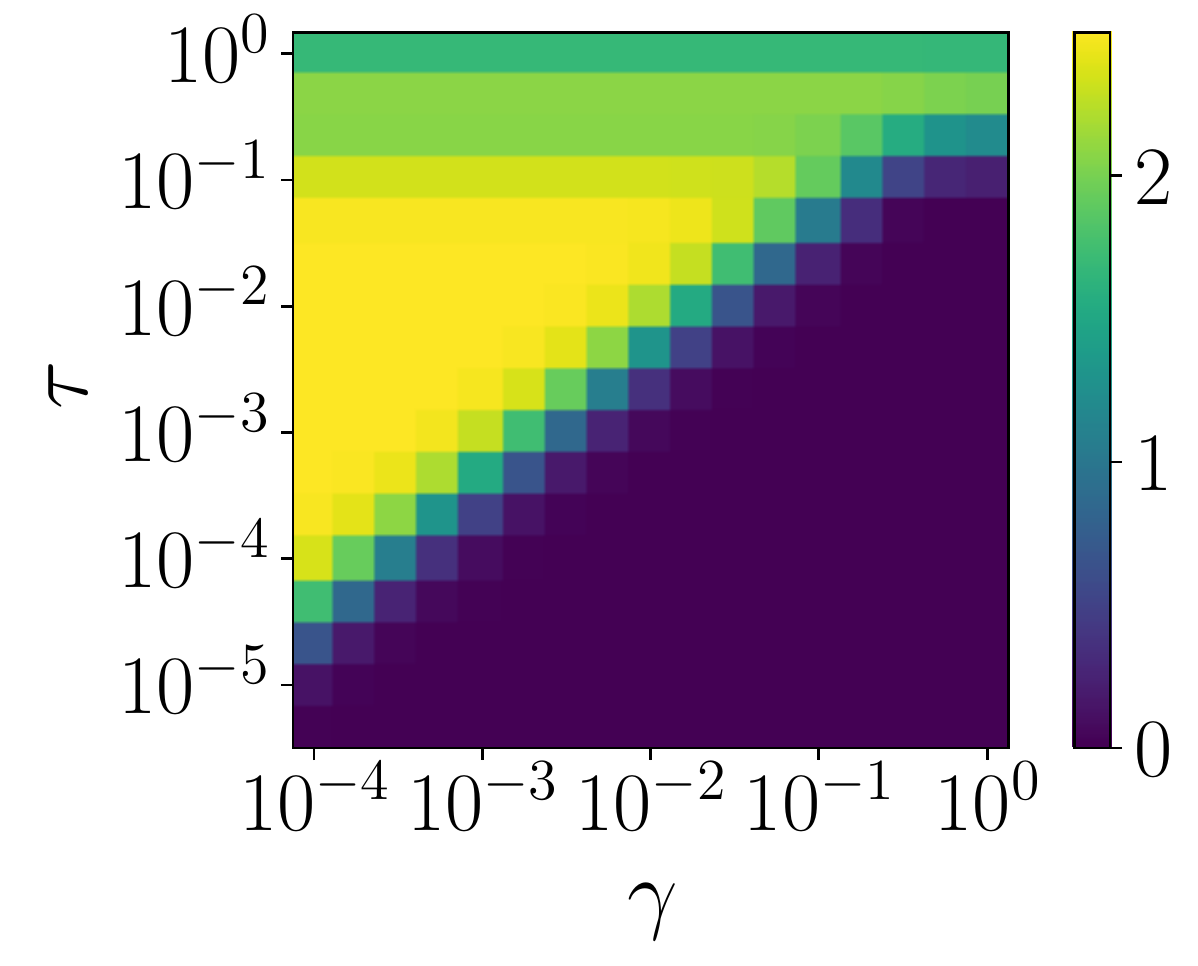}
\put(30,60){$c_\tau = 2.5$}
\end{overpic}}\\
\subfloat[$\alpha = 0.5$]{\begin{overpic}[width=0.32\textwidth]{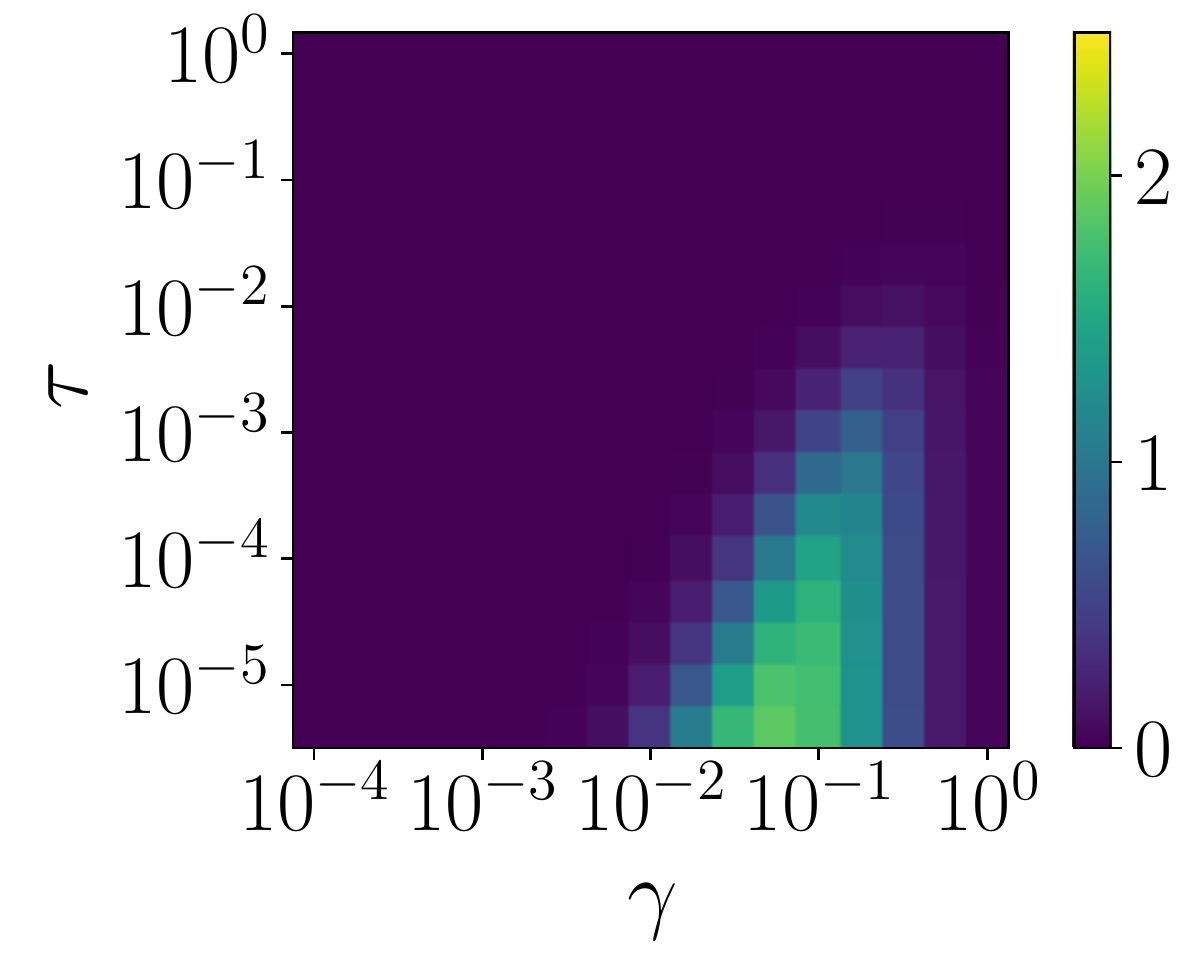}
\put(55,25){\color{white}$c_\gamma = 2$}
\end{overpic}
}
\subfloat[$\alpha = 1$]{\begin{overpic}[width=0.32\textwidth]{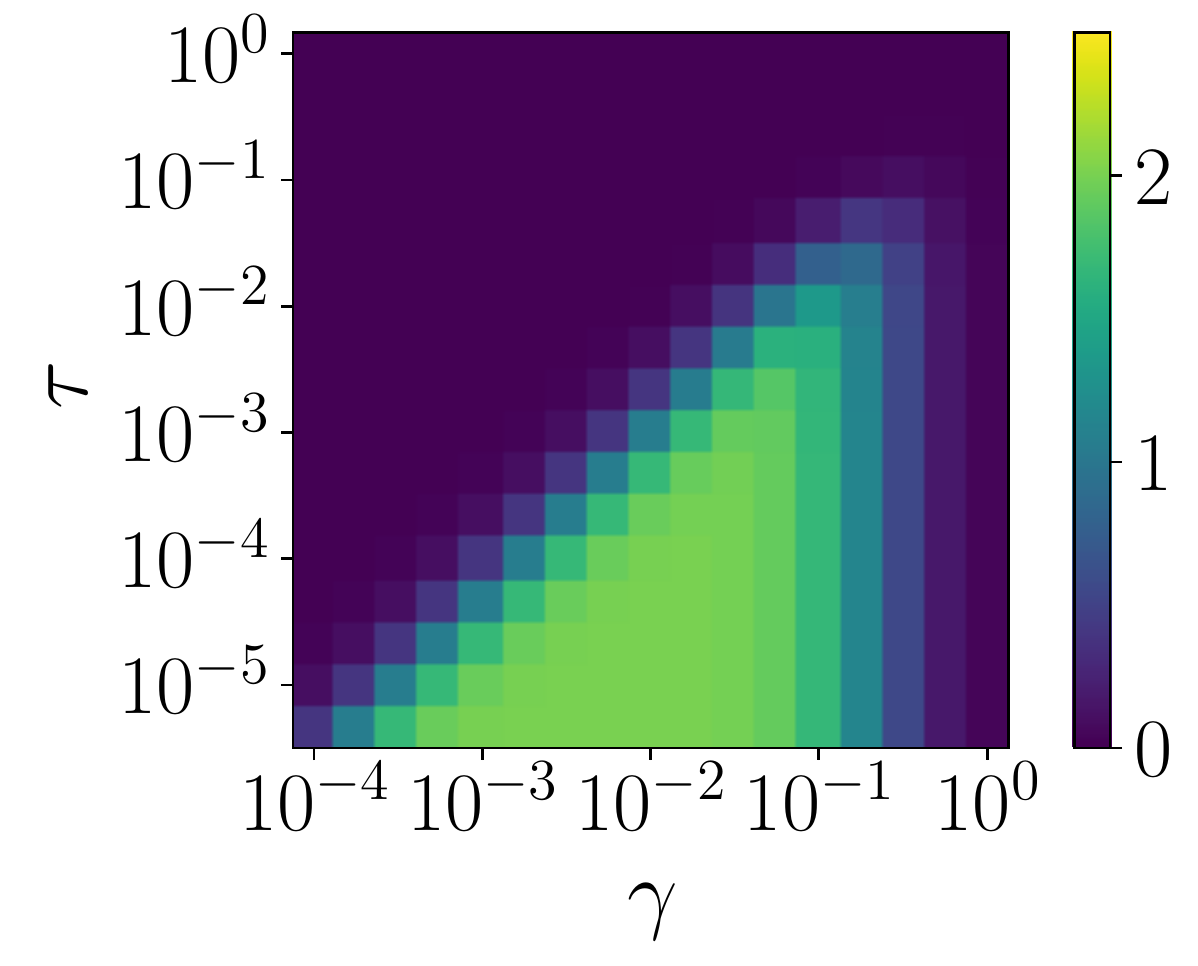}
\put(45,25){$c_\gamma = 2$}
\end{overpic}
}
\subfloat[$\alpha = 1.25$]{\begin{overpic}[width=0.32\textwidth]{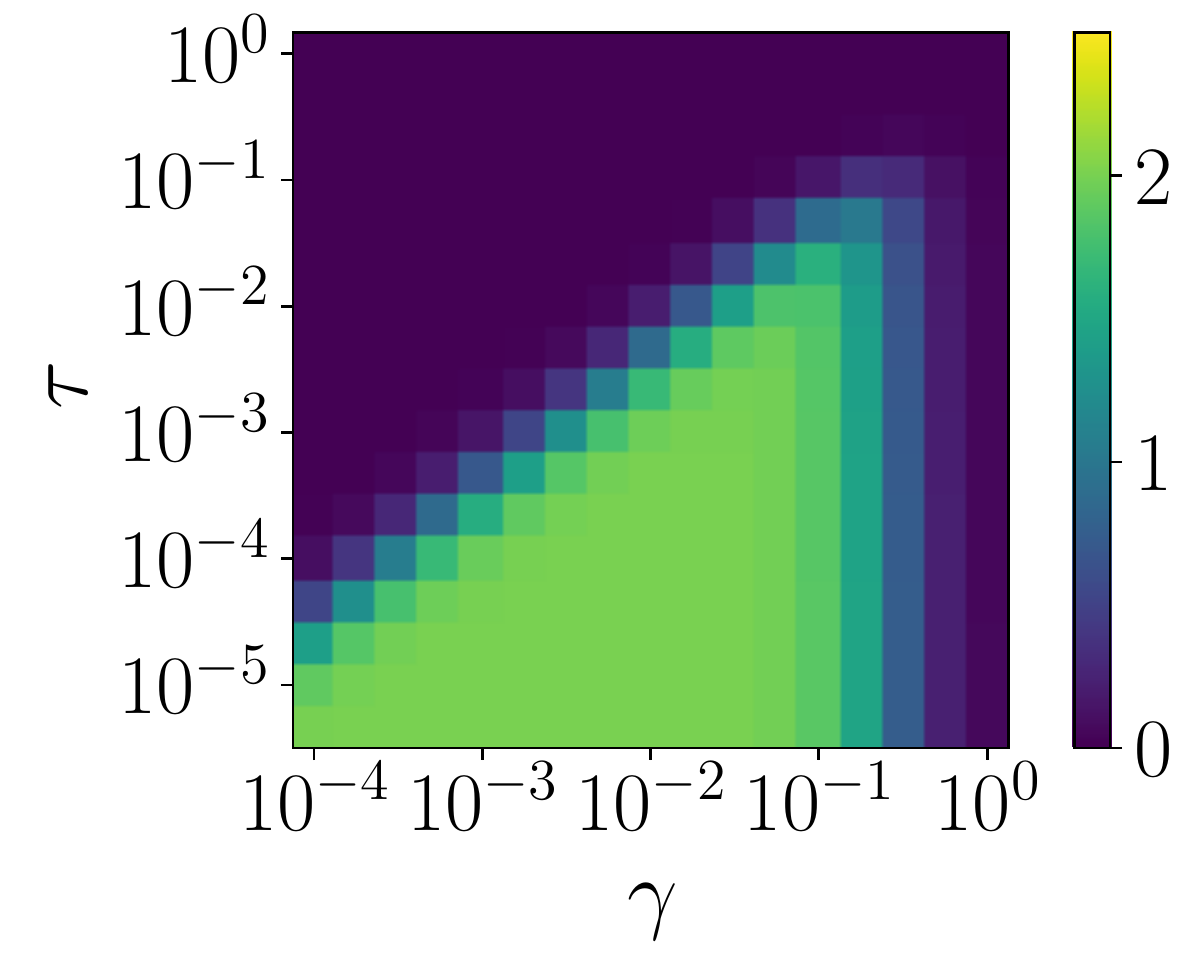}
\put(45,25){$c_\gamma = 2$}
\end{overpic}}
\caption{A numerical demonstration of Lemma~\ref{lemma:1} on the synthetic data set with $\epsilon = \tau^{2\alpha}$. The top panels showcase the numerical estimates of the logarithmic slope $c_\tau :=\frac{\partial\log(\mathrm{Tr}(C_\epsilon^\ast))}{\partial \log(\tau)}$ for different $\alpha$ values and the bottom panels showcase the numerical estimates of the logarithmic slope $c_\gamma := \frac{\partial \log(\mathrm{Tr}(C_\epsilon^\ast))}{\partial \log(\gamma)}$. 
In the dark blue region, $c_\tau,c_\gamma \approx 0$, indicating that $\mathrm{Tr}(C_\epsilon^\ast)$ stays approximately flat with respect to the respective variable $\tau$ or $\gamma$; the slope of the brighter regions is annotated in each panel. The transition between the dark and bright regions occurs approximately at $\tau = \gamma^{1/\alpha}$. \label{fig:dtrc_connected}}
\end{figure}

\begin{figure}
\subfloat[$\alpha = 0.5$]{\begin{overpic}[width=0.32\textwidth]{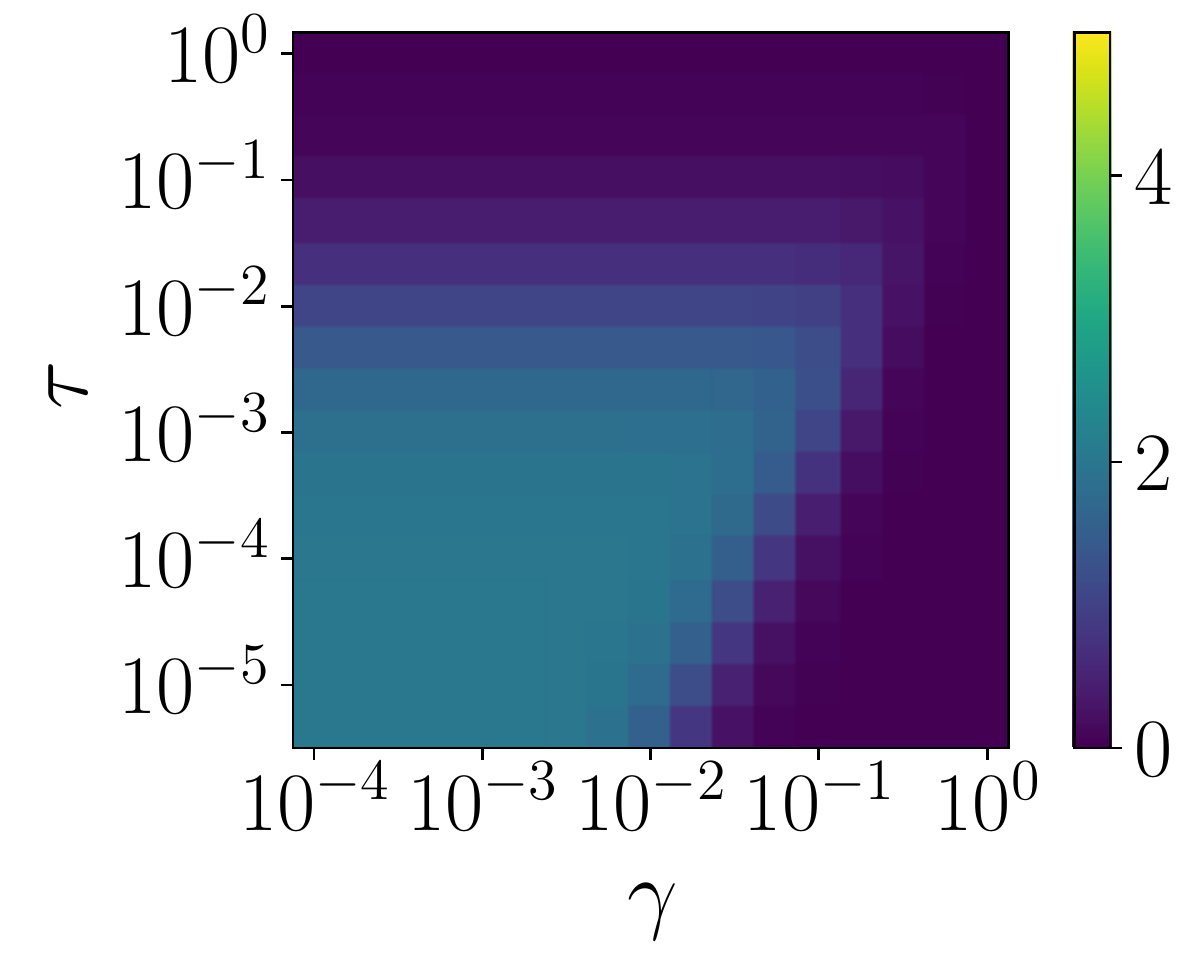}
\put(30,30){\color{white}$c_\tau = 2$}
\end{overpic}
}
\subfloat[$\alpha = 1$]{\begin{overpic}[width=0.32\textwidth]{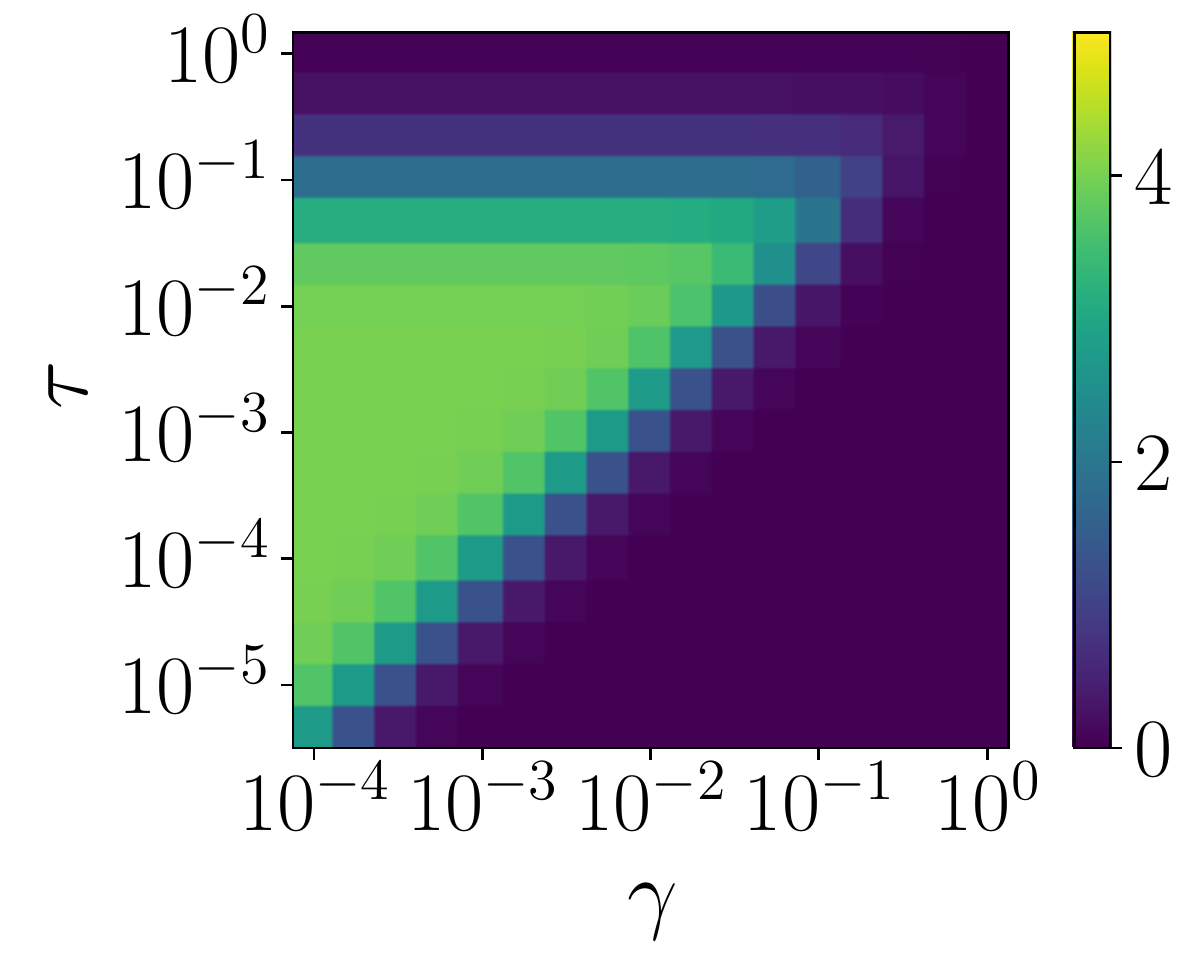}
\put(30,55){$c_\tau = 4$}
\end{overpic}
}
\subfloat[$\alpha = 1.25$]{\begin{overpic}[width=0.32\textwidth]{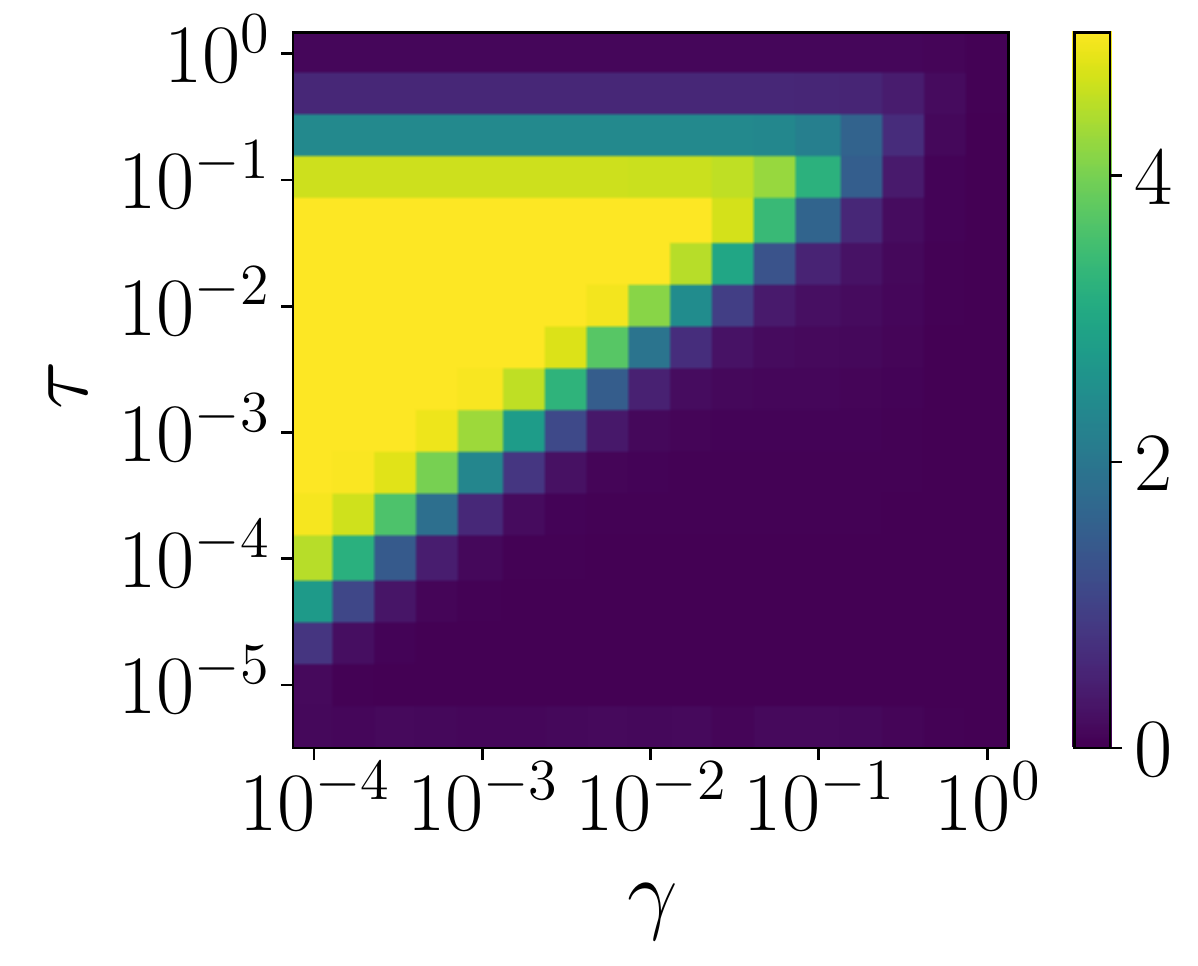}
\put(30,55){$c_\tau = 5$}
\end{overpic}}\\
\subfloat[$\alpha = 0.5$]{\begin{overpic}[width=0.32\textwidth]{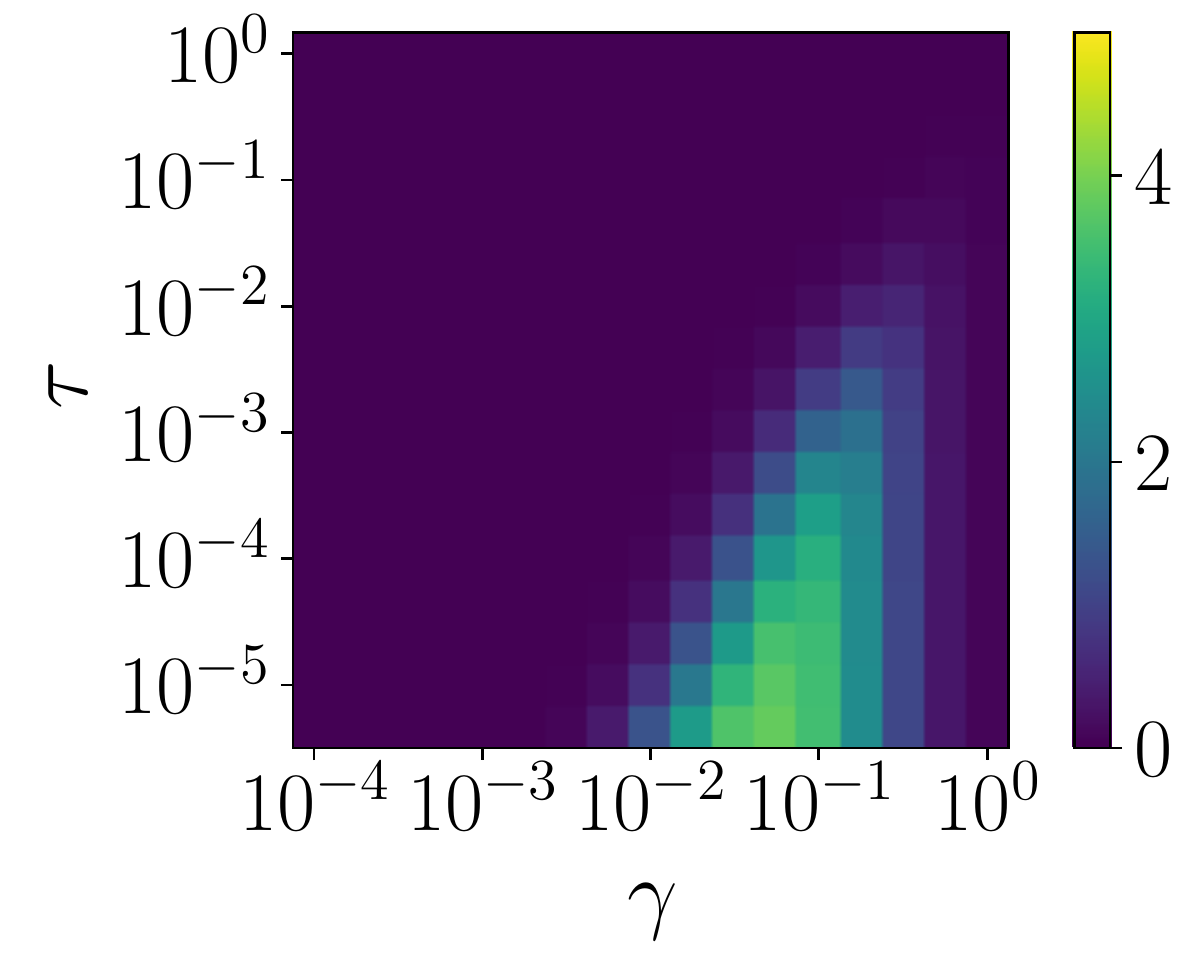}
\put(55,25){\color{white}$c_\gamma = 4$}
\end{overpic}
}
\subfloat[$\alpha = 1$]{\begin{overpic}[width=0.32\textwidth]{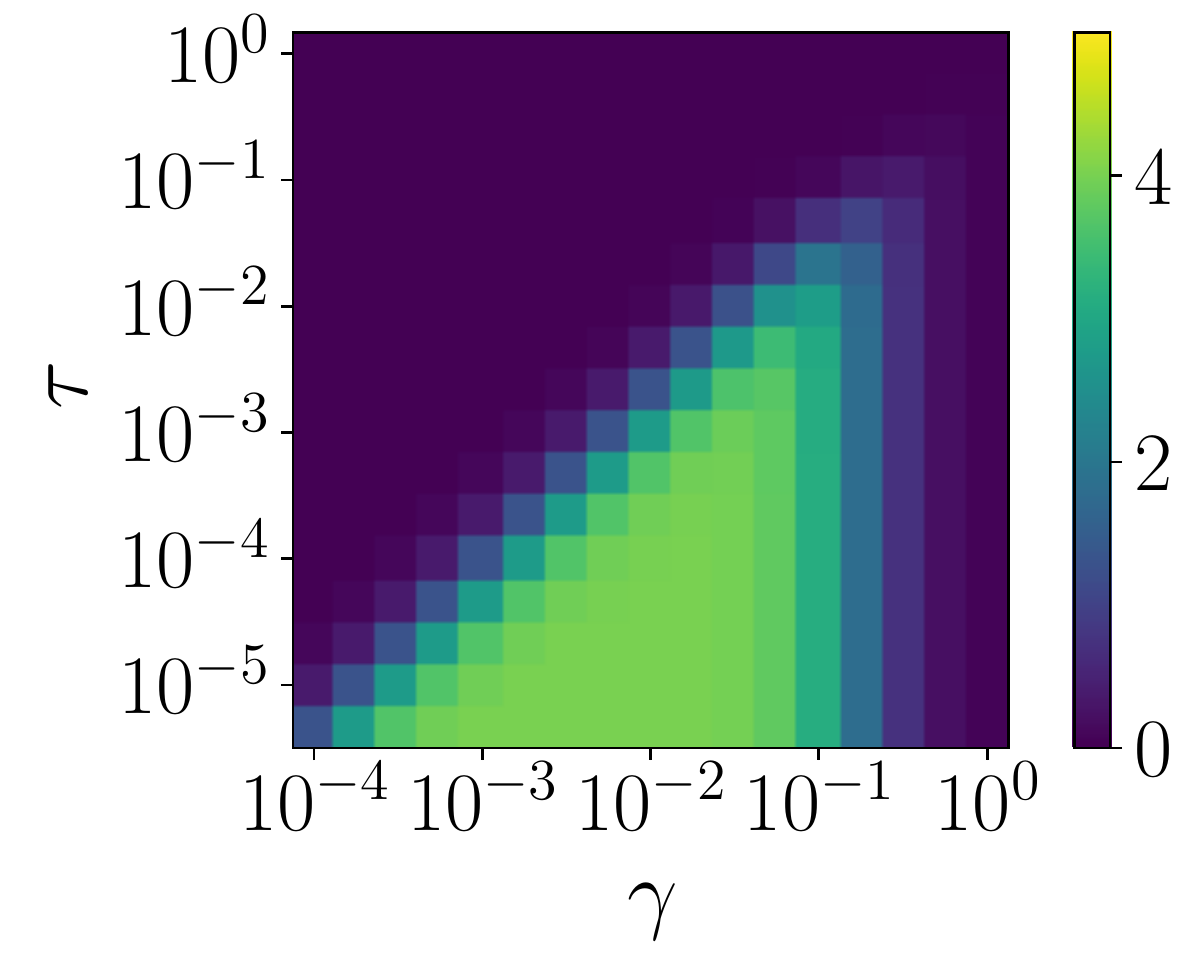}
\put(45,25){$c_\gamma = 4$}
\end{overpic}
}
\subfloat[$\alpha = 1.25$]{\begin{overpic}[width=0.32\textwidth]{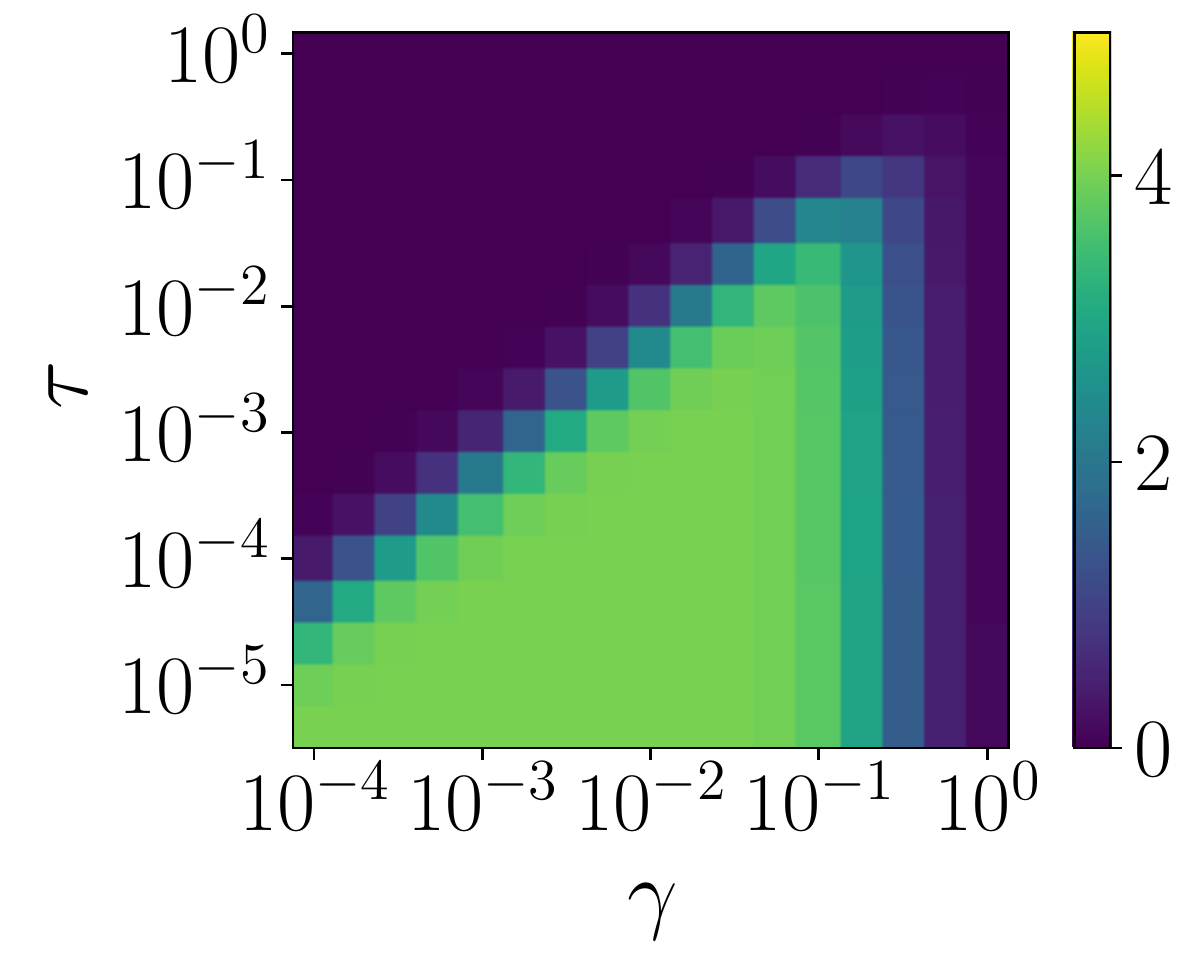}
\put(45,25){$c_\gamma = 4$}
\end{overpic}}
\caption{A numerical demonstration of Lemma~\ref{lemma:3} on a synthetic data set with $\epsilon = \tau^{\max\{2, 2\alpha\}}$. The top panels showcase the numerical estimates of the logarithmic slope $c_\tau := \frac{\partial \log(\biasc^2)}{\partial \log(\tau)}$ for different $\alpha$ values and the bottom panels showcase the numerical estimates of the logarithmic slope $c_\gamma := \frac{\partial \log(\biasc^2)}{\partial \log(\gamma)}$. 
In the dark blue region, $c_\tau,c_\gamma \approx 0$, indicating that $\biasc^2$ stays approximately flat with respect to the respective variable $\tau$ or $\gamma$; the slope of the brighter regions is annotated in each panel. The transition between the dark and bright regions occurs approximately at $\tau = \gamma^{1/\alpha}$. \label{fig:dbias_connected}}
\end{figure}

In Figures~\ref{fig:dtrc_connected} and~\ref{fig:dbias_connected} we present numerics that illustrate the convergence results for Lemmata~\ref{lemma:1} and~\ref{lemma:3} respectively. These lemmata respectively bound the first and third terms of the decomposition of $\mcl I$:
\[
  \mcl I(\gamma, \alpha, \tau, \eps)
  = M \text{Tr}(\poC_\eps) + \frac{M}{\gamma^2}\text{Tr}(\poC_\eps B \poC_\eps)
  + \sum_{m=1}^M
  \left|\frac{1}{\gamma^2}\poC_\eps B\bu^\dagger_m - \bu^\dagger_m \right|^2.
\]

Again, we omit numerics for the middle term in this decomposition since the corresponding bound in Lemma~\ref{lemma:2} is derived from the bound found for $\mathrm{Tr}(\poC_\eps)$ in Lemma~\ref{lemma:1} and exhibit nearly identical behavior numerically. Just as in Figures~\ref{fig:slope} and~\ref{fig:MNIST slope}, we have set the scaling $\eps = \tau^{\max\{2, 2\alpha\}}$. The top panels in Figure~\ref{fig:dtrc_connected} show the estimated rate of convergence of $\mathrm{Tr}(\poC_\eps)$ in terms of $\tau$ in the log-log scale, while the bottom panels show the estimated rate of convergence in terms of $\gamma$ in the log-log scale. Figure~\ref{fig:dbias_connected} likewise shows the estimated rate of convergence $\biasc$ 
in the parameters $\eps$ and $\gamma$.  From Figure~\ref{fig:dtrc_connected}, we read that in the region where $\gamma^2 \ll \tau^{2\alpha}$, $\partial \log(\mathrm{Tr}(\poC_\eps))/\partial\log(\tau)$ stays close to $2\alpha$ whereas $\partial \log(\mathrm{Tr}(\poC_\eps))/\partial\log(\gamma)$ is approximately 0. In the region where $\tau^{2\alpha} \ll \gamma^{2}$, we observe that $\partial \log(\mathrm{Tr}(\poC_\eps))/\partial\log(\tau)$ is close to 0 whereas $\partial \log(\mathrm{Tr}(\poC_\eps))/\partial\log(\gamma)$ is around $2$. These results confirm our bound presented in Lemma~\ref{lemma:1}.

In Figure~\ref{fig:dbias_connected}, we read that in the region where $\gamma^2 \ll \tau^{2\alpha}$, $\partial \log(\biasc^2)$ $/\partial\log(\tau)$ stays close to $4\alpha$ whereas $\partial \log(\biasc^2)/\partial\log(\gamma)$ is approximately 0. In the region where $\tau^{2\alpha} \ll \gamma^{2}$, we observe that $\partial \log(\biasc^2)$ $/\partial\log(\tau)$ is close to 0 whereas $\partial \log(\biasc^2)/\partial\log(\gamma)$ is around $4$. These results confirm our bounds presented in Lemma~\ref{lemma:3}.


\subsection{MNIST Data}\label{sec:mnist-data} 

In this subsection we use the MNIST dataset\cite{lecun2010mnist} to test our theory on
an empirical dataset. MNIST is a dataset of 70,000 grayscale $28\times28$ pixel images of
handwritten digits (0--9), of which we use only the digits 1, 4, and 7. Each image is represented by a vector $\bx_i \in \mbb R^{784}$ and we normalize the pixel values to range from 0 to 1.
To confirm our theory in practice presents the issue of determining 
how to control the parameter $\epsilon$ that is inherent to the 
clustering structure of a given fixed unlabeled dataset $X$ given 
in application. However, in this example, we may use the fact that every image
is labeled and so the clustering structure of the dataset is known.
Using this, we may devise an  $\epsilon$-dependent parameter set
to observe what happens in the $\epsilon \to 0$ limit. 

First, we create a similarity graph $G$ based on the unlabeled data $X$ of reshaped images $\bx_i \in \mbb R^{784}$.
Given the known clustering (i.e. class memberships) of the points in the MNIST dataset, we can identify the {\it inter-cluster} edges, those edges that connect vertices of different clusters corresponding to different digits. 
If the original weight matrix is given by $W$, with entries $w_{ij}$,
then we scale the inter-cluster edges by $\epsilon$ to obtain $W_\eps$ as: 
\begin{equation*}
    [W_\eps]_{ij} = 
    \begin{cases}
        w_{ij}, & \text{ if } i,j \in \tilde{Z}_k, \\
        \eps w_{ij}, & \text{ if } i \in \tilde{Z}_k, j \in \tilde{Z}_\ell, \text{ with } k \not= \ell.
    \end{cases}
\end{equation*}
Sending $\eps \rightarrow 0$ then results in a disconnected graph, where each cluster represents a different digit. For all $\epsilon$ sufficiently small,
the graph Laplacian will have the structure underlying our theory. 

\begin{figure}
\subfloat[$\alpha = 0.5$]{\begin{overpic}[width=0.32\textwidth]{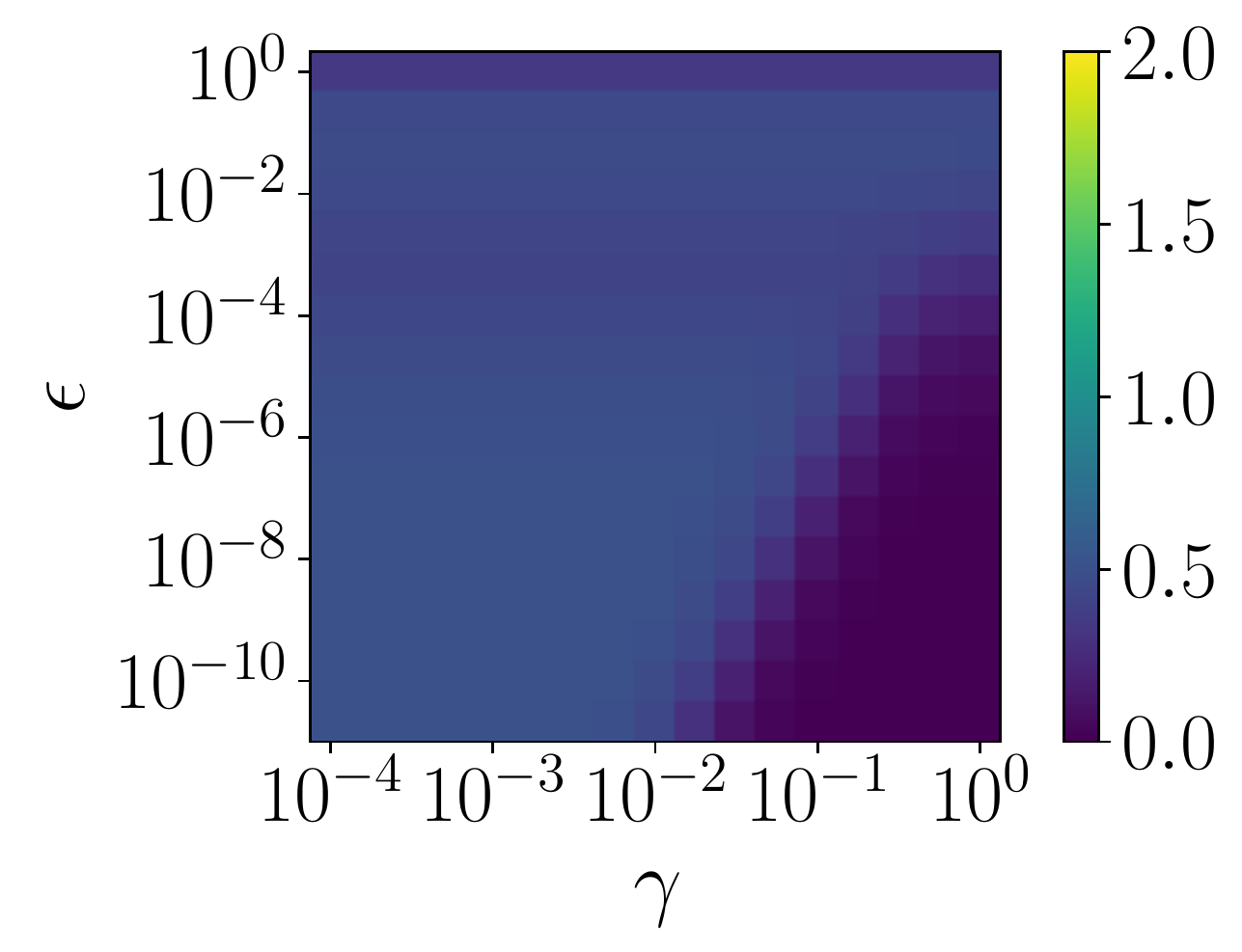}
\put(30,50){\color{white}$c_\epsilon = 0.5$}
\end{overpic}}
\subfloat[$\alpha = 1$]{\begin{overpic}[width=0.32\textwidth]{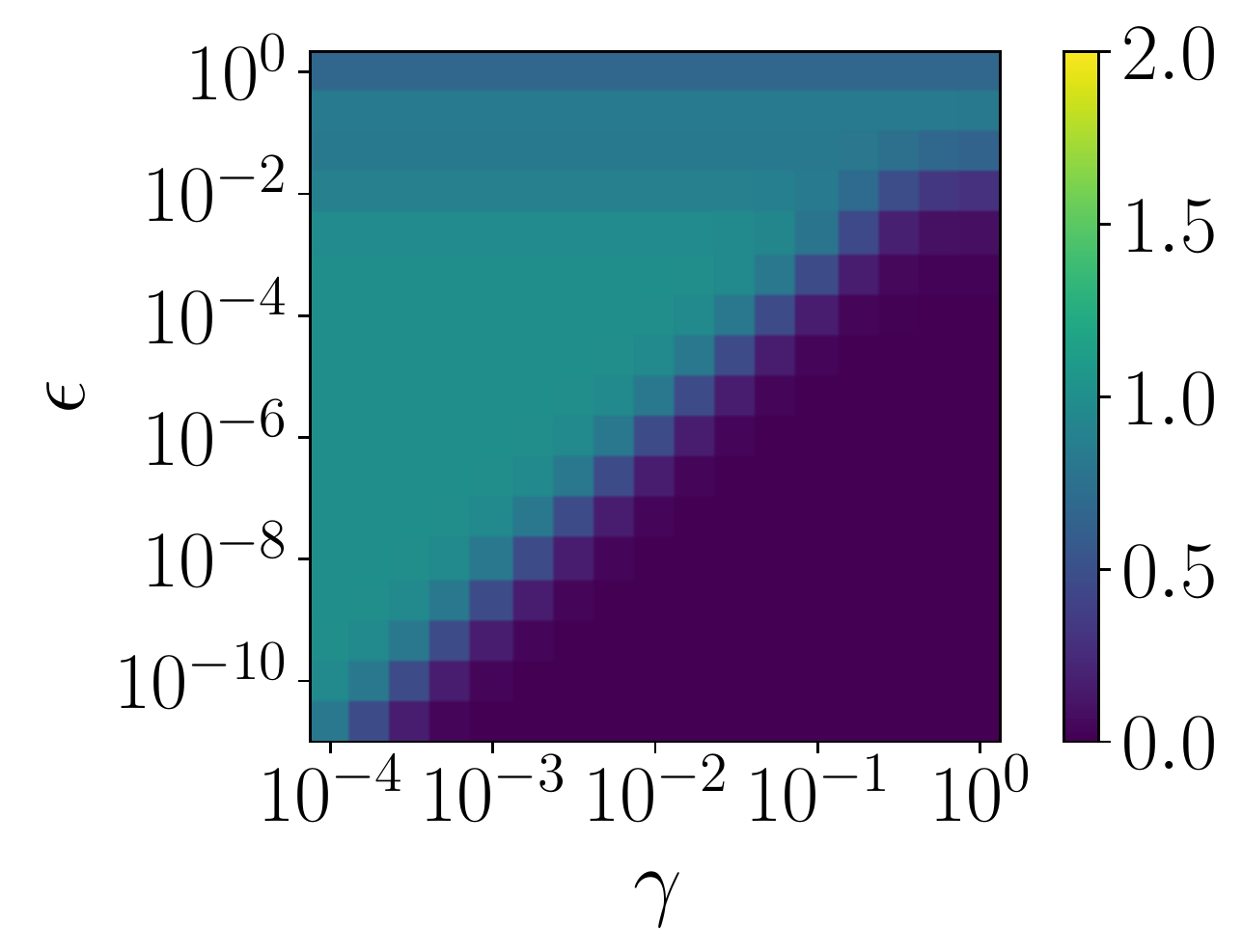}
\put(30,50){$c_\epsilon = 1$}
\end{overpic}}
\subfloat[$\alpha = 5$]{\begin{overpic}[width=0.32\textwidth]{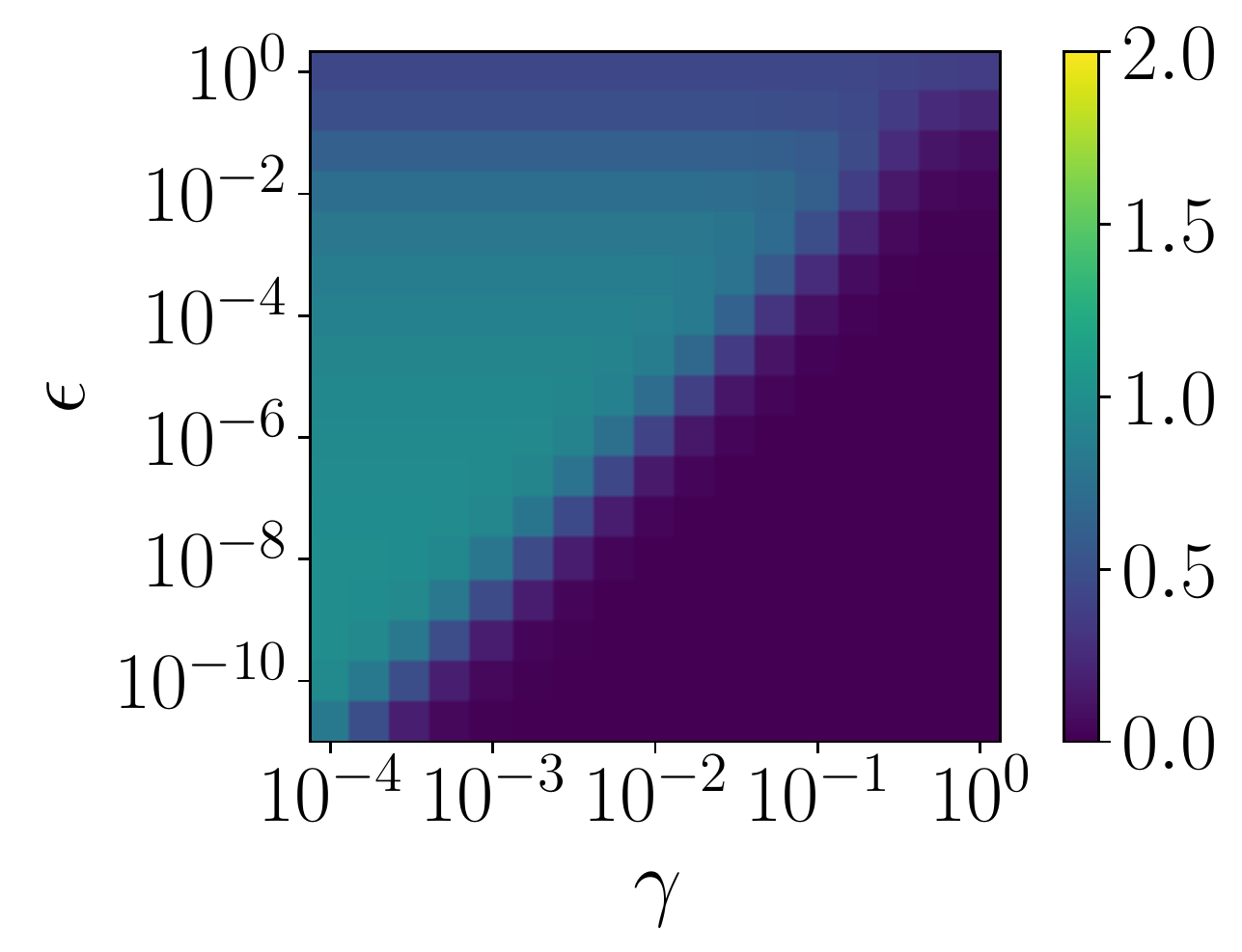}
\put(30,50){$c_\epsilon = 1$}
\end{overpic}
}\\
\subfloat[$\alpha = 0.5$]{\begin{overpic}[width=0.32\textwidth]{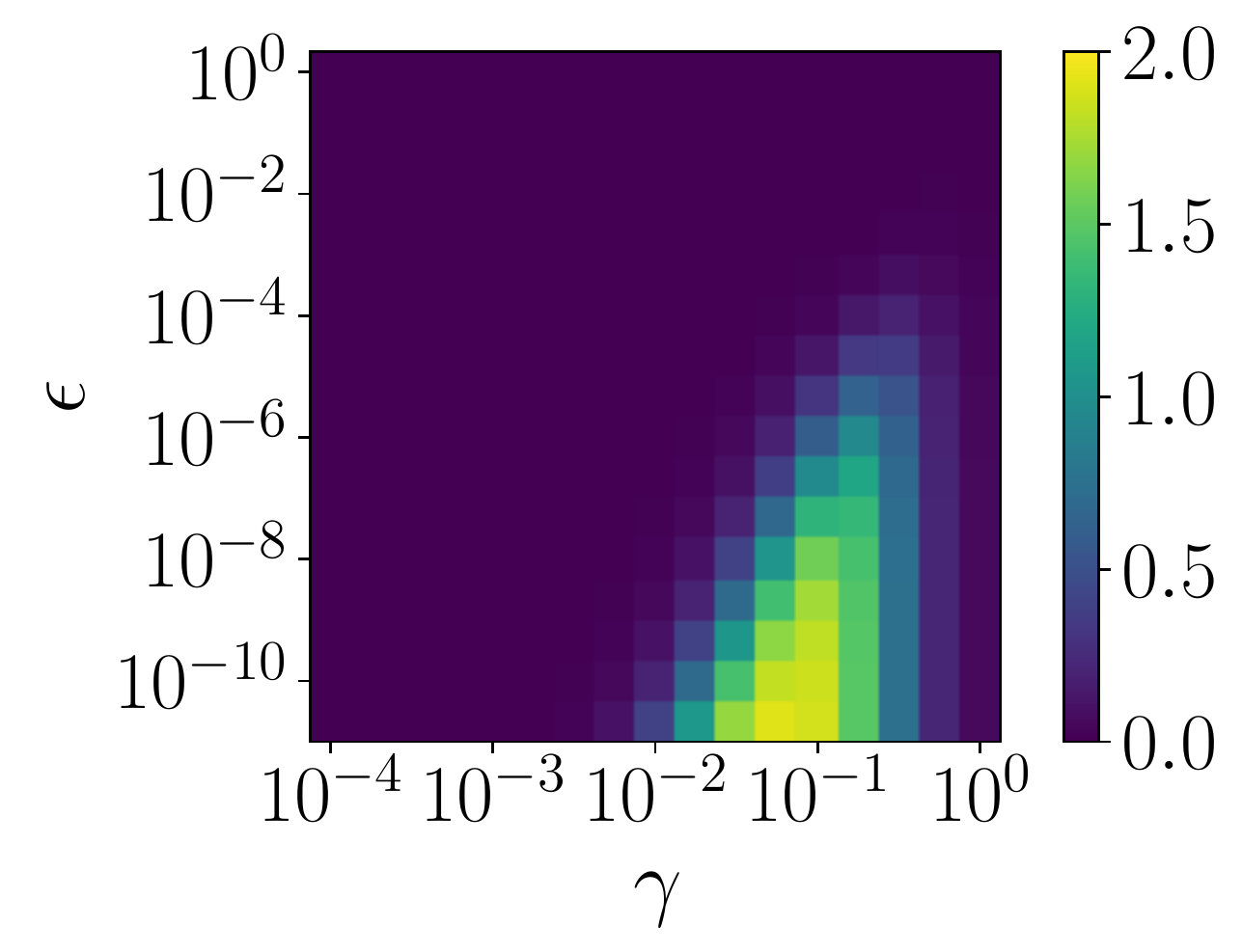}
\put(50,25){\color{white} $c_\gamma = 2$}
\end{overpic}}
\subfloat[$\alpha = 1$]{\begin{overpic}[width=0.32\textwidth]{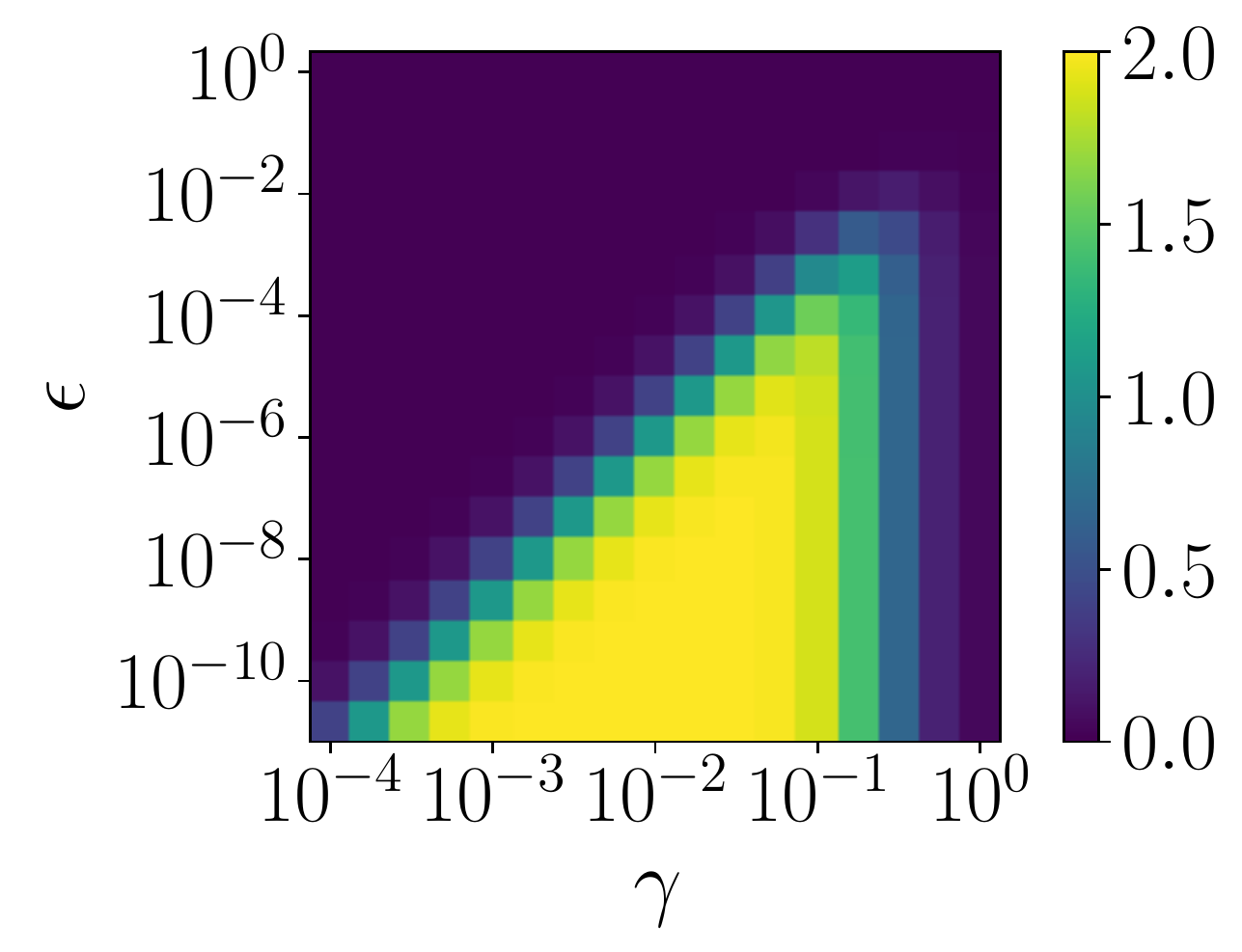}
\put(45,25){$c_\gamma = 2$}
\end{overpic}}
\subfloat[$\alpha = 5$]{\begin{overpic}[width=0.32\textwidth]{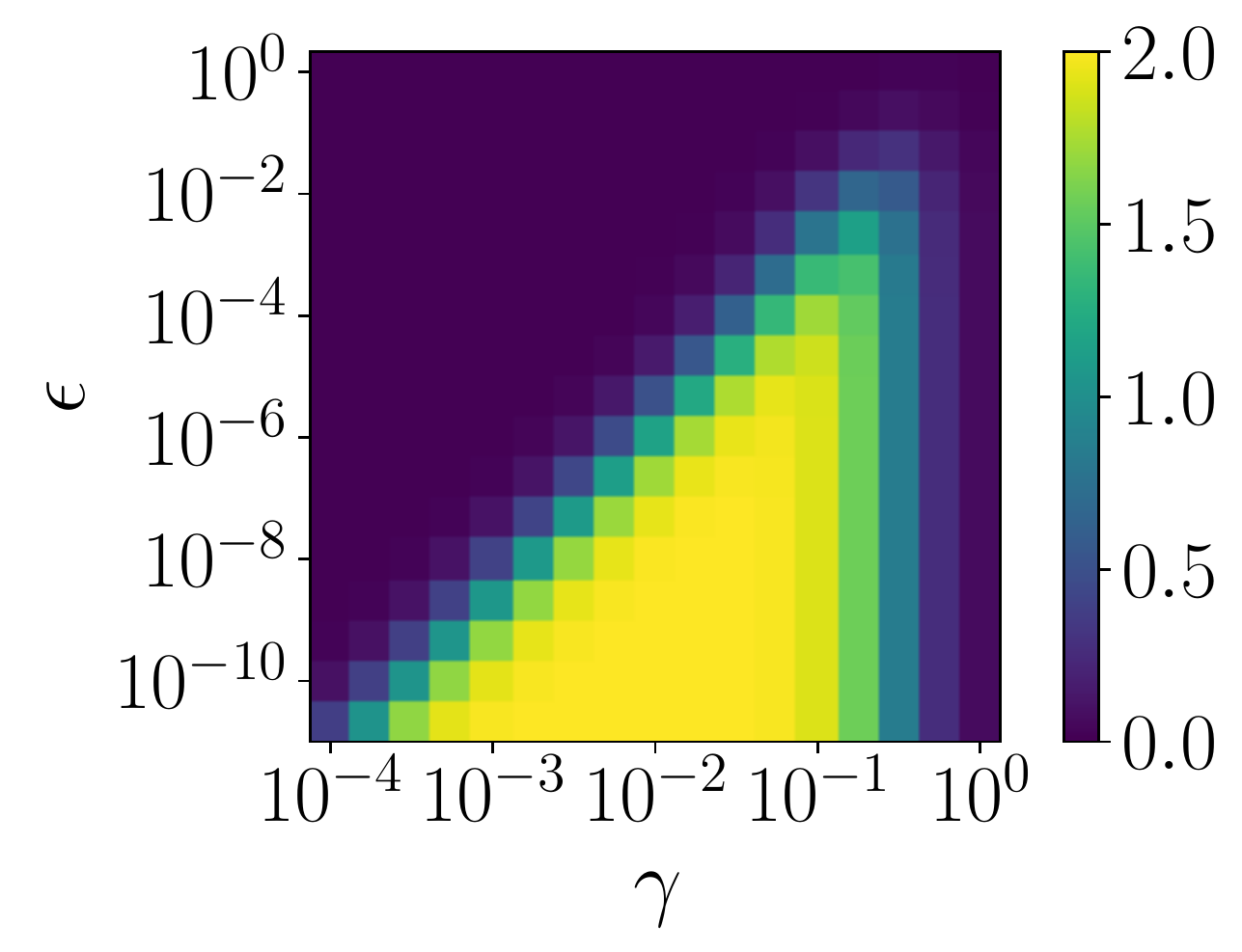}
\put(45,25){$c_\gamma = 2$}
\end{overpic}}
\caption{
  A numerical demonstration of the Main Theorem on the MNIST data set with digits 1, 4, and 7. The top panels showcase numerical estimates of $c_\eps = \frac{\partial \log(\mathcal{I})}{\partial \log(\epsilon)}$
    for different $\alpha$ values and the bottom panels showcase the numerical estimates of $c_\gamma = \frac{\partial \log(\mathcal{I})}{\partial \log(\gamma)}$. In the dark blue regions, $c_\eps, c_\gamma \approx 0$, indicating that $\mathcal{I}$ stays approximately flat with
    respect to the respective variable $\epsilon$ or $\gamma$ and so contraction has approximately ceased; the slope of the brighter regions is annotated in each panel and implies posterior contraction.
    The transition between the dark and bright regions occurs approximately at $\epsilon = \gamma^{2/\min\{1, \alpha\}}$. These results are similar to our synthetic experiment depicted in
    Figure~\ref{fig:slope}. \label{fig:MNIST slope}
  }
\end{figure}

For our experiment, we sample $100$ images uniformly at random from the digits 1, 4, and 7. The similarity graph $W=(w_{ij})$ is constructed via the Gaussian kernel and the Zelnik-Perona scaling \cite{zelnik2005self},
$w_{ij} = \exp(-|\bx_i-\bx_j|^2/r_ir_j),$
where $r_i$ is the Euclidean distance between data point $i$ and its 15th nearest neighbor. Following the same procedure as the synthetic data, we pick one vertex from each digit to be labeled and choose the ground truth $U^\dagger = [\bchi_1, \bchi_2, \bchi_3]^T$. We  evaluate the contraction measurement $\mathcal{I}$ for a range of $\epsilon$ and $\gamma$. We present the results in Figure~\ref{fig:MNIST slope}. It is clear that Figure~\ref{fig:MNIST slope} is nearly identical to Figure~\ref{fig:slope}, demonstrating that the behavior on this MNIST
dataset is close to that observed in the synthetic case; in turn
the two sets of experiments together attest to the sharpness of our contraction 
rate estimates in Theorem~\ref{thm:main}. 
Working with the MNIST dataset highlights the relevance of 
our analysis to real-world SSR applications.

\section{Conclusions}\label{sec:conclusions}

The work in this paper is, to the best of our knowledge, 
the first analysis of Bayesian posterior consistency in 
semi-supervised regression (SSR). The regression formulation 
of semi-supervised learning is convenient for both computations and analysis 
due to conjugacy of Gaussian likelihoods and priors, leading
to a Gaussian posterior. The resulting closed form is
useful in practice \cite{ZhuThesis2005} and for theory,
such as that developed in this paper. We formulate the SSR
problem as a BIP in which the unlabeled data
defines the prior and the labeled data defines the likelihood.
By postulating coherence between the labeled and unlabeled data
we are able to quantify the convergence of the posterior distribution
to the truth in terms of the noise in the labels and a measure
of clustering in the data. As a by-product of the analysis we also learn about
parameter choices within the data-informed prior construction.

However the SSR formulation has some undesirable model characteristics
relating to the fact that the latent variable $U$, which is 
real-valued, and the labels, which are categorical,
are seen as elements of the
same space. A fruitful avenue for future study is to
combine the work in this paper with that developed in
\cite{HHRS19}, where consistency of probit-based optimization
is studied, in order to analyze Bayesian posterior consistency for 
probit-based approaches to SSL. The probit methodology postulates
a link function connecting the latent variable to labels,
a concrete example being the use of the sign function in
binary classification \cite{bertozzi2018uncertainty}.
Another interesting direction for theoretical analyses of SSR
concerns active learning as pioneered in \cite{zhu2003combining}.
The framework and methodology developed here will be useful in developing
principled theories of active learning.

\vspace{0.1in}
\noindent{\bf Acknowledgements} We are grateful to Mason Porter for helpful comments
that improved the paper. 

\vspace{0.1in}

\appendix

\section{Proof of Lemmata}\label{sec:proof-main-results}

In this appendix we start by discussing useful properties of the 
posterior measure in Subsection~\ref{sec:char-post}; in
particular we show that the posterior is Gaussian and give closed form expressions for its mean and covariance. Subsections~\ref{ssec:A2}, \ref{ssec:A3} we present the  detailed proofs of the lemmata used to prove
our main results, Theorems~\ref{thm:post-contraction-disc-graph}
and \ref{thm:main}. Numerical experiments which illustrate 
these lemmata are contained in Subsection~\ref{sec:numer-supp-lemm-1}..
  
\subsection{Characterizing the Posterior}\label{sec:char-post}
Here we collect some results that completely characterize the posterior measure $\mu^Y$ as a Gaussian measure with explicit formulae for its mean and covariance.
\begingroup
\allowdisplaybreaks
\begin{proposition} \label{prop:posterior}
  Consider the  posterior measure $\mu^Y$ given by \eqref{eq:posterior}.
  Then
  \begin{enumerate}[(i)]
  \item $\mu^Y = \mcl N( U^\ast, I_M \otimes C^\ast)$ and 
 has Lebesgue density
  \begin{equation}
    \label{posterior-product-form}
    \begin{aligned}
      \mu^Y(\dd U) & = \frac{1}{\vartheta(Y)} \exp \left( - \frac{1}{2} \left\langle (U - \poU)^T ,
      (\poC)^{-1} (U - \poU)^T \right\rangle_{F} \right) \dd U                                      \\
                   & \equiv \frac{1}{\vartheta(Y)}
      \prod_{m=1}^M \exp \left( -\frac{1}{2}
      \big\langle (\bu_m - \pom_m), (\poC)^{-1} (\bu_m - \pom_m)  \big\rangle    \right)
      \dd \bu_\ell.
    \end{aligned}
  \end{equation}
  Here $U^\ast$ is the posterior mean with rows  $(\pom_m)^T$ 
  and $C^\ast$ is the covariance matrix of each row $(\pom_m)^T$, independent of $m$.
  \item The posterior means $\pom_m$ and covariances $C^\ast$ are given by
  \begin{equation}
    \pom_m = \frac{1}{\gamma^2}\poC H^T\by_m, \qquad
    \poC = \left(C_{\tau}^{-1} + \frac{1}{\gamma^2}B\right)^{-1},
    \label{eq:mean and cov}
  \end{equation}
  where $B = H^TH$ and $\by_m^T$ are the rows of $Y$.

  \item  The rows $\bu_m^T$ of $\ U \sim \mu^Y$
are i.i.d.~according to the Gaussian distribution $\mathcal{N}\left( \bu_m^\ast, \poC\right)$.
\end{enumerate}

\end{proposition}
\endgroup

\begin{proof}
To show (i) we begin by expressing the likelihood in terms of the rows of $U$ and $Y$,

\[
  \exp\left(-\Phi(U;Y)\right) = \exp\left(-\frac{1}{2\gamma^2}
  \Fnorm{H U^T - Y^T}^2\right)  = \exp\left(-\frac{1}{2\gamma^2}\sum_{m=1}^M \left|H\bu_m - \by_m \right|^2\right).
\]

Combining with \eqref{mu-0-generic-form} we can express the Lebesgue density of the posterior
 as 
\begin{dgroup*}
\begin{dmath*}
  \mu^Y(\mathrm{d}U) \propto\exp \left[-\frac{1}{2}\sum_{m=1}^M
    \left\langle \bu_m, C_{\tau}^{-1}\bu_m\right\rangle
    + \frac{1}{\gamma^2}\left|H\bu_m-\by_m\right|^2
    \right] \\
\end{dmath*}
\begin{dmath*}
    = \exp \left[-\frac{1}{2}\sum_{m=1}^M 
    \left\langle \bu_m, C_{\tau}^{-1}\bu_m\right\rangle
    + \frac{1}{\gamma^2}\left( \langle \bu_m, B \bu_m\rangle -2 \langle \bu_m, H^T\by_m \rangle + \left| \by_m\right|^2\right)
    \right] \\
\end{dmath*}
\begin{dmath*}
    \propto\exp \left[-\frac{1}{2}\sum_{m=1}^M
    \Langle \bu_m, \left(\poC\right)^{-1}\bu_m\Rangle
    -2 \Langle \bu_m, \frac{1}{\gamma^2}H^T\by_m \Rangle + \Langle \pom_m, \left(\poC\right)^{-1} \pom_m \Rangle \right] \\
\end{dmath*}
\begin{dmath*}
    =\exp \left[-\frac{1}{2}\sum_{m=1}^M
    \Langle \bu_m, (\poC)^{-1}\bu_m\Rangle
    -2 \Langle \bu_m, \left(\poC\right)^{-1} \pom_m \Rangle + \Langle \pom_m, \left(\poC\right)^{-1} \bu_m^* \Rangle 
    \right] \\
\end{dmath*}
\begin{dmath*}
    =\exp \left[-\frac{1}{2}\sum_{m=1}^M
    \Langle \bu_m - \pom_m, (\poC)^{-1}\left(\bu_m -\pom_m\right)\Rangle \right] \\
\end{dmath*}
\begin{dmath*}
    = \exp\left[-\frac{1}{2}\left\langle (U - \poU)^T, (\poC)^{-1}\left(U- \poU\right)^T
  \right\rangle_F\right],
\end{dmath*}
\label{eq:posterior-matrix}
\end{dgroup*}
with $\pom_m$,  and $\poC$  as in \eqref{eq:mean and cov}.
Assertion (ii) follows from \eqref{posterior-product-form}, and
the observation that the negative log posterior is a sum of identical
positive-definite quadratic forms in each $\bu_m$, from which the expressions for mean
and variance of $\bu_m$ may be inferred. Assertion (iii) is a consequence of the
fact that uncorrelated Gaussian random variables are also independent. 
\end{proof}

\subsection{Proofs of Lemmata~\ref{lemma:1-disc}, \ref{lemma:2-disc}, and \ref{lemma:3-disc}}
\label{ssec:A2}

\subsubsection{Proof of Lemma~\ref{lemma:1-disc}}
\begin{proof}
  Let $P_0\in\mathbb{R}^{N\times N}$ denote the projection matrix onto
  $\mathrm{span}\{\bchi_k\}_{k=1}^K$ (recall  \eqref{bchi-k-definition}) and define 
      \begin{equation}\label{eq:beta-disc}
    \beta = \sqrt{\frac{K}{K+\zeta^2/4}}, \qquad   \zeta := \min_{k\le K}\min_{i\in Z_k} |\bchi_{k}(i)|.
  \end{equation}
  Our method of proof is to obtain lower bounds on the
  Dirichlet energy $\left\langle \bv, (\poC_0)^{-1}\bv \right\rangle$ for unit vectors $\bv \in \mathbb{R}^{N}$
  by considering two cases where  $|P_0\bv| \ge \beta$ of $|P_0 \bv| < \beta$. This translates to a
  lower bound on the smallest eigenvalue of $(\poC_0)^{-1}$. Since 
  $ {\rm Tr}(\poC_0) = \sum_{j=1}^N \lambda_{j,0} $, with $\lambda_{j,0}$ denoting the strictly positive
  eigenvalues of $\poC_0$, the lower bound on the Dirichlet energy of $(\poC_0)^{-1}$
  translates to an upper bound on ${\rm Tr}(\poC_0)$.

  {\it Case 1 ($|P_0 \bv | \ge \beta$):}
  Since $\bv$ is a unit vector, we have that
  $\|(I-P_0)\bv\|_\infty \le |(I-P_0)\bv| \le \sqrt{1-\beta^2}$.
  The matrix $C_{\tau, 0}$ and its inverse are positive definite, and so

  \begin{equation}\label{eq:dirichlet-energy-lower-bound}
    \left\langle \bv, (\poC_0)^{-1}\bv\right\rangle
    = \left\langle \bv, \left( \frac{1}{\gamma^2}
    B\bv + C_{\tau,0}^{-1} \right)\bv \right\rangle
    \ge \left\langle \bv,
    \frac{1}{\gamma^2}B\bv \right\rangle
    =   \frac{1}{\gamma^2}\sum_{i\in Z'}
    v_i^2,
  \end{equation}
  where we used $v_i$ to denote the entries of $\bv$. 
Let us write $P_0\bv = \sum_{k=1}^K c_k\bchi_k$ with $c_k := \langle \bv, \bchi_k\rangle$
  denoting the basis coefficients of $\bv$ in
span of  $\left\{\bchi_k\right\}_{k=1}^K$ 
and define
  \begin{equation*}
    \tk := \argmax_k |c_k|,
  \end{equation*}
  the index of the absolutely maximal coefficient amongst the $c_k$.
    The assumption $|P_0\bv| \ge \beta$ implies
  $\sum_{k=1}^K c_k^2 \ge \beta^2$.
  It then follows that
  \[
    K\max_{k \le K}c_k^2 \ge \sum_{k=1}^Kc_k^2
    \ge \beta^2,
  \]
  hence    $ |c_{\tk}| = \max_{k \le K}|c_k| \ge \beta/\sqrt{K}$.
  Since each $\bchi_k$ is supported on $\tZ_k$ on which
  it takes values that are at least $\zeta$, we have
  \[
    | (P_0\bv)_i| = |c_{\tk}| ( \bchi_{\tk})_i
    \ge \frac{\beta \zeta}{\sqrt{K}} \qquad
    \text{for} \qquad i\in \tZ_{\tk},
  \]
  where we used $(P_0 \bv)_i$ to denote the $i$-th entry of the  vector $ P_0 \bv$. 
  It then follows that for $i\in \tZ_{\tk}$
  \begin{align*}
    |v_i| & = |(P_0\bv)_i + ((I-P_0)\bv )_i|
    \ge \max\left\{0, | (P_0\bv)_i| - \| (I-P_0)\bv\|_\infty\right\}                 \\
                & \ge \max\left\{0, \frac{\beta \zeta}{\sqrt{K}} -
    \sqrt{1-\beta^2}\right\}.           
  \end{align*}
  Substituting the value of $\beta$ from \eqref{eq:beta-disc}, we obtain
  %
    $|v_i| \ge \left(4K/\zeta^2 + 1\right)^{-1/2}$.
  %
  Following Assumption~\ref{assumption:Zprime},
  i.e. $\tZ'_{k}\neq\emptyset$ for all $k$, we have
  %
  \[
    \frac{1}{\gamma^2}\sum_{j \in Z'}v_j^2
    \ge \frac{1}{\gamma^2} |v_i|^2 \ \ge \gamma^{-2} \left(4K/\zeta^2 + 1\right)^{-1} \qquad 
    \text{for some index $i \in \tZ_{\tk}'$}.
  \]
  %
  Putting this lower bound together with \eqref{eq:dirichlet-energy-lower-bound}
  we conclude that for any $\bv$ such that $|P_0\bv|\ge \beta$,
  \[
    \langle \bv, (\poC_0)^{-1}\bv\rangle
    \ge \gamma^{-2} \left(4K/\zeta^2 + 1\right)^{-1}.
  \]
{\it Case 2 ($|P_0 \bv | < \beta$):}
  We naturally have $|(I-P_0)\bv|
    \ge \sqrt{1-\beta^2}$. Let $\{(\sigma_{k,0}, \bm{\phi}_{k,0} ) \}_{k=1}^N$ denote the eigenpairs of $L_{0}$, indexed by order of increasing eigenvalues. Recall from Subsection~\ref{sec:disconnected-graph} that $\sigma_{k,0} = 0$ for $k=1, 2, \ldots, K$ and $\{\bm{\phi}_{k,0}\}_{k=1}^K \subset \mathrm{span}\{ \bchi_k \}_{k=1}^K$. Moreover,
  the orthonormal eigenvectors $\{ \bm{\phi}_{k,0}\}_{k=1}^N$ are also eigenvectors of $C_{\tau, 0}^{-1}$.
With some abuse of notation we define $c_k := \langle \bv, \bm{\phi}_{k,0} \rangle$ for 
$k = K+1, \dots, N$ and write 
  $(I-P_0)\bv = \sum_{k=K+1}^N c_k\bm{\phi}_{k,0}.$
  In light of this identity we compute
  \begin{multline}    \label{eq:spectralgap_other}
    \langle \bv, (\poC_0)^{-1} \bv\rangle
    =  \left\langle\bx,
    \left(\frac{1}{\gamma^2}B + C^{-1}_{\tau,0} \right) \bv
    \right\rangle                          
    \ge\langle \bv, C_{\tau,0}^{-1}\bv\rangle   \\
    =\sum_{k=1}^K c_k^2 + \sum_{k=K+1}^N c_k^2 \tau^{-2\alpha}
    (\sigma_{k,0} + \tau^2)^\alpha                      
     \ge\sum_{k=K+1}^N c_k^2 \tau^{-2\alpha}
    (\sigma_{k,0} + \tau^2)^\alpha.
\end{multline}
  Here we have used the fact that $B$ is positive semi-definite in the first inequality.
  From Assumption~\ref{assumption:G0}(b), 
  it follows that $\sigma_{k,0} \ge \theta$ for $k \ge K$, and subsequently $\sigma_{k,0} + \tau^2 \ge \theta$ for $k \ge K$. With this observation and using the expression for $\beta$ in \eqref{eq:beta-disc},
  we further continue the calculation in  \eqref{eq:spectralgap_other} to obtain the lower bound
  \begin{align*}
    \langle \bv, (\poC_0)^{-1} \bv \rangle
     & \ge\sum_{k = K+1}^N c_k^2 \tau^{-2\alpha}\theta^\alpha =  \tau^{-2\alpha}\theta^\alpha|(I-P_0)\bv|^2        \\
     & \ge\frac{1}{4}\tau^{-2\alpha}\theta^\alpha \left(4K/\zeta^2 + 1\right)^{-1}.
  \end{align*}
  Putting together the lower bounds from Cases 1 and 2 gives
  \[
    \langle \bv, (\poC_0)^{-1}\bv \rangle\ge \min
    \left\{
    \gamma^{-2}(4K/\zeta^2+1)^{-1},
    \frac{1}{4}\tau^{-2\alpha}\theta^\alpha(4K/\zeta^2+1)^{-1}
    \right\}
  \]
  for all unit vectors $\bv$ and constants $\gamma, \tau, \alpha >0$.
  Since the trace of a matrix coincides with the sum of its eigenvalues,
  we conclude that
  \[
    \mathrm{Tr}(\poC_0) \le N\max
    \left\{
    \gamma^2(4K/\zeta^2+ 1),
    4\tau^{2\alpha}\theta^{-\alpha}\left(4K/\zeta^2+1\right)
    \right\},
  \]
  from which the  desired result  follows by taking
  $\Xi = N\left(4K/\zeta^2+1\right) \max\left\{  1, 4\theta^{-\alpha} \right\}.$
\end{proof}

\subsubsection{Proof of Lemma~\ref{lemma:2-disc}}
\begin{proof}
  Recall \eqref{eq:C-0-ast-definition}.
  Then 
  \[
    \poC_0 = \poC_0\left(\frac{1}{\gamma^2}B + C_{\tau,0}^{-1}\right) \poC_0
    = \frac{1}{\gamma^2}\poC_0 B\poC_0 + \poC_0 C_{\tau,0}^{-1}C^\ast,
  \]
  which gives the identity
  \[
    \mathrm{Tr}\left(\frac{1}{\gamma^2}\poC_0 B \poC_0\right)
     = \mathrm{Tr}\left(\poC_0\right) 
     - \mathrm{Tr}\left(\poC_0 C_{\tau,0}^{-1}\poC_0\right).
  \]
  Both $\poC_0$ and $C_{\tau,0}^{-1}$ are positive definite and so is their product
  $\poC_0 C_{\tau,0}^{-1} \poC_0$. Therefore, $\mathrm{Tr}\left(
    \poC_0 C_{\tau,0}^{-1} \poC_0 \right)\ge 0$ and  so using Lemma~\ref{lemma:1-disc} we have
  $
    \mathrm{Tr}\left(\frac{1}{\gamma^2} \poC_0 B \poC_0 \right)\le
    \mathrm{Tr}\left( \poC_0 \right) \le \Xi \max\left\{ \gamma^2, \tau^{2\alpha}\right\}.$
  {}
\end{proof}

\subsubsection{Proof of Lemma~\ref{lemma:3-disc}}
\begin{proof}
  Choose  any vector $\bv \in {\rm span} \{ \bchi_1, \dots, \bchi_K\}$ and recall
  \eqref{eq:C-0-ast-definition}, the definition of $\poC_0$. Then 
  \begin{multline*}
    \left| \frac{1}{\gamma^2}\poC_0 B\bv - \bv\right|
     =  \left|
    \poC_0 \left(\frac{1}{\gamma^2}B\bv - (\poC_0)^{-1}\bv\right)
    \right|
     \le\left\|\poC_0\right\|_2 \left|\frac{1}{\gamma^2}B\bv - (\poC_0)^{-1}\bv\right| \\
     =  \left\|\poC_0\right\|_2 \left| C_{\tau, 0}^{-1}\bv\right|                       
     \le \mathrm{Tr}(\poC_0) \left| C_{\tau, 0}^{-1}\bv\right|,
  \end{multline*}
  where we remind that the two norm of a symmetric positive definite matrix is bounded above by its trace.
  Recall from Subsection~\ref{sec:disconnected-graph} that the vectors $\bchi_k$ are eigenvectors of $L_0$ corresponding to an eigenvalue of $0$, and so they are also eigenvectors of $C_{\tau, 0}^{-1}$ with attendant
  eigenvalue $1$. Therefore, since $\bv\in\mathrm{span}\left\{\bchi_k\right\}_{k=1}^K$ it follows
  that $C_{\tau,0}^{-1}\bv = \bv$. Using this fact and Lemma~\ref{lemma:1-disc} we  conclude that
  \[
    \left| \frac{1}{\gamma^2}\poC_0 B\bv - \bv \right|\le \Xi\max\{\gamma^2, \tau^{2\alpha}\}|\bv|.
  \]
The desired bound for the vectors $\bu^\dagger_m$ now follows trivially from Assumption~\ref{assumption:U_dagger}. 
\end{proof}

\subsection{Proofs Of Lemmata~\ref{lemma:1}, \ref{lemma:2}, and \ref{lemma:3}}
\label{ssec:A3}

\subsubsection{Proof of Lemma~\ref{lemma:1}}

\begin{proof}
  We use a similar argument to the proof of Lemma~\ref{lemma:1-disc} and obtain lower bounds
  on the Dirichlet energy  $\langle \bv, (\poC_\eps)^{-1} \bv \rangle$
  for unit vectors $\bv \in \mbb R^N$.  Recall $P_0\in\mathbb{R}^{N\times N}$ denotes the projection matrix onto
  $\mathrm{span}\{\bchi_k\}_{k=1}^K$ and define $\zeta, \beta$ as in \eqref{eq:beta-disc}. Once
  again we obtain the lower bounds in two cases where $|P_0 \bv | \ge \beta$ and
  $|P_0 \bv | < \beta$. 

  The case of $|P_0\bv| \ge \beta$ follows from identical arguments to Case 1 in the proof of
  Lemma~\ref{lemma:1-disc}. In fact, the lower bound \eqref{eq:dirichlet-energy-lower-bound} holds
  for $\poC_\eps$ replacing $\poC_0$ and
  so whenever $|P_0\bv|\ge \beta$ we have
  \[
    \langle \bv, (\poC_\eps)^{-1}\bv \rangle
    \ge \gamma^{-2}\left(4K/\zeta^2 + 1\right)^{-1}.
  \]
  So we focus on the 
case where  $|P_0\bv|<\beta$ and naturally $|(I-P_0)\bv|
    \ge \sqrt{1-\beta^2}$. Let $\{(\sigma_{j,\epsilon}, \bm{\phi}_{j,\epsilon} ) \}_{j=1}^N$ denote the eigenpairs of $L_{\epsilon}$, indexed by order of increasing eigenvalue. Note that these orthonormal eigenvectors are also eigenvectors of $C_{\tau, \epsilon}^{-1}$. We let $P_\epsilon \in \mathbb{R}^{N \times N}$
  denote the projection matrix onto $\mathrm{span}\{\bm{\phi}_{1,\epsilon},
  \bm{\phi}_{2,\epsilon}, \cdots, \bm{\phi}_{K,\epsilon}\}$. The key difference in this proof,
    compared to Case 2 in the proof of Lemma~\ref{lemma:1-disc}, is that we need to
    establish a lower bound on $|(I - P_\eps) \bv |$. We show that
    if $\eps \in (0, \eps_0)$ for a  sufficiently small constant $\eps_0$, then
    \begin{equation}
      \label{eq:lower-bound-on-I-P-eps}
      |(I - P_\eps)\bv| \ge \frac{1}{2} \sqrt{1 - \beta^2} = \frac{1}{2}(4 K/\zeta^2 + 1)^{-1/2}.
    \end{equation}
    We delay proving \eqref{eq:lower-bound-on-I-P-eps} until the end of the proof. Using \eqref{eq:C-eps-ast-definition} and the fact that $B$ is positive semi-definite 
    we can then write
    \begin{multline}\label{eq:spectralgap}
    \langle \bv, (\poC_\eps)^{-1} \bv\rangle
     =  \left\langle \bv, \left(  \frac{1}{\gamma^2}B + C^{-1}_{\tau,\epsilon}\right) \bv
    \right\rangle                                           \\
     \ge\langle \bv, C_{\tau,\epsilon}^{-1}\bv\rangle 
     \ge\sum_{j=K+1}^N c_{j,\eps}^2 \tau^{-2\alpha}
    (\sigma_{j,\epsilon} + \tau^2)^\alpha,
\end{multline}
where $c_{j,\eps} := \langle \bv, \bm{\phi}_{j,\eps} \rangle$.
By Lemma~\ref{appB-lemm39},
for $\eps \in (0, \eps_1)$ with $\eps_1 > 0$ sufficiently small to ensure that the entries of $W_\eps$ are non-negative, the graph Laplacian $L_\eps$ satisfies
an expansion of the form
\begin{equation*}
  L_\eps = L_0 + \sum_{h=1}^\infty \eps^h L^{(h)}
\end{equation*}
where $\{ \|L^{(h)}\|_2 \}_{h=1}^\infty \in \ell^\infty$.   Moreover, by Proposition~\ref{appB-Prop40}
and the binomial theorem we have that 
\begin{multline*}
  \tau^{-2\alpha}
  (\sigma_{K+1,\epsilon} + \tau^2)^\alpha
  \ge \tau^{-2\alpha} \left(  \theta + \tau^2  - \eps \sum_{h=1}^\infty
    \eps^{h-1} \| L^{(h)} \|_2 \right)^\alpha \\
    > \tau^{-2\alpha} \left( \theta + \tau^2 \right)^{\alpha} \left( 1 - \frac{\eps}{\tau^2 +\theta} \sum_{h=1}^\infty \eps^{h-1}
    \| L^{(h)} \|_2 \right)^\alpha  \\
  > \theta^\alpha \tau^{-2\alpha} \left( 1 - \frac{\eps}{\tau^2} \sum_{h=1}^\infty \eps^{h-1}
    \| L^{(h)} \|_2 \right)^\alpha
  \ge  \theta^\alpha \tau^{-2\alpha} \left( 1 - \frac{\eps}{\tau^2} \Xi_1 \right)^\alpha,
\end{multline*}
where  $\Xi_1 := \sup_{\eps \in (0,\eps_1)} \sum_{h=1}^\infty \eps^{h-1} \| L^{(h)}\|_2$
which is bounded provided that $\eps_1 <1$. Substituting this lower bound
back into \eqref{eq:spectralgap} and recalling the increasing ordering 
of the $\sigma_{j,\eps}$ we obtain
\begin{multline*}
    \langle \bv, (\poC_\eps)^{-1} \bv\rangle 
    \ge \theta^\alpha \tau^{-2\alpha} \left( 1 - \frac{\eps}{\tau^2} \Xi_1 \right)^\alpha  \sum_{j=K+1}^N c_{j,\eps}^2 \\
    =  \theta^\alpha \tau^{-2\alpha} \left( 1 - \frac{\eps}{\tau^2} \Xi_1 \right)^\alpha  |(I - P_\eps) \bv|^2
    \ge  \frac{1}{4} \theta^\alpha \tau^{-2\alpha} \left( 1 - \frac{\eps}{\tau^2} \Xi_1 \right)^\alpha 
    ( 4 K /\zeta^2 + 1)^{-1},
\end{multline*}
where we have invoked \eqref{eq:lower-bound-on-I-P-eps}.
Putting this bound together with the lower bound from the first case where $| P_0 \bv| \ge \beta$,
we conclude that
\begin{equation*}
      \langle \bv, (\poC_\eps)^{-1}\bv \rangle\ge \min
    \left\{
    \gamma^{-2}(4K/\zeta^2+1)^{-1},
    \frac{1}{4}\tau^{-2\alpha} (1  - \eps \tau^{-2} \Xi_1)^\alpha \theta^\alpha(4K/\zeta^2+1)^{-1}
    \right\}
\end{equation*}
from which it follows that
\begin{equation*}
      {\rm Tr}(\poC_\eps) \le  N \max
    \left\{
    \gamma^{2}(4K/\zeta^2+1),
    \frac{1}{4}\tau^{2\alpha} (1  - \eps \tau^{-2} \Xi_1)^{-\alpha} \theta^{-\alpha}(4K/\zeta^2+1)
    \right\}
\end{equation*}
provided that the $\eps_0>0$ for \eqref{eq:lower-bound-on-I-P-eps} to hold is sufficiently small
which concludes the proof of the Lemma. 

It remains for us to prove the bound \eqref{eq:lower-bound-on-I-P-eps}. By 
Proposition~\ref{appB-Prop39} and
\cite[Proof of Prop.~41]{HHRS19} there exist uniform constants  $\eps_2, \Xi_2 >0$ so that  $\forall \eps \in (0, \eps_2)$ and
for any unit vector $\bv$ 
\begin{equation*}
  | ( I - P_\eps)P_0 \bv |^2 \le \Xi_2 \eps^2 \quad \text{and} \quad  | ( I - P_0)P_\eps \bv |^2 \le \Xi_2 \eps^2,
\end{equation*}
implying that the range of $P_\eps$ and $P_0$ are close when $\eps$ is small. 
Therefore, using the fact that $P_0$ and $P_\eps$ are symmetric and idempotent, as well as 
the Cauchy-Schwarz inequality, we can write
\begin{multline*}
  \left|\left(P_0-P_\epsilon\right)\bv\right|^2
   = \left\langle
    (P_0 - P_\epsilon)\bv, P_0\bv
    \right\rangle -
    \left\langle
    (P_0 - P_\epsilon)\bv, P_\epsilon\bv
    \right\rangle \\
    =  \left\langle
    \bv, \left(P_0 - P_\epsilon\right)P_0\bv
    \right\rangle -
    \left\langle
    \bv, \left(P_0 - P_\epsilon\right)P_\epsilon\bv
  \right\rangle 
  =
  \left\langle
    \bv, (I-P_\epsilon)P_0\bv
    \right\rangle +
    \left\langle
    \bv, (I-P_0)P_\epsilon \bv
    \right\rangle \\
   \le |\bv| \left(
    \left|(I-P_\epsilon)P_0\bv\right| +
    \left|(I-P_0)P_\epsilon\bv\right|
    \right) 
   \le \Xi_3\epsilon.
\end{multline*}
We then calculate
\begin{multline*}
     |(I-P_\epsilon)\bv|
     =  |(I-P_0)\bv + (P_0-P_\epsilon)\bv|
      \ge\max\left\{
    0,
    |(I-P_0)\bv| - |(P_0-P_\epsilon)\bv|
    \right\}\\
     \ge\max\left\{
    0,
    \sqrt{1-\beta^2} - (\Xi_3\epsilon)^{1/2}
    \right\}
     \ge\frac{\sqrt{1-\beta^2}}{2} =  \frac{1}{2}(4K/\zeta^2+1)^{-1/2},
\end{multline*}
where the last inequality holds if $\epsilon \le \frac{1-\beta^2}{4\Xi_3}$. The lower bound \eqref{eq:lower-bound-on-I-P-eps} then follows by letting $\eps_0 := \min\left\{ \eps_1, \eps_2, \frac{1-\beta^2}{4\Xi_3}\right\}$.
{}
\end{proof}

\subsubsection{Proof of Lemma~\ref{lemma:2}}
\begin{proof}
The proof is nearly identical to that of Lemma~\ref{lemma:2-disc} and is hence omitted.  
\end{proof}

\subsubsection{Proof of Lemma~\ref{lemma:3}}
\begin{proof}
  We proceed similarly to the proof of Lemma~\ref{lemma:3-disc} by choosing a  vector $\bv \in
  {\rm span}\{ \bchi_k\}_{k=1}^K$. We then have
  \begin{multline*}
    \left| \frac{1}{\gamma^2}\poC_\eps B\bv - \bv\right|
     =  \left|
    \poC_\eps \left(\frac{1}{\gamma^2}B\bv - (\poC_\eps)^{-1}\bv\right)
    \right| \\
    \le\left\|\poC_\eps \right\|_2 \left|\frac{1}{\gamma^2}B\bv - (\poC_\eps )^{-1}\bv\right| 
     =  \left\|\poC_\eps \right\|_2 \left| C_{\tau, \epsilon}^{-1}\bv\right|.
  \end{multline*}
  Now decompose $\bv = P_\epsilon \bv + (I-P_\epsilon)\bv$.
  Since we assumed that $\bv\in\mathrm{span}\left\{\bar{\bm{\chi}}_\ell\right\}_{\ell=1}^K$,
  it follows from \cite[Prop.~39]{HHRS19} that
  $|(I-P_\epsilon)\bv| \le \Xi_3\epsilon |\bv|$ for some $\Xi_3 > 0$ independent of $\epsilon$,
  and so
  \begin{align*}
    \left|C_{\tau,\epsilon}^{-1}\bv \right|
     & \le \left|C_{\tau,\epsilon}^{-1}P_\epsilon \bv\right| + \left|C_{\tau,\epsilon}^{-1}(I-P_\epsilon)\bv\right|    \\
     & \le \max_{k\le K} \frac{(\sigma_{k,\epsilon}+\tau^2)^\alpha}{\tau^{2\alpha}} |P_\epsilon \bv| + \max_{k>K} \frac{(\sigma_{k,\epsilon} + \tau^2)^\alpha}{\tau^{2\alpha}}|(I-P_\epsilon)\bv| \\
     & \le \Xi_4 \left[  \left( 1 + \frac{\epsilon}{\tau^{2}} \right)^\alpha
       +  \epsilon\left(1 + \frac{1}{\tau^{2\alpha}}\right) \right] |\bv|.
  \end{align*}
The third inequality follows from 
Proposition~\ref{appB-Prop39}(ii) and
the fact that the  $\sigma_{k,\epsilon}$ are uniformly bounded
  for all $\eps \in (0, \eps_0)$ and $\eps_0 <1$. In fact, by Lemma~\ref{appB-lemm39}, we have that
\begin{align*}
      \sigma_{k,\epsilon} & =  \langle \bm{\phi}_{k, \epsilon}, L_\epsilon \bm{\phi}_{k, \epsilon} \rangle  \le \left| \langle \bm{\phi}_{k, \epsilon}, L_0 \bm{\phi}_{k, \epsilon} \rangle \right| + \sum_{h=1}^\infty \epsilon^h \left| \langle \bm{\phi}_{k, \epsilon}, L_h \bm{\phi}_{k, \epsilon} \rangle \right| \\
      &\le \|L_0\|_2 + \frac{\epsilon}{1-\epsilon} \left( \max_{h=1,2,\ldots} \|L_h\|_2 \right) \le  \frac{1}{1 - \eps}  \left( \max_{h=0,1,\ldots} \|L_h\|_2 \right).
\end{align*} 
Now bounding $\|\poC_\eps\|_2$ by $\mathrm{Tr}(\poC_\eps)$ and envoking Lemma~\ref{lemma:1} yields
  \begin{align*}
    \|\poC_\eps\|_2\left|C_{\tau,\epsilon}^{-1}\bv\right|
    &\le \Xi_0 \Xi_4
      \max\left\{ \gamma^2,  \left( \frac{\tau^{2}}{1 - \Xi_1 \eps/\tau^2} \right)^\alpha\right\}
      \left[ \eps + \frac{\eps}{\tau^{2\alpha}} + \left( 1 + \frac{\eps}{\tau^2} \right)^\alpha  \right] |\bv|.
  \end{align*} 
  The theorem follows by setting $\Xi_2 = \Xi_0\Xi_4$.
\end{proof}

\section{Summary of Results from~\cite{HHRS19}}\label{app:results-HHRS}

In this section, we briefly state and discuss the spectral analysis results from~\cite{HHRS19} that are used throughout our 
proofs in Section~\ref{sec:proof-main-results}. These are Davis-Kahan type results \cite{davis1970rotation} 
which  give quantitative bounds on the distances between the spectra of graph Laplacians and their 
perturbations in terms of the parameters  $\eps$ and $\tau$ introduced in our definition of nearly-disconnected graphcs.

\begin{lemma}[Lemma 38 of~\cite{HHRS19}]\label{appB-lemm39}
    Let $W_\eps$ be as in \eqref{W-eps-expansion} and suppose Assumptions~\ref{assumption:G0} and \ref{assumption:G_k} are satisfied. Let $L_\eps$ be as in \eqref{L-eps-and-C-tau-eps-def}. Then there exists $\eps_0 > 0$ so that for all $\eps \in (0, \eps_0)$ the matrix $W_\eps$ has non-negative weights and the graph Laplacian operator $L_\eps$ satisfies an expansion of the form
    \[
        L_\eps = L_0 + \sum_{h=1}^\infty \eps^h L^{(h)}
    \]
    with $\{\|L^{(h)}\|_2\}_{h \in \mathbb{N}} \in \ell^\infty$.
\end{lemma}

This lemma provides the existence of an $\eps_0 > 0$ so that the matrix $W_\eps$ has non-negative weights, since the constraints on the entries $w_{ij}^{(h)}$ of Assumption~\ref{assumption:G_k} allow for negative weights. Ensuring the non-negativity of the entries of $W_\eps$ is necessary for the proper definition of a graph Laplacian matrix and the expansion we obtain.

\begin{proposition}[Proposition 39 of~\cite{HHRS19}] \label{appB-Prop39}
    Suppose Assumptions~\ref{assumption:G0} and \ref{assumption:G_k} are satisfied and let $\{\lambda_{j,\eps}, \bphi_{j,\eps}\}$ denote the orthonormal eigenpairs of $C_{\tau, \eps}^{-1}$. Then there exists $\eps_0 > 0$ so that
    \begin{itemize}
        \item[(i)] $\lambda_{1, \eps} = 1$ and $\bphi_{1,\eps} = \bchi$, where $\bchi = \frac{\sum_{k=1}^K \bchi_k}{\left\| \sum_{k=1}^K \bchi_k \right\|_2}$;
        \item[(ii)] for all $\eps \in (0, \eps_0)$, there exists constants $\Xi_1(K, \|L^{(1)}\|_2) > 0$ and $\Xi_2(K,$ $\sup_{h\ge 2} \|L^{(h)}\|_2, \eps_0) > 0$ independent of $\eps$ with property that
        \[
            \lambda_{k,\eps} \le \left( 1 + \Xi_2\eps \tau^{-2} + \Xi_2 \eps^2 \tau^{-2} \right)^\alpha, \qquad \forall k \in \{1, 2, \ldots, K\};
        \]
        \item[(iii)] if there exists a uniform constant $\vartheta > 0$ so that $\lambda_{K+1, \eps} - 1 \ge \vartheta$ then there exists a constant $\Xi_3(K, \|L^{(1)}\|_2, \vartheta) > 0$ independent of $\eps \in (0, \eps_0)$ with property that
        \[
            \left| (I - P_0) P_\eps\bphi_{j, \eps}\right|^2 = \left| 1 - \sum_{j=1}^K \langle \bphi_{j, \eps}, \bchi_k\rangle^2 \right| \le \Xi_3 \eps^2 + \mcl O(\eps^3), \quad \forall k \in \{1, 2, \ldots, K\}.
        \]
    \end{itemize}
\end{proposition}

The above result gives  bounds on the first $K$ eigenvalues of the graph Laplacian matrix $L_\eps$ and 
characterizes the geometry of the corresponding eigenvectors in relation to the weighted set functions $\bchi_k$ that correspond to the subgraphs $\tG_k$. This shows that for properly scaled $\eps$ and $\tau$, we can ensure that these eigenvalues and eigenvectors are ``close'' to the $\bchi_k$.

\begin{proposition}[Proposition 40 of~\cite{HHRS19}]\label{appB-Prop40}
    Suppose Assumptions~\ref{assumption:G0} and \ref{assumption:G_k} are satisfied. Then the eigenvalues $\sigma_{K+1,\eps}$ of $L_\eps$ and the eigenvalues $\lambda_{K+1,\eps}$ of $C_{\tau, \eps}^{-1}$ satisfy the bounds
    \[
        \sigma_{K+1, \eps} \ge \theta - \sum_{h=1}^\infty \eps^h \|L^{(h)}\|_2 \qquad \text{and} \qquad \lambda_{K+1,\eps} \ge \tau^{-2\alpha} \left( \tau^2 + \theta - \sum_{h=1}^{\infty} \eps^h \|L^{(h)}\|_2\right)^\alpha,
    \]
    where $\theta > 0$ is the constant appearing in Assumption~\ref{assumption:G0}(b).
\end{proposition}

This result gives useful lower bounds on the high frequency eigenvalues of $L_\eps$ and $C_{\tau, \eps}^{-1}$ allowing us to bound away from $0$ the eigenvalues of $(\poC_\eps)^{-1}$.  These lead to spectral gap results that are crucial in 
characterizing the geometry of the eigenvectors and in turn obtaining posterior contraction rates.


\bibliographystyle{siamplain}
\bibliography{references}


\end{document}